\documentclass[runningheads,a4paper]{llncs}

\usepackage[english]{babel}         
\usepackage[utf8]{inputenc}         
\usepackage{amsmath}                
\usepackage{amsfonts}               
\usepackage{amssymb}
\usepackage{latexsym}               
\usepackage{stmaryrd}               
\usepackage{exscale}                
\usepackage{verbatim}               
\usepackage[pdftex]{graphicx}       
\usepackage[pdftex]{hyperref}       
\usepackage{hypbmsec}               
\usepackage{listings}               
\usepackage{color}
\usepackage{array}
\usepackage{rotating}				
\usepackage{pdflscape}				
\usepackage{longtable}				
\usepackage{colortbl}				
\usepackage{ifthen}
\usepackage{todonotes}	

\hypersetup{
	plainpages=false, pdfpagelabels,
	pdfauthor={Michael Schneider, Geoff Sutcliffe},
    pdftitle={Reasoning in the OWL 2 Full Ontology Language using First-Order Automated Theorem Proving},
    pdfkeywords={Semantic Web, OWL, First-order logic, ATP}
}
\newcounter{IsExtendedPaper}
\newcommand{\SetExtendedPaper}{\setcounter{IsExtendedPaper}{1}}
\newcommand{\IfExtendedPaperThenElse}[2] {%
	\ifthenelse{\value{IsExtendedPaper}=1}{{#1}}{{#2}}%
}

\newcommand{\CellEmpty}{{\cellcolor[rgb]{0.00,0.00,0.00}}}

\newcommand{\CellTextSuccess}{$\mathbf{+}$}
\newcommand{\CellColorSuccess}{\cellcolor[rgb]{0.67,1.00,0.67}}
\newcommand{\CellSuccess}{\CellTextSuccess\CellColorSuccess}

\newcommand{\CellTextWrong}{$\mathbf{-}$}
\newcommand{\CellColorWrong}{\cellcolor[rgb]{1.00,1.00,1.00}}
\newcommand{\CellWrong}{\CellTextWrong\CellColorWrong}

\newcommand{\CellTextUnknown}{$\mathbf{?}$}
\newcommand{\CellColorUnknown}{\cellcolor[rgb]{1.00,1.00,1.00}}
\newcommand{\CellUnknown}{\CellTextUnknown\CellColorUnknown}

\newcommand{\FullishSuiteTableEntryColumnWidth}{0.49\textwidth}
\newcommand{\FullishSuiteTableEntryFontSize}{scriptsize}
\newcommand{\FullishSuiteTableEntryTitleInconsistencyGraph}{Graph}
\newcommand{\FullishSuiteTableEntryTitleEntailmentPremiseGraph}{Premise Graph}
\newcommand{\FullishSuiteTableEntryTitleEntailmentConclusionGraph}{Conclusion Graph}

\SetExtendedPaper

\begin{document}

\mainmatter

\title{Reasoning in the OWL 2 Full Ontology Language using First-Order Automated Theorem Proving}

\titlerunning{Reasoning in OWL~2 Full using First-Order ATP}

\author{%
Michael Schneider\inst{1}\thanks{%
Partially supported by the projects 
\emph{SEALS} 
(European Commission, EU-IST-2009-238975)
and 
\emph{THESEUS} 
(German Federal Ministry of Economics and Technology, FK~OIMQ07019).%
}
\and 
Geoff Sutcliffe\inst{2}
}
\institute{%
FZI Research Center for Information Technology, Germany\\
\and
University of Miami, USA%
}

\maketitle

\begin{abstract}
OWL~2 has been standardized by the World Wide Web Consortium (W3C) as a family of ontology languages for the Semantic Web. The most expressive of these languages is OWL~2 Full, but to date no reasoner has been implemented for this language. Consistency and entailment checking are known to be undecidable for OWL~2 Full. We have translated a large fragment of the OWL~2 Full semantics into first-order logic, and used automated theorem proving systems to do reasoning based on this theory. The results are promising, and indicate that this approach can be applied in practice for effective OWL reasoning, beyond the capabilities of current Semantic Web reasoners.

\IfExtendedPaperThenElse{%
This is an \emph{extended version} of
a paper with the same title 
that has been published
at CADE~2011, LNAI 6803, pp.~446--460.
The extended version provides appendices with additional resources
that were used in the reported evaluation.
}%
{}

\end{abstract}

\section{Introduction}
\label{toc:introduction}
%
The Web Ontology Language OWL~2~\cite{w3c09-owl2-document-overview}
has been standardized
by the World Wide Web Consortium~(W3C)
as a family of ontology languages for the Semantic Web.
OWL~2 includes 
OWL~2 DL~\cite{w3c09-owl2-structural-specification}, 
the OWL~2 RL/RDF rules~\cite{w3c09-owl2-profiles}, 
as well as OWL~2 Full~\cite{w3c09-owl2-rdf-semantics}.
The focus of this work is on reasoning in OWL~2 Full,
the most expressive of these languages.
So far,
OWL~2 Full has largely been ignored by the research community,
and no reasoner has been implemented for this language.

%
OWL~2 Full does not enforce any of the numerous syntactic restrictions 
of the description logic-style language OWL~2 DL.
Rather, OWL~2 Full treats arbitrary RDF graphs~\cite{w3c04-rdf-concepts} 
as valid input ontologies,
and can safely be used 
with weakly structured RDF data
as is typically found on the Web.
Further, 
OWL~2 Full provides for reasoning 
outside the scope of 
OWL~2 DL 
and the OWL~2 RL/RDF rules,
including sophisticated reasoning based on meta-modeling.
In addition, OWL~2 Full is semantically fully compatible
with RDFS~\cite{w3c04-rdf-semantics} 
and also with the OWL~2 RL/RDF rules,
and there is even a strong semantic correspondence~\cite{w3c09-owl2-rdf-semantics} 
with OWL~2 DL,
roughly stating that any OWL~2 DL conclusion 
can be reflected in OWL~2 Full.
This makes OWL~2 Full largely interoperable with the other OWL~2 languages,
and allows an OWL~2 Full reasoner to be combined
with most existing OWL reasoners 
to provide higher syntactic flexibility and semantic expressivity
in reasoning-enabled applications.

%
Due to its combination of flexibility and expressivity, 
OWL~2 Full is computationally undecidable with regard to
consistency and entailment checking~\cite{motik07-metamodeling}.
While there cannot be any complete decision procedure for OWL~2 Full,
the question remains 
to what extent practical OWL~2 Full reasoning is possible.
%
This paper presents the results of a series of experiments
about reasoning in OWL~2 Full  
using first-order logic (FOL) theorem proving.
A large fragment of the OWL~2 Full semantics has been translated
into a FOL theory,
and automated theorem proving (ATP) systems have been used
to do reasoning based on this theory.
The primary focus of these experiments was on the question 
of what can be achieved at all;
a future study may shift the focus to efficiency aspects.

%
The basic idea used in this work is not new.
An early application of this approach 
to a preliminary version of RDF and a precursor of OWL
was reported by Fikes et al.~\cite{fikes02-semweb-fol-semantics}.
That work focused on identifying technical problems 
in the original language specifications,
rather than on practical reasoning.
Hayes~\cite{hayes05-swlang-cl-translation} 
provided fairly complete translations
of RDF(S) and OWL~1 Full into Common Logic,
but did not report on any reasoning experiments.
This gap was filled by Hawke's reasoner 
\emph{Surnia}~\cite{hawke03-surnia},
which applied an ATP system to an FOL axiomatisation of OWL~1 Full.
For unknown reasons, however,
Surnia performed rather poorly
on reasoning tests~\cite{w3c04-owl1-testresults}.
Comparable studies have been carried out 
for ATP-based OWL DL reasoning,
as for \emph{Hoolet}~\cite{tsarkov04-hoolet},
an OWL DL reasoner implemented
on top of a previous version of the Vampire ATP system 
(\url{http://www.vprover.org}).
The work of Horrocks and Voronkov~\cite{horvor06-ontology-theorem-prover}
addresses reasoning over large ontologies,
which is crucial for practical Semantic Web reasoning.
Finally,
\cite{baader03-dl-handbook} reports on some historic 
knowledge representation systems
using ATP for description logic-style reasoning,
such as \emph{Krypton} in the~1980s.

%
All these previous efforts are outdated, in that they refer to precursors
of OWL~2 Full, and
appear to have been discontinued after publication.
The work reported in this paper refers
to the current specification of OWL~2 Full,
and makes a more extensive experimental evaluation of the FOL-based
approach than any previous work.
Several aspects of OWL~2 Full reasoning have been studied:
the degree of language coverage of OWL~2 Full;
semantic conclusions that are characteristic specifically of OWL~2 Full;
reasoning on large data sets;
and the ability of first-order systems 
to detect non-entailments and consistent ontologies in OWL~2 Full.
The FOL-based results have been compared
with the results of a selection of well-known Semantic Web reasoners,
to determine whether the FOL-based approach
is able to add significant value 
to the state-of-the-art in Semantic Web reasoning.

%
This paper is organized as follows:
Section~\ref{toc:preliminaries} provides an
introduction to the technologies used in this paper.
Section~\ref{toc:approach} describes the FOL-based reasoning approach.
Section~\ref{toc:setting} describes the evaluation setting,
including the test data, 
the reasoners, 
and the computers 
used in the experiments.
The main part of the paper is Section~\ref{toc:results},
which presents the results of the experiments.
Section~\ref{toc:conclusions} concludes,
and gives an outlook on possible future work.
\IfExtendedPaperThenElse{%
The appendices present
the raw result data underlying the evaluation results
(\ref{toc:detailedresults});
the complete test suite of ``characteristic OWL~2 Full conclusions''
that has been used in the evaluation
(\ref{toc:fullishtestsuite});
and 
an example showing how RDF data and the semantics of OWL~2 Full
have been translated into the first-order logic formalism
(\ref{toc:tptptranslation}).
}%
{}

\section{Preliminaries}
\label{toc:preliminaries}
\subsection{RDF and OWL 2 Full}
\label{toc:preliminaries:owl2full}

OWL~2 Full is specified as the language
that uses the OWL~2 RDF-Based Semantics~\cite{w3c09-owl2-rdf-semantics}
to interpret arbitrary RDF graphs.
RDF graphs are defined by the 
RDF Abstract Syntax \cite{w3c04-rdf-concepts}.
The OWL~2 RDF-Based Semantics
is defined as a semantic extension
of the RDF Semantics \cite{w3c04-rdf-semantics}.

According to the RDF Abstract Syntax,
an \emph{RDF graph}~$G$ 
is a set of RDF triples:~$G = \{ t_1, \ldots, t_n \}$.
Each \emph{RDF triple}~$t$ is given 
as an ordered ternary tuple~$t = s\,p\,o$ of \emph{RDF nodes}.
The RDF nodes $s$, $p$, and $o$ 
are called 
the \emph{subject}, \emph{predicate}, and \emph{object}
of the triple~$t$, respectively.
Each RDF node is either a \emph{URI}, 
a (plain, language-tagged or typed) \emph{literal}, 
or a \emph{blank node}.

The \emph{RDF Semantics} is defined on top of the RDF Abstract Syntax 
as a model theory for arbitrary RDF graphs.
For an \emph{interpretation}~$I$ and a \emph{domain}~$U$,
a URI denotes an individual in the domain,
a literal denotes a concrete data value 
(also considered a domain element),
and a blank node is used as an existentially quantified variable
indicating the existence of some domain element.
The meaning of a triple~$t = s\,p\,o$ is a truth value
of the relationship
$\langle I(s),I(o) \rangle \in \text{IEXT}(I(p))$,
where $\text{IEXT}$ 
is a mapping from domain elements 
that are \emph{properties} to associated binary relations.
The meaning of a graph~$G = \{ t_1, \ldots, t_n \}$ is a truth value
determined by the conjunction of the meaning of all the triples,
taking into account the existential semantics 
of blank nodes occurring in~$G$.
If an RDF graph~$G$ is true under an interpretation~$I$,
then $I$~\emph{satisfies}~$G$.
An RDF graph~$G$ is \emph{consistent}
if there is an interpretation~$I$ that satisfies~$G$.
An RDF graph~$G$ \emph{entails} another RDF graph~$H$
if every interpretation~$I$ that satisfies~$G$ also satisfies~$H$.

Whether an interpretation satisfies a given graph
is primarily determined by a collection of
model-theoretic \emph{semantic conditions}
that constrain the mapping $\text{IEXT}$.
There are different sets of model-theoretic semantic conditions
for the different semantics defined by the RDF Semantics specification.
For example,
the semantics of class subsumption in \emph{RDFS} 
is defined mainly by the semantic condition
defined for the RDFS vocabulary term \texttt{rdfs:subClassOf}:
\[
\langle c,d \rangle \in \text{IEXT}(I(\texttt{rdfs:subClassOf})) 
\Rightarrow
c, d \in IC \, \wedge \, \text{ICEXT}(c) \subseteq \text{ICEXT}(d)
\]
where ``$c$'' and ``$d$'' are universally quantified variables.
Analogous to the mapping~$\text{IEXT}$, 
the mapping $\text{ICEXT}$ associates \emph{classes}
with subsets of the domain.
%
%
The two mappings are responsible
for the \emph{metamodeling capabilities} of RDFS 
and its semantic extensions:
Although the quantifiers in the RDFS semantic conditions 
range over exclusively domain elements,
which keeps RDFS in the realm of first-order logic,
the associations provided by the two mappings 
allow domain elements (properties and classes)
to indirectly refer to sets and binary relations.
This enables a limited but useful form
of higher order-style modeling and reasoning.

The \emph{OWL~2 RDF-Based Semantics},
i.e.\ the semantics of OWL~2 Full, 
extends the RDF Semantics specification
by additional semantic conditions
for the OWL-specific vocabulary terms,
such as \texttt{owl:unionOf} and \texttt{owl:disjointWith}.

\subsection{FOL, the TPTP language, and ATP}
\label{toc:preliminaries:atp}

The translation of the OWL~2 Full semantics is to classical untyped
first-order logic.
The concrete syntax is the TPTP language~\cite{Sut09}, which is the de facto
standard for state-of-the-art ATP systems for first-order logic.
The ATP systems used in the evaluation were taken from their web sites
(see Section~\ref{toc:setting:reasoners}) or from the archives of the 5th 
IJCAR ATP System Competition, CASC-J5 (\url{http://www.tptp.org/CASC/J5/}).
Most of the systems are also available online as part of the SystemOnTPTP
service (\url{http://www.tptp.org/cgi-bin/SystemOnTPTP/}).

\section{Approach}
\label{toc:approach}

Each of the model-theoretic semantic conditions 
of the OWL~2 Full semantics
is translated into a corresponding FOL axiom.
The result is an axiomatization of OWL~2 Full.
The RDF graphs to reason about 
are also converted into FOL formulae.
In the case of \emph{consistency checking}
there is a single RDF graph that is converted into a FOL axiom,
for which satisfiability needs to be checked.
In the case of \emph{entailment checking},
there is a premise graph that is converted into a FOL axiom, and a conclusion 
graph that is converted into a FOL conjecture.
The FOL formulae 
(those representing the input RDF graphs
and 
those building the FOL axiomatization of the OWL~2 Full semantics)
are passed to an ATP system,
which tries to prove the conclusion or establish consistency.


We apply a straight-forward \emph{translation of the semantic conditions},
making use of the fact that all semantic conditions 
have the form of FOL formulae.
A semantic relationship of the form
``$\langle s,\,o \rangle \in \text{IEXT}(p)$''
that appears within a semantic condition
is converted into an atomic FOL formula of the form
``$\text{iext}(p, s, o)$''.
Likewise, a relationship
``$x \in \text{ICEXT}(c)$''
is converted into
``$\text{icext}(c, x)$''.
Apart from this,
the basic logical structure of the semantic conditions is retained.
For example,
the semantic condition specifying RDFS class subsumption
shown in Section~\ref{toc:preliminaries:owl2full}
is translated into
\[
\begin{array}{ll}
\forall c,d:
[\;
&
	\text{iext}(\texttt{rdfs:subClassOf},c,d) \Rightarrow \\
&
(\,
	\text{ic}(c)
	\wedge \text{ic}(d)
	\wedge \forall x: ( \text{icext}(c,x) \Rightarrow \text{icext}(d,x) )
\,)
\;]
\end{array}
\]


The \emph{translation of RDF graphs}
amounts to converting the set of triples~``$s\,p\,o$'' 
into a conjunction of corresponding ``$\text{iext}(p,s,o)$'' atoms.
A \emph{URI} occurring in an RDF graph
is converted into a constant.
An \emph{RDF literal} is converted into a function term,
with a constant for the literal's lexical form as one of its arguments.
Different functions are used for the different kinds of literals:
function terms for \emph{plain literals} have arity~1; 
function terms for \emph{language-tagged literals} 
have a constant representing the language tag as their second argument;
function terms for \emph{typed literals}
have a constant for the datatype URI as their second argument.
For each \emph{blank node},
an existentially quantified variable is introduced,
and the scope of the corresponding existential quantifier
is the whole conjunction of the ``$\text{iext}$'' atoms.
For example, the RDF graph
\[
\begin{array}{l}
\texttt{\_:x}\; \texttt{rdf:type}\; \texttt{foaf:Person}\; \texttt{.} \\
\texttt{\_:x}\; \texttt{foaf:name}\; \texttt{"Alice"\^{}\^{}xsd:string}\; \texttt{.}
\end{array}
\]
which contains the blank node~``$\texttt{\_:x}$'', 
the typed literal ``$\texttt{"Alice"\^{}\^{}xsd:string}$'',
and the URIs ``$\texttt{rdf:type}$'', ``$\texttt{foaf:Person}$'', and ``$\texttt{foaf:name}$'', 
is translated into the FOL formula
\[
\begin{array}{ll}
\exists x: [\; 
	& \text{iext}(
		\texttt{rdf:type},
		x,
		\texttt{foaf:Person}
	) 
	\,\wedge \\
	& \text{iext}(
		\texttt{foaf:name},
		x,
		\texttt{literal}_\texttt{typed}(
			\texttt{Alice}, 
			\texttt{xsd:string}
		)
	)
\;]
\end{array}
\]

\section{Evaluation Setting}
\label{toc:setting}
This section describes the evaluation setting:
the OWL~2 Full axiomatization,
the test cases,
the reasoners, and
the computing resources.
\emph{Supplementary material}
including the 
axiomatizations, 
test data, 
raw results, 
and the software
used for this paper
can be found online at:\\
\mbox{\small{\url{http://www.fzi.de/downloads/ipe/schneid/cade2011-schneidsut-owlfullatp.zip}}}.

\subsection{The FOL Axiomatization and RDF Graph Conversion}
\label{toc:setting:axiomset}

Following the approach described in Section~\ref{toc:approach},
most of the normative semantic conditions of the OWL~2 Full semantics 
have been converted into the corresponding FOL axioms,
using the TPTP language~\cite{Sut09}.
The main omission is that most
of the semantics concerning \emph{reasoning on datatypes} 
has not been treated,
as we were only interested in evaluating the ``logical core'' of the language.
All other language features of OWL~2 Full were covered in their full form,
with a restriction that was sufficient for our tests:
while OWL~2 Full has many size-parameterized language features,
for example the intersection of arbitrarily many classes,
our axiomatization generally supports these language feature schemes
only up to a size of~3.
The resulting FOL axiomatization consists of 558~formulae.
The axiom set is fully first-order with equality, 
 but equality accounts for less than 10\% of the atoms.
The first-order ATP systems used (see Section~\ref{toc:setting:reasoners}) 
convert the formulae to clause normal form. 
The resultant clause set is non-Horn.
Almost all the clauses are range-restricted, which can result in reasoning 
that produces mostly ground clauses.

In addition,
a converter from RDF graphs to FOL formulae was implemented.
This allowed the use of RDF-encoded OWL test data in the experiments,
without time consuming and error prone manual conversion.

\subsection{Test Data}
\label{toc:setting:testdata}

Two complementary test suites were used for the experiments:
one test suite to evaluate the degree of language coverage of OWL~2 Full,
and another suite consisting of 
characteristic conclusions for OWL~2 Full reasoning.
For scalability experiments a large set of RDF data was also used.

\subsubsection{The Language Coverage Test Suite.}
\label{toc:setting:testdata:coverage}

For the language coverage experiments,
the test suite described in~\cite{schneid09-testsuite-owl2-rdfbased} was 
used.\footnote{%
There is an official W3C test suite for OWL~2 at
\url{http://owl.semanticweb.org/page/OWL_2_Test_Cases} (2011-02-09).
However, it does not cover OWL~2 Full sufficiently well, and
was not designed in a systematic way that allows easy determination of 
which parts of the language specification are not supported by a reasoner.}
The test suite was created specifically 
as a conformance test suite for OWL~2 Full and its main sub languages,
including RDFS and the OWL~2 RL/RDF rules.
The test suite consists of one or more test cases 
for each of the semantic conditions of the OWL~2 RDF-Based Semantics,
i.e., the test suite provides a systematic coverage of OWL~2 Full
at a specification level.
Most of the test cases are positive entailment and inconsistency tests,
but there are also a few
negative entailment tests and positive consistency tests.
The complete test suite consists of 736~test cases.
A large fraction of the test suite deals with datatype reasoning.
As the FOL axiomatization has almost no support for datatype reasoning,
only the test cases that cover the ``logical core'' of OWL~2 Full were used.
Further, only the positive entailment and inconsistency tests were used.
The resultant test suite has 411~test cases.

\subsubsection{OWL 2 Full-characteristic Test Cases.}
\label{toc:setting:testdata:fullish}

In order to investigate the extent of the reasoning possible using the
FOL axiomatization, a set of test cases that are characteristic
conclusions of OWL~2 Full was created.
``Characteristic'' means that the test cases represent OWL~2 Full reasoning
that cannot normally be expected
from either OWL~2 DL reasoning 
or from reasoners implementing the OWL~2 RL/RDF rules.
The test suite consists of 32~tests,
with 28~entailment tests
and 4~inconsistency tests.
There are test cases probing
semantic consequences 
from meta-modeling,
annotation properties,
the unrestricted use of complex properties,
and consequences from 
the use of OWL vocabulary terms as regular entities
(sometimes called ``syntax reflection'').

\subsubsection{Bulk RDF Data.}
\label{toc:setting:testdata:bulk}

For the scalability experiments,
a program that generates RDF graphs of arbitrary size
(``bulk RDF data'') was written.
The data consist of RDF triples using URIs that do
not conflict with the URIs in the test cases.
Further, no OWL vocabulary terms are used in the data sets.
This ensures that adding this bulk RDF data to test cases
does not affect the reasoning results.

\subsection{Reasoners}
\label{toc:setting:reasoners}

This section lists the different reasoning systems 
that were used in the experiments.
The idea behind the selection was to have a small number 
of representative systems for
(i) first-order proving,
(ii) first-order model finding, and
(iii) OWL reasoning.
Details of the ATP systems can be found on their web sites, and (for most)
in the system descriptions on the CASC-J5 web site.
The OWL reasoners were tested
to provide comparisons with existing 
state of the art Semantic Web reasoners.
Unless explicitly stated otherwise,
the systems were used in their default modes.

\subsubsection{Systems for first-order theorem proving}

\begin{itemize}

    \item\textsl{\textbf{Vampire~0.6}}
    (\url{http://www.vprover.org}).
    A powerful superposition-based ATP system,
    including strategy scheduling.
    \item\textsl{\textbf{Vampire-SInE~0.6}}
    A variant of Vampire that always runs the SInE strategy 
    (\url{http://www.cs.man.ac.uk/~hoderk/sine/desc/})
    to select axioms that are expected to be relevant.

	\item\textsl{\textbf{iProver-SInE~0.8}}
	(\url{http://www.cs.man.ac.uk/~korovink/iprover}).
	An instantiation-based ATP system, using
    the SInE strategy, and including strategy scheduling.


    
    

\end{itemize}

\subsubsection{Systems for first-order model finding}

\begin{itemize}
    \item\textsl{\textbf{Paradox~4.0}}
    (\url{http://www.cse.chalmers.se/~koen/code/}).
    A finite model finder, based on conversion to propositional form 
    and the use of a SAT solver.

    \item\textsl{\textbf{DarwinFM~1.4.5}}
    (\url{http://goedel.cs.uiowa.edu/Darwin}).
    A finite model finder, based on conversion to function-free first-order 
    logic and the use of the Darwin ATP system.
    
    
    
    
    
\end{itemize}

\subsubsection{Systems for OWL reasoning}

\begin{itemize}
	\item\textsl{\textbf{Pellet~2.2.2}}
	(\url{http://clarkparsia.com/pellet}).
	An OWL~2 DL reasoner 
	that implements a tableaux-based decision procedure.

	\item\textsl{\textbf{HermiT~1.3.2}}
	(\url{http://hermit-reasoner.com}).
	An OWL~2 DL reasoner 
	that implements a tableaux-based decision procedure.
	
    \item\textsl{\textbf{FaCT++~1.5.0}}
    (\url{http://owl.man.ac.uk/factplusplus}).
	An OWL~2 DL reasoner 
	that implements a tableaux-based decision procedure.
    
	\item\textsl{\textbf{BigOWLIM~3.4}}
	(\url{http://www.ontotext.com/owlim}).
	An RDF entailment-rule reasoner
	that comes with predefined rule sets.
	The OWL~2 RL/RDF rule set (\texttt{owl2-rl}) was used.
	The commercial ``BigOWLIM'' variant of the reasoning engine was applied,
	because it provides inconsistency checking.

	\item\textsl{\textbf{Jena~2.6.4}}
	(\url{http://jena.sourceforge.net}).
	A Java-based RDF framework
	that supports RDF entailment-rule reasoning
	and comes with predefined rule sets.
	The most expressive rule set,
	\texttt{OWL\_MEM\_RULE\_INF}, was used.

	\item\textsl{\textbf{Parliament~2.6.9}}
	(\url{http://parliament.semwebcentral.org}).
	An RDF triple store
	with some limited OWL reasoning capabilities.
	Parliament cannot detect inconsistencies in ontologies.

\end{itemize}

\subsection{Evaluation Environment}
\label{toc:setting:environment}

Testing was done on computers with a 2.8GHz Intel 
Pentium~4 CPU, 2GB~memory, running Linux~FC8.
A 300s CPU time limit was imposed on each run.

\section{Evaluation Results}
\label{toc:results}
This section presents the results of the following reasoning experiments:
a \emph{language coverage analysis}, to determine the degree of conformance 
to the language specification of OWL~2 Full;
\emph{``characteristic'' OWL 2 Full reasoning} experiments
to determine the extent 
to which distinguishing OWL~2 Full reasoning is possible;
some basic \emph{scalability testing};
and several \emph{model finding experiments}
to determine whether first-order model finders can be used in practice
for the recognition of non-entailments and consistent ontologies.
The following markers are used in the result tables
to indicate the outcomes of the experiments:
\begin{itemize}
  \item
  \textbf{\emph{success (`\CellTextSuccess')}:}
  a test run that provided the correct result.
  \item
  \textbf{\emph{wrong (`\CellTextWrong')}:}
  a test run that provided a wrong result, e.g., when a reasoner
  claims that an entailment test case is a non-entailment.
  \item
  \textbf{\emph{unknown (`\CellTextUnknown')}:}
  a test run that did not provide a result, e.g.,
  due to a processing error or time out.
\end{itemize}

This section also presents comparative evaluation results
for the OWL reasoners listed in Section~\ref{toc:setting:reasoners}.
This illustrates the degree to which OWL~2 Full reasoning
can already be achieved with existing OWL reasoners,
and the added value of our reasoning approach 
compared to existing Semantic Web technology. 
This means, for example, that an OWL~2 DL reasoner will produce a wrong result
if it classifies an OWL~2 Full entailment test case as a non-entailment.
However, this negative evaluation result
refers to only the level of conformance with respect to OWL~2 Full reasoning,
i.e., the reasoner may still be a compliant implementation of OWL~2 DL.

\subsection{Language Coverage}
\label{toc:results:coverage}

This experiment used the FOL axiomatization with the 411~test cases in
the language coverage suite described in 
Section~\ref{toc:setting:testdata:coverage}.
The results of the experiment
are shown in Table~\ref{tab:results-coverage-atp-complete}.
iProver-SInE succeeded on~93\% of the test cases, and 
Vampire succeeded on 85\%.
%
It needs to be mentioned that the results were not perfectly stable.
Over several runs the number of successes varied for iProver-SInE 
between 382 and~386.
This is caused by small variations in the timing of strategy
changes within iProver-SInE's strategy scheduling.

\begin{table}[b!]
	\begin{center}
\begin{tabular}{|l||r|r|r|}
\hline
\textbf{Reasoner} & 
\textbf{Success} & 
\textbf{Wrong} & 
\textbf{Unknown} \\ 
\hline
\hline
Vampire
& 349 & 0 & 62 \\
\hline
iProver-SInE
& 383 & 0 & 28 \\
\hline
\end{tabular}

	\end{center}
	\caption{Language coverage: ATPs with OWL~2 Full axiom set.}
	\label{tab:results-coverage-atp-complete}
\end{table}

\begin{figure}[tb!]
	\begin{center}
	\includegraphics[width=0.75\textwidth]{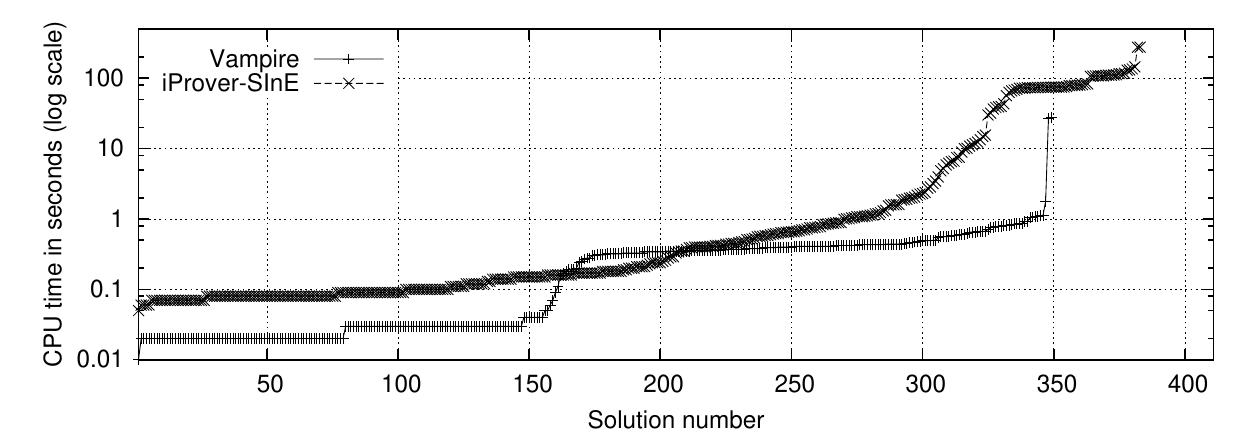}
	\caption{Language coverage: ordered system times of ATPs.}
	\label{fig:results-coverage-orderedtimes}
	\end{center}
\end{figure}

%
Figure~\ref{fig:results-coverage-orderedtimes} 
shows the runtime behavior of the two systems, with the times 
for successes sorted into increasing order.
Both systems take less than 1s for the majority of their successes.
Although Vampire succeeded on less cases than iProver-SInE, it
is typically faster in the case of a success.

%
An analysis of the 28~test cases for which both Vampire and iProver-SInE 
did not succeed revealed that 14 of them require 
support for OWL~2 Full language features
not covered by the FOL axiomatization,
including certain forms of datatype reasoning
and support for the RDF container vocabulary~\cite{w3c04-rdf-semantics}.
A future version of the axiomatization will encode these parts
of the OWL~2 Full semantics,
which might lead to improved results. 
%
For each of the remaining 14~test cases, 
subsets of axioms sufficient for a solution 
were hand-selected from the FOL axiomatization.
These axiom sets were generally very small,
with up to 16~axioms,
and in most cases less than 10~axioms.
iProver-SInE succeeded on 13 of these 14~test cases.
The remaining test case is a considerably complex one,
involving the semantics of qualified cardinality restrictions.
It was solved by Vampire.
Thus, all test cases were solved except for the 14 
that are beyond the current axiomatization.

%
For comparison, the OWL reasoners
listed in Section~\ref{toc:setting:reasoners} 
were also tested.
The results are shown in Table~\ref{tab:results-coverage-semweb}.
The OWL~2 DL reasoners Pellet and HermiT 
both succeeded on about 60\% of the test cases.
A comparison of the individual results showed
that the two reasoners succeeded mostly on the same test cases.
Interestingly, 
although most of the test cases are formally invalid OWL~2 DL ontologies,
reasoning rarely resulted in a processing error.
Rather, in ca.~40\% of the cases,
the reasoners wrongly reported a test case 
to be a non-entailment or a consistent ontology.
The third OWL~2 DL reasoner, FaCT++, signaled a processing error more often,
and succeeded on less than 50\% of the test cases.
 
The OWL~2 RL/RDF rule reasoner BigOWLIM 
succeeded on roughly 70\% of the test cases.
Although the number of successful tests
was larger than for all the OWL~2 DL reasoners,
there was a considerable number of test cases 
for which the OWL~2 DL reasoners were successful
but not BigOWLIM, 
and vice versa.
The Jena OWL reasoner, 
which is an RDF entailment rule reasoner like BigOWLIM,
succeeded on about only 30\% of the test cases,
which is largely due to missing support for OWL~2 features. 
Finally, 
Parliament succeeded on only 14 of the test cases.
In particular, it did not solve any of the inconsistency test cases.
The low success rate reflects the style of ``light-weight reasoning''
used in many reasoning-enabled RDF triple stores.

\begin{table}[tb!]
	\begin{center}
	\begin{tabular}{|l||r|r|r|}
\hline
\textbf{Reasoner} & 
\textbf{Success} & 
\textbf{Wrong} & 
\textbf{Unknown} \\ 
\hline
\hline
Pellet
& 237 & 168 & 6 \\
\hline
HermiT
& 246 & 157 & 8 \\
\hline
FaCT++
& 190 & 45 & 176 \\
\hline
BigOWLIM
& 282 & 129 & 0 \\
\hline
Jena
& 129 & 282 & 0 \\
\hline
Parliament
& 14 & 373 & 24 \\
\hline
\end{tabular}

	\end{center}
	\caption{Language coverage: OWL reasoners.}
	\label{tab:results-coverage-semweb}
\end{table}

\subsection{Characteristic OWL 2 Full Conclusions}
\label{toc:results:fullish}


The test suite of characteristic OWL~2 Full conclusions
focuses on semantic consequences
that are typically beyond the scope of OWL~2 DL 
or RDF rule reasoners.
This is reflected in Table~\ref{tab:results-fullish-semweb},
which presents the results for the OWL reasoning systems.
The column numbers correspond to the test case numbers in the test suite.
In general, the OWL reasoners show significantly weaker performance
on this test suite than on the language coverage test suite.
Note that the successful test cases 
for the OWL~2 DL reasoners 
(Pellet, HermiT and FaCT++)
have only little overlap with the successful test cases
for the RDF rule reasoners
(BigOWLIM and Jena).
Parliament succeeded on only two test cases.

\begin{table}[tb!]
	\begin{tiny}
	\begin{center}
	\begin{tabular}{|l||c|c|c|c|c|c|c|c|c|c|c|c|c|c|c|c|c|c|c|c|c|c|c|c|c|c|c|c|c|c|c|c|}
\hline
& 01 & 02 & 03 & 04 & 05 & 06 & 07 & 08 & 09 & 10 & 11 & 12 & 13 & 14 & 15 & 16 & 17 & 18 & 19 & 20 & 21 & 22 & 23 & 24 & 25 & 26 & 27 & 28 & 29 & 30 & 31 & 32 \\
\hline
\hline
PE
&\CellSuccess&\CellSuccess&\CellSuccess&\CellWrong&\CellWrong&\CellWrong&\CellWrong&\CellWrong&\CellSuccess&\CellSuccess&\CellWrong&\CellWrong&\CellWrong&\CellWrong&\CellSuccess&\CellWrong&\CellWrong&\CellWrong&\CellWrong&\CellSuccess&\CellSuccess&\CellWrong&\CellWrong&\CellWrong&\CellWrong&\CellSuccess&\CellWrong&\CellWrong&\CellUnknown&\CellWrong&\CellWrong&\CellWrong\\
\hline
HE
&\CellSuccess&\CellUnknown&\CellSuccess&\CellWrong&\CellWrong&\CellUnknown&\CellWrong&\CellSuccess&\CellSuccess&\CellSuccess&\CellWrong&\CellWrong&\CellWrong&\CellWrong&\CellSuccess&\CellWrong&\CellWrong&\CellWrong&\CellWrong&\CellSuccess&\CellSuccess&\CellWrong&\CellWrong&\CellSuccess&\CellUnknown&\CellSuccess&\CellWrong&\CellWrong&\CellUnknown&\CellWrong&\CellWrong&\CellWrong\\
\hline
FA
&\CellSuccess&\CellUnknown&\CellUnknown&\CellUnknown&\CellUnknown&\CellUnknown&\CellUnknown&\CellWrong&\CellUnknown&\CellSuccess&\CellWrong&\CellWrong&\CellWrong&\CellUnknown&\CellSuccess&\CellUnknown&\CellWrong&\CellWrong&\CellWrong&\CellSuccess&\CellSuccess&\CellUnknown&\CellUnknown&\CellUnknown&\CellUnknown&\CellSuccess&\CellWrong&\CellUnknown&\CellUnknown&\CellWrong&\CellWrong&\CellUnknown\\
\hline
BO
&\CellSuccess&\CellWrong&\CellWrong&\CellSuccess&\CellWrong&\CellWrong&\CellSuccess&\CellSuccess&\CellWrong&\CellWrong&\CellSuccess&\CellSuccess&\CellWrong&\CellWrong&\CellSuccess&\CellWrong&\CellWrong&\CellSuccess&\CellSuccess&\CellWrong&\CellWrong&\CellWrong&\CellWrong&\CellWrong&\CellWrong&\CellWrong&\CellWrong&\CellWrong&\CellWrong&\CellWrong&\CellWrong&\CellWrong\\
\hline
JE
&\CellSuccess&\CellWrong&\CellWrong&\CellWrong&\CellWrong&\CellSuccess&\CellSuccess&\CellSuccess&\CellWrong&\CellWrong&\CellSuccess&\CellWrong&\CellWrong&\CellWrong&\CellWrong&\CellWrong&\CellWrong&\CellSuccess&\CellWrong&\CellWrong&\CellWrong&\CellWrong&\CellSuccess&\CellWrong&\CellWrong&\CellSuccess&\CellWrong&\CellWrong&\CellWrong&\CellWrong&\CellWrong&\CellSuccess\\
\hline
PA
&\CellSuccess&\CellWrong&\CellWrong&\CellWrong&\CellWrong&\CellWrong&\CellWrong&\CellSuccess&\CellWrong&\CellWrong&\CellUnknown&\CellWrong&\CellWrong&\CellWrong&\CellWrong&\CellWrong&\CellWrong&\CellWrong&\CellUnknown&\CellWrong&\CellWrong&\CellWrong&\CellWrong&\CellWrong&\CellWrong&\CellWrong&\CellWrong&\CellWrong&\CellWrong&\CellUnknown&\CellUnknown&\CellWrong\\
\hline
\end{tabular}

	\end{center}
	\end{tiny}
	\caption{Characteristic conclusions: OWL reasoners.
\mbox{PE=Pellet},
\mbox{HE=HermiT},
\mbox{FA=FaCT++},
\mbox{BO=BigOWLIM},
\mbox{JE=Jena},
\mbox{PA=Parliament}.%
}
	\label{tab:results-fullish-semweb}
\end{table}

%
The first two rows of Table~\ref{tab:results-fullish-atp} show that
the ATP systems achieved much better results than the OWL reasoners, 
using the complete OWL~2 Full axiomatization.
iProver-SInE succeeded on 28 of the 32~test cases, and
Vampire succeeded on 23.
As was done for the language coverage test cases, small subsets of axioms 
sufficient for each of the test cases were hand-selected from the FOL 
axiomatization.
As the last two rows of Table~\ref{tab:results-fullish-atp} show,
both ATP systems succeeded on all these simpler test cases.

\begin{table}
	\begin{tiny}
	\begin{center}
	\begin{tabular}{|l||c|c|c|c|c|c|c|c|c|c|c|c|c|c|c|c|c|c|c|c|c|c|c|c|c|c|c|c|c|c|c|c|}
\hline
& 01 & 02 & 03 & 04 & 05 & 06 & 07 & 08 & 09 & 10 & 11 & 12 & 13 & 14 & 15 & 16 & 17 & 18 & 19 & 20 & 21 & 22 & 23 & 24 & 25 & 26 & 27 & 28 & 29 & 30 & 31 & 32 \\
\hline
\hline
VA/c
&\CellSuccess&\CellSuccess&\CellSuccess&\CellSuccess&\CellSuccess&\CellSuccess&\CellSuccess&\CellSuccess&\CellSuccess&\CellUnknown&\CellSuccess&\CellUnknown&\CellUnknown&\CellSuccess&\CellSuccess&\CellSuccess&\CellSuccess&\CellSuccess&\CellSuccess&\CellUnknown&\CellUnknown&\CellUnknown&\CellSuccess&\CellSuccess&\CellUnknown&\CellSuccess&\CellUnknown&\CellUnknown&\CellSuccess&\CellSuccess&\CellSuccess&\CellSuccess\\
\hline
IS/c
&\CellSuccess&\CellSuccess&\CellSuccess&\CellSuccess&\CellSuccess&\CellSuccess&\CellSuccess&\CellSuccess&\CellSuccess&\CellSuccess&\CellSuccess&\CellUnknown&\CellUnknown&\CellSuccess&\CellSuccess&\CellSuccess&\CellSuccess&\CellSuccess&\CellSuccess&\CellUnknown&\CellUnknown&\CellSuccess&\CellSuccess&\CellSuccess&\CellSuccess&\CellSuccess&\CellSuccess&\CellSuccess&\CellSuccess&\CellSuccess&\CellSuccess&\CellSuccess\\
\hline
VA/s
&\CellSuccess&\CellSuccess&\CellSuccess&\CellSuccess&\CellSuccess&\CellSuccess&\CellSuccess&\CellSuccess&\CellSuccess&\CellSuccess&\CellSuccess&\CellSuccess&\CellSuccess&\CellSuccess&\CellSuccess&\CellSuccess&\CellSuccess&\CellSuccess&\CellSuccess&\CellSuccess&\CellSuccess&\CellSuccess&\CellSuccess&\CellSuccess&\CellSuccess&\CellSuccess&\CellSuccess&\CellSuccess&\CellSuccess&\CellSuccess&\CellSuccess&\CellSuccess\\
\hline
IS/s
&\CellSuccess&\CellSuccess&\CellSuccess&\CellSuccess&\CellSuccess&\CellSuccess&\CellSuccess&\CellSuccess&\CellSuccess&\CellSuccess&\CellSuccess&\CellSuccess&\CellSuccess&\CellSuccess&\CellSuccess&\CellSuccess&\CellSuccess&\CellSuccess&\CellSuccess&\CellSuccess&\CellSuccess&\CellSuccess&\CellSuccess&\CellSuccess&\CellSuccess&\CellSuccess&\CellSuccess&\CellSuccess&\CellSuccess&\CellSuccess&\CellSuccess&\CellSuccess\\
\hline
\end{tabular}

	\end{center}
	\end{tiny}
	\caption{
Characteristic conclusions: ATPs with complete and small axiom sets.
\mbox{VA/c=Vampire/complete},
\mbox{IS/c=iProver-SInE/complete},
\mbox{VA/s=Vampire/small},
\mbox{IS/s=iProver-SInE/small}.%
}
	\label{tab:results-fullish-atp}
\end{table}

\begin{figure}[tb!]
	\begin{center}
	\includegraphics[width=0.75\textwidth]{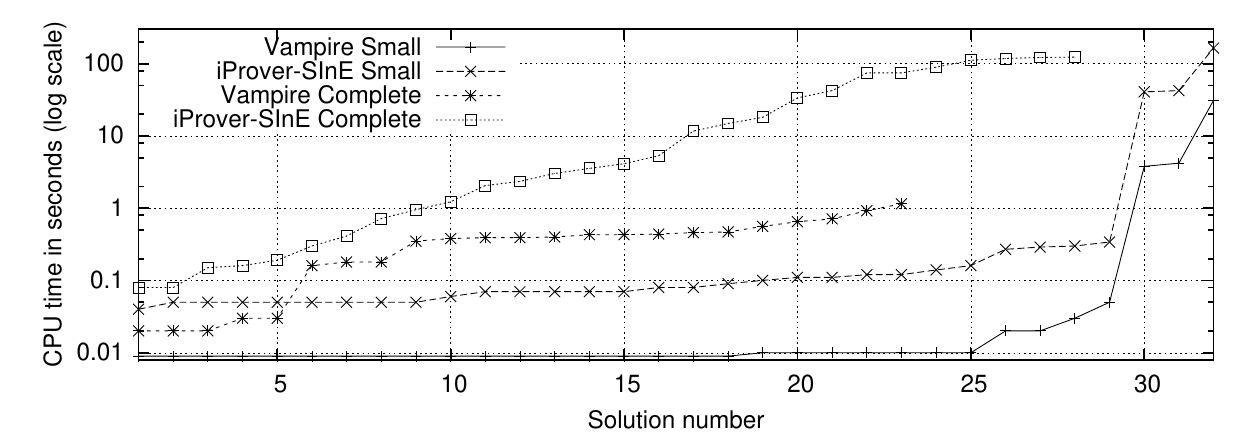}
	\caption{Characteristic conclusions: ordered system times of ATPs.}
	\label{tab:results-fullish-orderedtimes}
	\end{center}
\end{figure}

Figure~\ref{tab:results-fullish-orderedtimes} 
shows the runtime behavior of the two systems.
For the complete axiomatization, Vampire either succeeds in less than 1s
or does not succeed.
In contrast, iProver's performance degrades more gracefully.
The reasoning times using the small-sufficient axiom sets are generally 
up to several magnitudes lower than for the complete axiomatization.
In the majority of cases they are below~1s.

\subsection{Scalability}
\label{toc:results:scalability}

The Semantic Web consists of huge data masses, but single reasoning results 
presumably 
often depend on only a small fraction of that data.
As a basic test of the ATP systems' abilities to ignore irrelevant background
axioms, a set of one million ``bulk RDF axioms'' (as described in
Section~\ref{toc:setting:testdata}) was added 
to the test cases of characteristic OWL~2 Full conclusions.
This was done using the complete FOL axiomatization, 
and also the small-sufficient
sets of axioms for each test case.

Table~\ref{tab:results-bulk-atp} shows the results.
The default version of Vampire produced very poor results, as is shown in 
the first and fourth rows of the table.
(Strangely, Vampire had two more successes with the complete axiomatization
than with the small-sufficient axiom sets.
That can be attributed to differences in the strategies selected
for the different axiomatizations.)
In contrast, as shown in the second, third, fifth and sixth rows,
the version of Vampire-SInE and iProver-SInE did much better. 
The use of the SInE strategy for selecting relevant axioms clearly helps.

\begin{table}[tb!]
	\begin{tiny}
	\begin{center}
	   \begin{tabular}{|l||c|c|c|c|c|c|c|c|c|c|c|c|c|c|c|c|c|c|c|c|c|c|c|c|c|c|c|c|c|c|c|c|}
   \hline
   & 01 & 02 & 03 & 04 & 05 & 06 & 07 & 08 & 09 & 10 & 11 & 12 & 13 & 14 & 15 & 16 & 17 & 18 & 19 & 20 & 21 & 22 & 23 & 24 & 25 & 26 & 27 & 28 & 29 & 30 & 31 & 32 \\
   \hline
   \hline
   VA/c
   &\CellSuccess&\CellSuccess&\CellSuccess&\CellUnknown&\CellUnknown&\CellUnknown&\CellUnknown&\CellUnknown&\CellUnknown&\CellUnknown&\CellUnknown&\CellUnknown&\CellUnknown&\CellUnknown&\CellSuccess&\CellUnknown&\CellUnknown&\CellUnknown&\CellUnknown&\CellUnknown&\CellUnknown&\CellUnknown&\CellUnknown&\CellUnknown&\CellUnknown&\CellUnknown&\CellUnknown&\CellUnknown&\CellUnknown&\CellUnknown&\CellUnknown&\CellUnknown\\
   \hline
   VS/c
   &\CellSuccess&\CellSuccess&\CellSuccess&\CellSuccess&\CellSuccess&\CellSuccess&\CellUnknown&\CellSuccess&\CellUnknown&\CellUnknown&\CellSuccess&\CellUnknown&\CellUnknown&\CellUnknown&\CellSuccess&\CellSuccess&\CellUnknown&\CellSuccess&\CellSuccess&\CellUnknown&\CellUnknown&\CellUnknown&\CellSuccess&\CellUnknown&\CellUnknown&\CellSuccess&\CellUnknown&\CellUnknown&\CellUnknown&\CellSuccess&\CellUnknown&\CellSuccess\\
   \hline
   IS/c
   &\CellSuccess&\CellSuccess&\CellSuccess&\CellSuccess&\CellSuccess&\CellSuccess&\CellSuccess&\CellSuccess&\CellSuccess&\CellSuccess&\CellSuccess&\CellUnknown&\CellUnknown&\CellSuccess&\CellSuccess&\CellSuccess&\CellSuccess&\CellSuccess&\CellSuccess&\CellUnknown&\CellUnknown&\CellSuccess&\CellSuccess&\CellSuccess&\CellSuccess&\CellSuccess&\CellSuccess&\CellSuccess&\CellSuccess&\CellSuccess&\CellSuccess&\CellSuccess\\
   \hline
   VA/s
   &\CellSuccess&\CellUnknown&\CellSuccess&\CellUnknown&\CellUnknown&\CellUnknown&\CellUnknown&\CellUnknown&\CellUnknown&\CellUnknown&\CellUnknown&\CellUnknown&\CellUnknown&\CellUnknown&\CellUnknown&\CellUnknown&\CellUnknown&\CellUnknown&\CellUnknown&\CellUnknown&\CellUnknown&\CellUnknown&\CellUnknown&\CellUnknown&\CellUnknown&\CellUnknown&\CellUnknown&\CellUnknown&\CellUnknown&\CellUnknown&\CellUnknown&\CellUnknown\\
   \hline
   VS/s
   &\CellSuccess&\CellSuccess&\CellSuccess&\CellSuccess&\CellSuccess&\CellSuccess&\CellUnknown&\CellSuccess&\CellSuccess&\CellSuccess&\CellSuccess&\CellSuccess&\CellUnknown&\CellSuccess&\CellSuccess&\CellSuccess&\CellUnknown&\CellSuccess&\CellSuccess&\CellUnknown&\CellUnknown&\CellSuccess&\CellSuccess&\CellSuccess&\CellSuccess&\CellSuccess&\CellSuccess&\CellUnknown&\CellSuccess&\CellSuccess&\CellUnknown&\CellSuccess\\
   \hline
   IS/s
   &\CellSuccess&\CellSuccess&\CellSuccess&\CellSuccess&\CellSuccess&\CellSuccess&\CellSuccess&\CellSuccess&\CellSuccess&\CellSuccess&\CellSuccess&\CellSuccess&\CellUnknown&\CellSuccess&\CellSuccess&\CellSuccess&\CellSuccess&\CellSuccess&\CellSuccess&\CellSuccess&\CellSuccess&\CellSuccess&\CellSuccess&\CellSuccess&\CellSuccess&\CellSuccess&\CellSuccess&\CellSuccess&\CellSuccess&\CellSuccess&\CellSuccess&\CellSuccess\\
   \hline
   \end{tabular}

	\end{center}
	\end{tiny}
	\caption{Scalability: ATPs with complete and small axiom sets, 1M~RDF triples.
\mbox{VA/c=Vampire/complete},
\mbox{VS/c=Vampire-SInE/complete},
\mbox{IS/c=iProver-SInE/complete},
\mbox{VA/s=Vampire/small},
\mbox{VS/s=Vampire-SInE/small},
\mbox{IS/s=iProver-SInE/small}.%
}
	\label{tab:results-bulk-atp}
\end{table}

\begin{figure}[tb!]
	\begin{center}
	\includegraphics[width=0.75\textwidth]{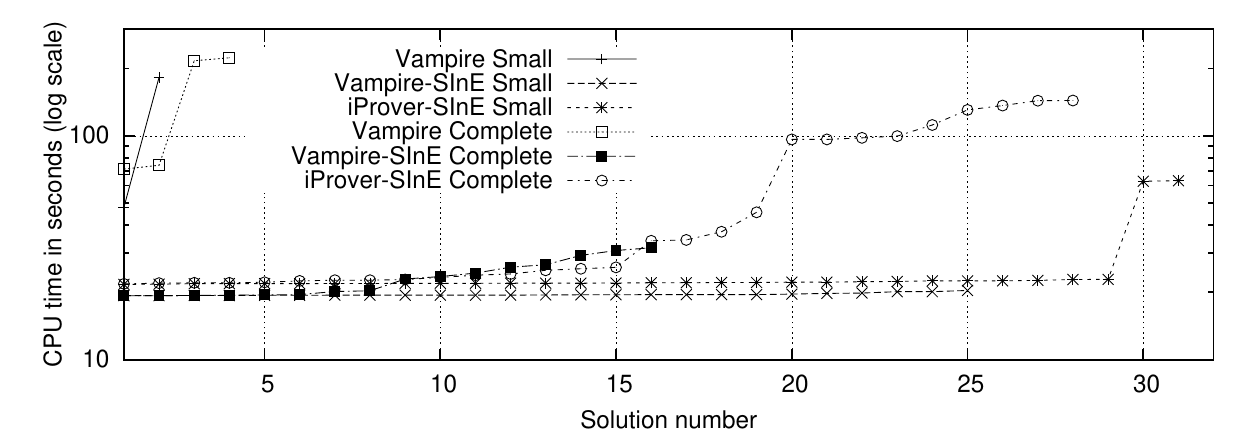}
	\caption{Scalability: ordered system times of ATPs, 1M~RDF triples.}
	\label{fig:results-bulk-orderedtimes}
	\end{center}
\end{figure}

Figure~\ref{fig:results-bulk-orderedtimes} shows the runtime behavior
of the systems.
The bulk axioms evidently add a constant overhead of about 20s to all
successes, which is believed to be taken parsing the large files.
In an application setting this might be done only once at the start,
so that the time would be amortized over multiple reasoning tasks.
The step in iProver's performance at the 20th problem is an
artifact of strategy scheduling.

The bulk axioms were designed to have no connection to the FOL axiomatization
or the RDF graphs.
As such, simple analysis of inference chains
from the conjecture~\cite{PY03} would be sufficient to determine that the 
bulk axioms could not be used in a solution.
This simplistic approach is methodologically an appropriate way to start
testing robustness against irrelevant axioms, and potentially not too
far off the reality of Semantic Web reasoning.
However, future work using axioms that are not so obviously redundant would 
properly exercise the power of the SInE approach to axiom selection.

\subsection{Model Finding}
\label{toc:results:saturation}

This section presents the results from experiments
concerning the detection of non-entailments and consistent ontologies 
w.r.t.\ OWL~2 Full and two of its sub languages:
ALCO Full~\cite{motik07-metamodeling} 
and RDFS~\cite{w3c04-rdf-semantics}.
ALCO Full is interesting because it is a small fragment of OWL~2 Full
that is known to be undecidable~\cite{motik07-metamodeling}.
RDFS is interesting because it is a minimally meaningful language
that shares the main characteristics of OWL~2 Full.
The RDFS axioms included the  ``extensional'' semantic extension,
as non-normatively defined in Section~4.2 of~\cite{w3c04-rdf-semantics}.
Similarly, the original definition of ALCO Full was extended to include
extensional RDFS.
%
%
No report is given for the OWL reasoners,
as only the OWL~2 DL reasoners have model-finding capabilities,
and not for any of the three languages considered here.

Consistency checking for an RDF graph~$G$ w.r.t.\ some ontology language~$L$
corresponds to consistency checking for the combination
of a complete axiomatization of~$L$ and the FOL translation of~$G$.
Hence, a minimum requirement is to confirm that the FOL axiomatization of 
OWL~2 Full is consistent.
Unfortunately, for the OWL 2 Full axiomatization 
no model finder was able to confirm consistency.\footnote{%
This raised the question of whether our positive entailment reasoning results 
were perhaps due to an inconsistent axiomatization.
However, none of the theorem provers was able to establish inconsistency.
In addition, the model finders confirmed the consistency of all the 
small-sufficient axiom sets mentioned in 
Section~\ref{toc:results:fullish}.
Hence, it is at least ensured that those positive reasoning results
are achievable from consistent subsets of the OWL~2 Full axiomatization.
}

For the ALCO Full axioms, Paradox found a finite model of size~5 in ca.~5s 
CPU time, while DarwinFM timed out.
Paradox was then used on the characteristic OWL~2 Full test cases, with the 
OWL~2 Full axiomatization replaced by the ALCO Full axioms. 
As ALCO Full is a sub language of OWL~2 Full,
24 of the 32 test cases are either 
non-entailments or consistent ontologies,
out of which 15~were correctly recognized by Paradox. 
iProver-SInE was used to confirm that the
remaining 8~test cases are positive entailments or inconsistent ontologies.
The results are shown in the first row of Table~\ref{tab:results-sat-atp}.

%
For the RDFS axioms, analogous experiments were done.
Paradox found a finite model of the axioms, of size~1, in about 1s. 
The consistency was confirmed by DarwinFM in less than 1s.
With the OWL~2 Full axiomatization replaced by the RDFS axioms, 29 of the 
32~characteristic test cases are non-entailments or consistent ontologies.
Paradox and Darwin confirmed all of these, mostly in ca.~1s, with a maximum 
time of ca.~2s.
iProver-SInE confirmed that the remaining 3~test cases are positive 
entailments or inconsistent ontologies.
These results are shown in the second and third rows of 
Table~\ref{tab:results-sat-atp}.

\begin{table}[tb!]
	\begin{tiny}
	\begin{center}
	   \begin{tabular}{|l||c|c|c|c|c|c|c|c|c|c|c|c|c|c|c|c|c|c|c|c|c|c|c|c|c|c|c|c|c|c|c|c|}
   \hline
   & 01 & 02 & 03 & 04 & 05 & 06 & 07 & 08 & 09 & 10 & 11 & 12 & 13 & 14 & 15 & 16 & 17 & 18 & 19 & 20 & 21 & 22 & 23 & 24 & 25 & 26 & 27 & 28 & 29 & 30 & 31 & 32 \\
   \hline
   \hline
   PA/A
   &\CellEmpty&\CellEmpty&\CellEmpty&\CellSuccess&\CellSuccess&\CellSuccess&\CellSuccess&\CellSuccess&\CellEmpty&\CellEmpty&\CellSuccess&\CellEmpty&\CellUnknown&\CellEmpty&\CellSuccess&\CellSuccess&\CellSuccess&\CellSuccess&\CellSuccess&\CellUnknown&\CellUnknown&\CellUnknown&\CellUnknown&\CellSuccess&\CellUnknown&\CellUnknown&\CellUnknown&\CellSuccess&\CellEmpty&\CellSuccess&\CellUnknown&\CellSuccess\\
   \hline
   PA/R
   &\CellEmpty&\CellEmpty&\CellEmpty&\CellSuccess&\CellSuccess&\CellSuccess&\CellSuccess&\CellSuccess&\CellSuccess&\CellSuccess&\CellSuccess&\CellSuccess&\CellSuccess&\CellSuccess&\CellSuccess&\CellSuccess&\CellSuccess&\CellSuccess&\CellSuccess&\CellSuccess&\CellSuccess&\CellSuccess&\CellSuccess&\CellSuccess&\CellSuccess&\CellSuccess&\CellSuccess&\CellSuccess&\CellSuccess&\CellSuccess&\CellSuccess&\CellSuccess\\
   \hline
   DF/R
   &\CellEmpty&\CellEmpty&\CellEmpty&\CellSuccess&\CellSuccess&\CellSuccess&\CellSuccess&\CellSuccess&\CellSuccess&\CellSuccess&\CellSuccess&\CellSuccess&\CellSuccess&\CellSuccess&\CellSuccess&\CellSuccess&\CellSuccess&\CellSuccess&\CellSuccess&\CellSuccess&\CellSuccess&\CellSuccess&\CellSuccess&\CellSuccess&\CellSuccess&\CellSuccess&\CellSuccess&\CellSuccess&\CellSuccess&\CellSuccess&\CellSuccess&\CellSuccess\\
   \hline
   \end{tabular}

	\end{center}
	\end{tiny}
	\caption{Model Finding: ATPs with ALCO Full and RDFS axiom sets.
The black entries indicate positive entailments or inconsistent ontologies.
\mbox{PA/A=Paradox/ALCO Full},
\mbox{PA/R=Paradox/RDFS},
\mbox{DF/R=DarwinFM/RDFS}.
}
	\label{tab:results-sat-atp}
\end{table}

An interesting observation made during the model finding experiments was
that \emph{finite} model finders were effective, e.g., the results of 
Paradox and DarwinFM above.
In contrast, other model finders such as iProver-SAT (a variant of iProver
tuned for model finding) and Darwin (the plain Model Evolution core of
DarwinFM) were less effective, e.g., taking 80s and 37s respectively to
confirm the satisfiability of the RDFS axiom set.

\section{Conclusions and Future Work}
\label{toc:conclusions}

This paper has described how first order ATP systems
can be used for reasoning in the OWL~2 Full ontology language,
using a straight-forward translation 
of the underlying model theory into a FOL axiomatization.
The results were obtained from two complementary test suites,
one for language coverage analysis
and one for probing characteristic conclusions of OWL~2 Full.
The results indicate that this approach can be applied in practice 
for effective OWL reasoning, and offers a viable alternative
to current Semantic Web reasoners.
Some scalability testing was done by adding large sets of 
semantically unrelated RDF data to the test case data.
While the ATP systems that include the SInE strategy effectively 
ignored this redundant data, 
it was surprising that other ATP systems
did not use simple reachability analysis to detect and ignore this bulk data --
this suggests an easy way for developers 
to adapt their systems to such problems.

In contrast to the successes of the ATP systems proving theorems, model
finders were less successful in identifying non-entailments and 
consistent ontologies w.r.t.\ OWL~2 Full.
However, some successes were obtained for ALCO Full.
Since ALCO Full is an undecidable sub-language of OWL~2 Full,
there is hope that the failures were not due to undecidability
but rather due to the large number of axioms.
This needs to be investigated further.
Model finding for RDFS worked quite efficiently, which
is interesting because we do not know of any tool 
that detects RDFS non-entailments.

In the future we plan to extend the approach to datatype reasoning,
which is of high practical relevance in the Semantic Web.
It may be possible to take advantage of the typed first-order or typed
higher-order form of the TPTP language to effectively encode the datatypes,
and reason using ATP systems that take advantage of the type information.
Another topic for further research is to develop techniques
for identifying parts of the FOL axiomatization that are relevant to a given
reasoning task.
It is hoped that by taking into account OWL~2 Full specific knowledge,
more precise axiom selection than offered by the generic SInE approach
will be possible.
An important area of development will be query answering,
i.e., the ability to obtain explicit answers to users' questions.
For future OWL~2 Full reasoners
this will be a very relevant reasoning task, particularly with respect to
the current extension of the standard RDF query language SPARQL
towards ``entailment regimes''
(\url{http://www.w3.org/TR/sparql11-entailment}). 
This topic is also of growing interest in the ATP community, 
with a proposal being
considered for expressing questions and answers in the TPTP language
(\url{http://www.tptp.org/TPTP/Proposals/AnswerExtraction.html}).

\bibliography{biblio}
\bibliographystyle{splncs03}

\IfExtendedPaperThenElse{

\newpage

\appendix

\section{Detailed Raw Result Data}
\label{toc:detailedresults}
This appendix provides detailed raw result data 
that underlies the experimental results 
reported in Section~\ref{toc:results}.
The data is given in tables, 
which present the names of the test cases in the first column, 
and the results of individual reasoning experiments in the other columns.
The result of each experiment consists of 
one of the possible outcomes 
defined at the beginning of Section~\ref{toc:results}
and, optionally, the duration of the reasoning experiment,
given as the number of seconds it took.
All result data presented here is also available in electronic form
as part of the \emph{supplementary material} for this paper
(see the download link at the beginning of Section~\ref{toc:setting}).

\subsection{Language Coverage Results}
\label{toc:detailedresults:coverage}

The following tables provide the raw result data 
that underlies the aggregated results 
for the language coverage experiments,
as reported in Section~\ref{toc:results:coverage}.
Table~\ref{tab:detailedresults-coverage-allreasoners}
contains the result data that was obtained from evaluating
the Semantic Web reasoners 
and from the FOL theorem provers 
when used with the complete OWL~2 Full axiomatization;
the corresponding aggregated results were reported in Tables
\ref{tab:results-coverage-semweb} 
and
\ref{tab:results-coverage-atp-complete},
respectively.
The remaining tables provide the raw results 
from testing the FOL reasoners
on those test cases where they had failed originally,
now using small but sufficient subaxiomatizations
that were manually crafted for each of the test cases.

\begin{center}
\begin{tiny}
\begin{longtable}{|l||c|c|c|c|c|c|c|c|}
\caption{%
Result data of the language coverage experiments for
Semantic Web reasoners
and for FOL theorem provers
when used with the complete OWL~2 Full axiomatization.
Time values have only be measured for the FOL reasoners.
\mbox{PE=Pellet},
\mbox{HE=HermiT},
\mbox{FA=FaCT++},
\mbox{BO=BigOWLIM},
\mbox{JE=Jena},
\mbox{PA=Parliament},
\mbox{VA=Vampire},
\mbox{IS=iProver-SInE}%
}
\label{tab:detailedresults-coverage-allreasoners}
\\
\hline
\textbf{Test Case} & \textbf{PE} & \textbf{HE} & \textbf{FA} & \textbf{BO} & \textbf{JE} & \textbf{PA} & \textbf{VA} & \textbf{IS} \\
\hline
\hline
rdfbased-sem-bool-complement-data & -\CellColorWrong & -\CellColorWrong & ?\CellColorUnknown & -\CellColorWrong & -\CellColorWrong & -\CellColorWrong & + (0.07)\CellColorSuccess & + (0.16)\CellColorSuccess \\
\hline
rdfbased-sem-bool-complement-ext & +\CellColorSuccess & +\CellColorSuccess & +\CellColorSuccess & -\CellColorWrong & -\CellColorWrong & -\CellColorWrong & + (0.33)\CellColorSuccess & + (0.61)\CellColorSuccess \\
\hline
rdfbased-sem-bool-complement-inst & +\CellColorSuccess & +\CellColorSuccess & +\CellColorSuccess & +\CellColorSuccess & -\CellColorWrong & ?\CellColorUnknown & + (0.28)\CellColorSuccess & + (0.29)\CellColorSuccess \\
\hline
rdfbased-sem-bool-demorgan & +\CellColorSuccess & +\CellColorSuccess & +\CellColorSuccess & -\CellColorWrong & -\CellColorWrong & -\CellColorWrong & ? (300.00)\CellColorUnknown & ? (300.00)\CellColorUnknown \\
\hline
rdfbased-sem-bool-intersection-data-localize & +\CellColorSuccess & +\CellColorSuccess & ?\CellColorUnknown & -\CellColorWrong & -\CellColorWrong & -\CellColorWrong & + (0.60)\CellColorSuccess & + (38.46)\CellColorSuccess \\
\hline
rdfbased-sem-bool-intersection-ext & +\CellColorSuccess & +\CellColorSuccess & +\CellColorSuccess & -\CellColorWrong & -\CellColorWrong & -\CellColorWrong & ? (300.00)\CellColorUnknown & ? (300.00)\CellColorUnknown \\
\hline

\end{longtable}
\end{tiny}
\end{center}

\begin{center}
\begin{tiny}
\begin{longtable}{|l||c|} 
\caption{%
Result data of the language coverage experiments for
the FOL theorem prover \emph{iProver-SInE}, 
used with the small-sufficient OWL~2 Full subaxiomatizations
on those test cases where it and Vampire had failed 
when using the complete axiomatization.
}
\label{tab:detailedresults-coverage-iproversine-small}
\\
\hline
\textbf{Test Case} & \textbf{iProver-SInE} \\
\hline
\hline
rdfbased-sem-bool-demorgan & + (36.39)\CellColorSuccess \\
\hline
rdfbased-sem-bool-intersection-ext & + (0.64)\CellColorSuccess \\
\hline
rdfbased-sem-bool-union-ext & + (0.54)\CellColorSuccess \\
\hline
rdfbased-sem-chain-ext & + (35.00)\CellColorSuccess \\
\hline
rdfbased-sem-enum-ext & + (0.17)\CellColorSuccess \\
\hline
rdfbased-sem-restrict-exactcard-inst-subj-two & + (55.83)\CellColorSuccess \\
\hline
rdfbased-sem-restrict-exactqcr-inst-subj-two & ? (300.00)\CellColorUnknown \\
\hline
rdfbased-sem-restrict-maxcard-cmp-prop & + (0.18)\CellColorSuccess \\
\hline
rdfbased-sem-restrict-maxqcr-cmp-prop & + (0.30)\CellColorSuccess \\
\hline
rdfbased-sem-restrict-maxqcr-inst-subj-one & + (1.30)\CellColorSuccess \\
\hline
rdfbased-sem-restrict-maxqcr-inst-subj-zero & + (2.34)\CellColorSuccess \\
\hline
rdfbased-sem-restrict-term-dataqcr & + (0.59)\CellColorSuccess \\
\hline
rdfbased-sem-restrict-term-minmaxexact & + (39.47)\CellColorSuccess \\
\hline
rdfbased-sem-restrict-term-minmaxthing & + (2.03)\CellColorSuccess \\
\hline

\end{longtable}
\end{tiny}
\end{center}

\begin{center}
\begin{tiny}
\begin{longtable}{|l||c|} 
\caption{%
Result data of the language coverage experiments for
the FOL theorem prover \emph{Vampire}, 
used with the small-sufficient OWL~2 Full subaxiomatizations
on those test cases where iProver-SInE failed
when using the small-sufficient axiomatizations.
}
\label{tab:detailedresults-coverage-vampire-small}
\\
\hline
\textbf{Test Case} & \textbf{Vampire} \\
\hline
\hline
rdfbased-sem-restrict-exactqcr-inst-subj-two & + (2.90)\CellColorSuccess \\
\hline

\end{longtable}
\end{tiny}
\end{center}

\subsection{OWL 2 Full-Characteristic Conclusions and Scalability Results}
\label{toc:detailedresults:scalability}

The following tables provide the raw result data
that underlies the results
for the scalability experiments,
as reported in Section~\ref{toc:results:scalability}.
All experiments were conducted
using the test suite of characteristic OWL~2 Full conclusions,
as introduced in Section~\ref{toc:setting:testdata:fullish}
(see Appendix~\ref{toc:fullishtestsuite} for more detailed information
about the test suite).
There is one table per combination
of a FOL reasoner
(\emph{iProver-SInE},
\emph{Vampire} in auto mode,
and \emph{Vampire} using the \emph{SInE} strategy)
and either the complete OWL 2 Full axiomatization
or the small-sufficient subaxiomatizations for the different test cases.
While Section~\ref{toc:results:scalability}
lists only the results for bulk RDF data of size 1~million triples,
the tables here also show results for several intermediate sizes:
1200, 10,000, and 100,000~triples.
In addition, the results for no bulk data (0~triples) are presented,
which were the base for the results
reported in Section~\ref{toc:results:fullish}
for the test suite of characteristic OWL~2 Full conclusions.
No result data for the characteristic conclusion tests is given here
for the Semantic Web reasoners,
since the data provided in Section~\ref{toc:results:fullish}
is already complete for them.
The first column of each table gives the name of the test case,
and the remaining columns gives the results
for the different bulk data sizes.

\begin{center}
\begin{tiny}
\begin{longtable}{|l||c|c|c|c|c|}
\caption{%
Scalability results for the theorem prover \emph{iProver-SInE}
using the complete OWL~2 Full axiomatization.
}
\label{tab:detailedresults-scalability-iproversine-complete}
\\
\hline
\textbf{Test Case} & \textbf{0} & \textbf{1200} & \textbf{10k} & \textbf{100k} & \textbf{1M} \\
\hline
\hline
001\_Subgraph\_Entailment & + (0.15)\CellColorSuccess & + (0.17)\CellColorSuccess & + (0.34)\CellColorSuccess & + (2.16)\CellColorSuccess & + (21.98)\CellColorSuccess \\
\hline
002\_Existential\_Blank\_Nodes & + (0.08)\CellColorSuccess & + (0.10)\CellColorSuccess & + (0.27)\CellColorSuccess & + (2.15)\CellColorSuccess & + (21.81)\CellColorSuccess \\
\hline
003\_Blank\_Nodes\_for\_Literals & + (0.08)\CellColorSuccess & + (0.10)\CellColorSuccess & + (0.27)\CellColorSuccess & + (2.06)\CellColorSuccess & + (22.51)\CellColorSuccess \\
\hline
004\_Axiomatic\_Triples & + (1.23)\CellColorSuccess & + (1.27)\CellColorSuccess & + (1.44)\CellColorSuccess & + (3.43)\CellColorSuccess & + (23.31)\CellColorSuccess \\
\hline
005\_Everything\_is\_a\_Resource & + (3.03)\CellColorSuccess & + (3.03)\CellColorSuccess & + (2.56)\CellColorSuccess & + (5.19)\CellColorSuccess & + (25.07)\CellColorSuccess \\
\hline
006\_Literal\_Values\_represented\_by\_URIs\_and\_Blank\_Nodes & + (11.66)\CellColorSuccess & + (11.65)\CellColorSuccess & + (11.77)\CellColorSuccess & + (13.61)\CellColorSuccess & + (34.00)\CellColorSuccess \\
\hline

\end{longtable}
\end{tiny}
\end{center}

\newpage

\begin{center}
\begin{tiny}
\begin{longtable}{|l||c|c|c|c|c|}
\caption{%
Scalability results for the theorem prover \emph{Vampire} (auto mode)
using the complete OWL~2 Full axiomatization.
}
\label{tab:detailedresults-scalability-vampire-complete}
\\
\hline
\textbf{Test Case} & \textbf{0} & \textbf{1200} & \textbf{10k} & \textbf{100k} & \textbf{1M} \\
\hline
\hline
001\_Subgraph\_Entailment & + (0.02)\CellColorSuccess & + (0.04)\CellColorSuccess & + (0.21)\CellColorSuccess & + (2.31)\CellColorSuccess & + (71.14)\CellColorSuccess \\
\hline
002\_Existential\_Blank\_Nodes & + (0.02)\CellColorSuccess & + (0.03)\CellColorSuccess & + (0.21)\CellColorSuccess & + (2.35)\CellColorSuccess & + (216.63)\CellColorSuccess \\
\hline
003\_Blank\_Nodes\_for\_Literals & + (0.02)\CellColorSuccess & + (0.04)\CellColorSuccess & + (0.21)\CellColorSuccess & + (1.90)\CellColorSuccess & + (73.91)\CellColorSuccess \\
\hline
004\_Axiomatic\_Triples & + (0.43)\CellColorSuccess & + (0.46)\CellColorSuccess & + (0.67)\CellColorSuccess & + (2.49)\CellColorSuccess & ? (300.00)\CellColorUnknown \\
\hline
005\_Everything\_is\_a\_Resource & + (0.03)\CellColorSuccess & + (0.05)\CellColorSuccess & + (0.22)\CellColorSuccess & + (2.38)\CellColorSuccess & ? (300.00)\CellColorUnknown \\
\hline
006\_Literal\_Values\_represented\_by\_URIs\_and\_Blank\_Nodes & + (0.18)\CellColorSuccess & + (0.21)\CellColorSuccess & + (0.38)\CellColorSuccess & + (2.34)\CellColorSuccess & ? (300.00)\CellColorUnknown \\
\hline

\end{longtable}
\end{tiny}
\end{center}

\begin{center}
\begin{tiny}
\begin{longtable}{|l||c|c|c|c|c|}
\caption{%
Scalability results for the theorem prover \emph{Vampire-SInE}
using the complete OWL~2 Full axiomatization.
}
\label{tab:detailedresults-scalability-vampiresine-complete}
\\
\hline
\textbf{Test Case} & \textbf{0} & \textbf{1200} & \textbf{10k} & \textbf{100k} & \textbf{1M} \\
\hline
\hline
001\_Subgraph\_Entailment & + (0.04)\CellColorSuccess & + (0.06)\CellColorSuccess & + (0.20)\CellColorSuccess & + (1.84)\CellColorSuccess & + (20.19)\CellColorSuccess \\
\hline
002\_Existential\_Blank\_Nodes & + (0.04)\CellColorSuccess & + (0.06)\CellColorSuccess & + (0.21)\CellColorSuccess & + (1.81)\CellColorSuccess & + (19.28)\CellColorSuccess \\
\hline
003\_Blank\_Nodes\_for\_Literals & + (0.02)\CellColorSuccess & + (0.04)\CellColorSuccess & + (0.19)\CellColorSuccess & + (1.82)\CellColorSuccess & + (19.34)\CellColorSuccess \\
\hline
004\_Axiomatic\_Triples & + (4.99)\CellColorSuccess & + (5.04)\CellColorSuccess & + (5.21)\CellColorSuccess & + (6.77)\CellColorSuccess & + (24.38)\CellColorSuccess \\
\hline
005\_Everything\_is\_a\_Resource & + (0.06)\CellColorSuccess & + (0.07)\CellColorSuccess & + (0.23)\CellColorSuccess & + (1.86)\CellColorSuccess & + (20.32)\CellColorSuccess \\
\hline
006\_Literal\_Values\_represented\_by\_URIs\_and\_Blank\_Nodes & + (9.63)\CellColorSuccess & + (9.74)\CellColorSuccess & + (9.66)\CellColorSuccess & + (11.43)\CellColorSuccess & + (29.27)\CellColorSuccess \\
\hline

\end{longtable}
\end{tiny}
\end{center}

\begin{center}
\begin{tiny}
\begin{longtable}{|l||c|c|c|c|c|}
\caption{%
Scalability results for the theorem prover \emph{iProver-SInE}
using the small-sufficient OWL~2 Full subaxiomatizations.
}
\label{tab:detailedresults-scalability-iproversine-small}
\\
\hline
\textbf{Test Case} & \textbf{0} & \textbf{1200} & \textbf{10k} & \textbf{100k} & \textbf{1M} \\
\hline
\hline
001\_Subgraph\_Entailment & + (0.04)\CellColorSuccess & + (0.07)\CellColorSuccess & + (0.24)\CellColorSuccess & + (2.03)\CellColorSuccess & + (22.87)\CellColorSuccess \\
\hline
002\_Existential\_Blank\_Nodes & + (0.05)\CellColorSuccess & + (0.07)\CellColorSuccess & + (0.24)\CellColorSuccess & + (2.09)\CellColorSuccess & + (21.85)\CellColorSuccess \\
\hline
003\_Blank\_Nodes\_for\_Literals & + (0.05)\CellColorSuccess & + (0.07)\CellColorSuccess & + (0.26)\CellColorSuccess & + (2.14)\CellColorSuccess & + (22.07)\CellColorSuccess \\
\hline
004\_Axiomatic\_Triples & + (0.10)\CellColorSuccess & + (0.12)\CellColorSuccess & + (0.30)\CellColorSuccess & + (2.12)\CellColorSuccess & + (21.97)\CellColorSuccess \\
\hline
005\_Everything\_is\_a\_Resource & + (0.05)\CellColorSuccess & + (0.08)\CellColorSuccess & + (0.26)\CellColorSuccess & + (2.07)\CellColorSuccess & + (22.50)\CellColorSuccess \\
\hline
006\_Literal\_Values\_represented\_by\_URIs\_and\_Blank\_Nodes & + (0.07)\CellColorSuccess & + (0.09)\CellColorSuccess & + (0.26)\CellColorSuccess & + (2.08)\CellColorSuccess & + (21.96)\CellColorSuccess \\
\hline

\end{longtable}
\end{tiny}
\end{center}

\begin{center}
\begin{tiny}
\begin{longtable}{|l||c|c|c|c|c|}
\caption{%
Scalability results for the theorem prover \emph{Vampire} (auto mode)
using the small-sufficient OWL~2 Full subaxiomatizations.
}
\label{tab:detailedresults-scalability-vampire-small}
\\
\hline
\textbf{Test Case} & \textbf{0} & \textbf{1200} & \textbf{10k} & \textbf{100k} & \textbf{1M} \\
\hline
\hline
001\_Subgraph\_Entailment & + (0.00)\CellColorSuccess & + (0.01)\CellColorSuccess & + (0.20)\CellColorSuccess & + (2.41)\CellColorSuccess & + (48.09)\CellColorSuccess \\
\hline
002\_Existential\_Blank\_Nodes & + (0.00)\CellColorSuccess & + (0.05)\CellColorSuccess & + (1.89)\CellColorSuccess & ? (300.00)\CellColorUnknown & ? (301.00)\CellColorUnknown \\
\hline
003\_Blank\_Nodes\_for\_Literals & + (0.00)\CellColorSuccess & + (0.03)\CellColorSuccess & + (1.49)\CellColorSuccess & + (161.52)\CellColorSuccess & + (182.70)\CellColorSuccess \\
\hline
004\_Axiomatic\_Triples & + (0.01)\CellColorSuccess & + (0.06)\CellColorSuccess & + (2.02)\CellColorSuccess & ? (300.00)\CellColorUnknown & ? (300.00)\CellColorUnknown \\
\hline
005\_Everything\_is\_a\_Resource & + (0.00)\CellColorSuccess & + (0.08)\CellColorSuccess & + (3.17)\CellColorSuccess & ? (300.00)\CellColorUnknown & ? (300.00)\CellColorUnknown \\
\hline
006\_Literal\_Values\_represented\_by\_URIs\_and\_Blank\_Nodes & + (0.00)\CellColorSuccess & + (0.03)\CellColorSuccess & + (1.60)\CellColorSuccess & ? (300.00)\CellColorUnknown & ? (300.00)\CellColorUnknown \\
\hline

\end{longtable}
\end{tiny}
\end{center}

\begin{center}
\begin{tiny}
\begin{longtable}{|l||c|c|c|c|c|}
\caption{%
Scalability results for the theorem prover \emph{Vampire-SInE}
using the small-sufficient OWL~2 Full subaxiomatizations.
}
\label{tab:detailedresults-scalability-vampiresine-small}
\\
\hline
\textbf{Test Case} & \textbf{0} & \textbf{1200} & \textbf{10k} & \textbf{100k} & \textbf{1M} \\
\hline
\hline
001\_Subgraph\_Entailment & + (0.00)\CellColorSuccess & + (0.01)\CellColorSuccess & + (0.16)\CellColorSuccess & + (1.80)\CellColorSuccess & + (19.53)\CellColorSuccess \\
\hline
002\_Existential\_Blank\_Nodes & + (0.00)\CellColorSuccess & + (0.01)\CellColorSuccess & + (0.16)\CellColorSuccess & + (1.76)\CellColorSuccess & + (20.15)\CellColorSuccess \\
\hline
003\_Blank\_Nodes\_for\_Literals & + (0.00)\CellColorSuccess & + (0.01)\CellColorSuccess & + (0.16)\CellColorSuccess & + (1.81)\CellColorSuccess & + (19.52)\CellColorSuccess \\
\hline
004\_Axiomatic\_Triples & + (0.00)\CellColorSuccess & + (0.02)\CellColorSuccess & + (0.17)\CellColorSuccess & + (1.79)\CellColorSuccess & + (19.35)\CellColorSuccess \\
\hline
005\_Everything\_is\_a\_Resource & + (0.00)\CellColorSuccess & + (0.01)\CellColorSuccess & + (0.16)\CellColorSuccess & + (1.78)\CellColorSuccess & + (19.54)\CellColorSuccess \\
\hline
006\_Literal\_Values\_represented\_by\_URIs\_and\_Blank\_Nodes & + (0.00)\CellColorSuccess & + (0.01)\CellColorSuccess & + (0.16)\CellColorSuccess & + (1.80)\CellColorSuccess & + (19.54)\CellColorSuccess \\
\hline

\end{longtable}
\end{tiny}
\end{center}

\subsection{Model Finding Results}
\label{toc:detailedresults:modelfinding}

The following table provides the raw result data
that underlies the results
for the model-finding experiments,
as reported in Section~\ref{toc:results:saturation}.
The only additional data here is the CPU time for each experiment.
All experiments were conducted
using the test suite of characteristic OWL~2 Full conclusions,
as introduced in Section~\ref{toc:setting:testdata:fullish}
(see Appendix~\ref{toc:fullishtestsuite} for more detailed information
about the test suite).

\begin{center}
\begin{tiny}
\begin{longtable}{|l||c|c|c|}
\caption{%
Model finding results
for the model-finders \emph{Paradox} and \emph{DarwinFM}
on the ALCO Full and RDFS axiom sets.
The black entries indicate positive entailments or inconsistent ontologies.
\mbox{PA/A=Paradox/ALCO Full},
\mbox{PA/R=Paradox/RDFS},
\mbox{DF/R=DarwinFM/RDFS}.
}
\label{tab:detailedresults-modelfinding}
\\
\hline
\textbf{Test Case} & \textbf{PA/A} & \textbf{PA/R} & \textbf{DF/R} \\
\hline
\hline
001\_Subgraph\_Entailment & \CellEmpty & \CellEmpty & \CellEmpty  \\
\hline
002\_Existential\_Blank\_Nodes & \CellEmpty & \CellEmpty & \CellEmpty  \\
\hline
003\_Blank\_Nodes\_for\_Literals & \CellEmpty & \CellEmpty & \CellEmpty  \\
\hline
004\_Axiomatic\_Triples & + (13.60)\CellColorSuccess & + (0.73)\CellColorSuccess & + (0.45)\CellColorSuccess \\
\hline
005\_Everything\_is\_a\_Resource & + (15.08)\CellColorSuccess & + (0.90)\CellColorSuccess & + (0.12)\CellColorSuccess \\
\hline
006\_Literal\_Values\_represented\_by\_URIs\_and\_Blank\_Nodes & + (20.95)\CellColorSuccess & + (0.81)\CellColorSuccess & + (0.04)\CellColorSuccess  \\
\hline
007\_Equal\_Classes & + (13.01)\CellColorSuccess & + (1.03)\CellColorSuccess & + (7.19)\CellColorSuccess  \\
\hline
008\_Inverse\_Functional\_Data\_Properties & + (11.74)\CellColorSuccess & + (0.99)\CellColorSuccess & + (0.08)\CellColorSuccess  \\
\hline
009\_Existential\_Restriction\_Entailments & \CellEmpty & + (1.17)\CellColorSuccess & + (0.05)\CellColorSuccess  \\
\hline
010\_Negative\_Property\_Assertions & \CellEmpty & + (1.61)\CellColorSuccess & + (0.07)\CellColorSuccess  \\
\hline
011\_Entity\_Types\_as\_Classes & + (14.15)\CellColorSuccess & + (0.86)\CellColorSuccess & + (0.01)\CellColorSuccess  \\
\hline
012\_Template\_Class & \CellEmpty & + (1.70)\CellColorSuccess & + (0.33)\CellColorSuccess  \\
\hline
013\_Cliques & ? (300.11)\CellColorUnknown & + (2.17)\CellColorSuccess & + (0.05)\CellColorSuccess  \\
\hline
014\_Harry\_belongs\_to\_some\_Species & \CellEmpty & + (1.16)\CellColorSuccess & + (0.56)\CellColorSuccess  \\
\hline
015\_Reflective\_Tautologies\_I & + (10.68)\CellColorSuccess & + (0.75)\CellColorSuccess & + (0.04)\CellColorSuccess  \\
\hline
016\_Reflective\_Tautologies\_II & + (8.21)\CellColorSuccess & + (0.77)\CellColorSuccess & + (2.05)\CellColorSuccess  \\
\hline
017\_Builtin\_Based\_Definitions & + (14.61)\CellColorSuccess & + (0.99)\CellColorSuccess & + (0.06)\CellColorSuccess  \\
\hline
018\_Modified\_Logical\_Vocabulary\_Semantics & + (89.21)\CellColorSuccess & + (0.93)\CellColorSuccess & + (7.35)\CellColorSuccess \\
\hline
019\_Disjoint\_Annotation\_Properties & + (14.55)\CellColorSuccess & + (0.89)\CellColorSuccess & + (0.01)\CellColorSuccess \\
\hline
020\_Logical\_Complications & ? (300.28)\CellColorUnknown & + (1.80)\CellColorSuccess & + (0.85)\CellColorSuccess \\
\hline
021\_Composite\_Enumerations & ? (300.15)\CellColorUnknown & + (2.21)\CellColorSuccess & + (0.11)\CellColorSuccess \\
\hline
022\_List\_Member\_Access & ? (300.15)\CellColorUnknown & + (1.79)\CellColorSuccess & + (0.06)\CellColorSuccess \\
\hline
023\_Unique\_List\_Components & ? (300.15)\CellColorUnknown & + (1.17)\CellColorSuccess & + (0.05)\CellColorSuccess \\
\hline
024\_Cardinality\_Restrictions\_on\_Complex\_Properties & + (16.76)\CellColorSuccess & + (1.16)\CellColorSuccess & + (0.10)\CellColorSuccess \\
\hline
025\_Cyclic\_Dependencies\_between\_Complex\_Properties & ? (300.17)\CellColorUnknown & + (1.65)\CellColorSuccess & + (0.06)\CellColorSuccess \\
\hline
026\_Inferred\_Property\_Characteristics\_I & ? (301.78)\CellColorUnknown & + (1.20)\CellColorSuccess & + (0.07)\CellColorSuccess \\
\hline
027\_Inferred\_Property\_Characteristics\_II & ? (300.12)\CellColorUnknown & + (1.04)\CellColorSuccess & + (0.07)\CellColorSuccess \\
\hline
028\_Inferred\_Property\_Characteristics\_III & + (17.62)\CellColorSuccess & + (1.10)\CellColorSuccess & + (0.07)\CellColorSuccess \\
\hline
029\_Ex\_Falso\_Quodlibet & \CellEmpty & + (1.27)\CellColorSuccess & + (0.07)\CellColorSuccess \\
\hline
030\_Bad\_Class & + (17.88)\CellColorSuccess & + (1.05)\CellColorSuccess & + (0.01)\CellColorSuccess \\
\hline
031\_Large\_Universe & ? (300.55)\CellColorUnknown & + (0.93)\CellColorSuccess & + (0.01)\CellColorSuccess \\
\hline
032\_Datatype\_Relationships & + (9.69)\CellColorSuccess & + (0.85)\CellColorSuccess & + (0.06)\CellColorSuccess \\
\hline

\end{longtable}
\end{tiny}
\end{center}

\newpage

\section{OWL 2 Full Characteristic Conclusions Test Suite}
\label{toc:fullishtestsuite}


This appendix presents the suite of
OWL~2 Full-characteristic conclusion test cases
that was used in the evaluation
and has been introduced in Section~\ref{toc:setting:testdata:fullish}.
The appendix is divided into two parts:
Section~\ref{toc:fullishtestsuite:testcases} 
lists the \emph{test cases},
and Section~\ref{toc:fullishtestsuite:proofs} 
provides \emph{correctness proofs}
for them.
The test suite is also available in electronic form
as part of the \emph{supplementary material} for this paper
(see the download link at the beginning of Section~\ref{toc:setting}),
and can alternatively be obtained as a \emph{separate package} from
\url{http://www.fzi.de/downloads/ipe/schneid/testsuite-fullish.zip}. 

\subsection{Test Cases}
\label{toc:fullishtestsuite:testcases}

Each test case is given by its \emph{name},
its \emph{type}
(one of ``Entailment'' or ``Inconsistency''),
a textual \emph{description},
and the 
\emph{testing data}
as one or two RDF graphs
for an inconsistency test or entailment test, 
respectively.
The RDF graphs are represented in \emph{Turtle} syntax\footnote
{%
Turtle RDF syntax: \url{http://www.w3.org/TeamSubmission/turtle/}
}.
The electronic form of the test suite 
additionally contains serializations 
in \emph{RDF/XML} syntax\footnote
{
RDF/XML syntax: \url{http://www.w3.org/TR/rdf-syntax-grammar/}
}
and in the \emph{TPTP} syntax~\cite{Sut09}. 

\subsubsection{001\_Subgraph\_Entailment (Entailment)}
\label{toc:fullishtestsuite:testcases:001}

In OWL~2 Full, a given RDF graph entails any of its sub graphs, even sub graphs 
that appear to encode broken language constructs of OWL. For example, the 
encoding of a class subsumption axiom that uses a property restriction as 
its superclass entails the single \texttt{owl:onProperty} triple of the serialization. 
This is a characteristic feature of the whole family of RDF-based languages, 
starting with RDF Simple Entailment, and it demonstrates the strictly 
triple-centered view that OWL~2 Full adopts. This behavior is typically 
shown by RDF entailment-rule reasoners, but not by OWL DL reasoners.

\begin{center}
\begin{\FullishSuiteTableEntryFontSize}
\begin{tabular}{p{\FullishSuiteTableEntryColumnWidth}|p{\FullishSuiteTableEntryColumnWidth}}
\textbf{\FullishSuiteTableEntryTitleEntailmentPremiseGraph}
&
\textbf{\FullishSuiteTableEntryTitleEntailmentConclusionGraph} \\
\hline
\begin{verbatim}
ex:c rdfs:subClassOf ex:r .
ex:r rdf:type owl:Restriction .
ex:r owl:onProperty ex:p .
ex:r owl:someValuesFrom ex:d .
\end{verbatim}
&
\begin{verbatim}
ex:r rdf:type owl:Restriction .
ex:r owl:onProperty ex:p .
\end{verbatim}
\end{tabular}
\end{\FullishSuiteTableEntryFontSize}
\end{center}

\subsubsection{002\_Existential\_Blank\_Nodes (Entailment)}
\label{toc:fullishtestsuite:testcases:002}

In OWL~2 Full, every blank node in an RDF graph is interpreted as an 
existentially quantified variable. On the one hand, this means that triples 
with URIs entail corresponding triples with blank nodes substituting the URIs. 
On the other hand, this means that triples with blank nodes entail 
corresponding triples with alternative blank nodes. This feature stems from 
RDF Simple Entailment. Many reasoners, in particular most RDF entailment-rule
reasoners, do not provide the existential semantics of blank nodes.

\begin{center}
\begin{\FullishSuiteTableEntryFontSize}
\begin{tabular}{p{\FullishSuiteTableEntryColumnWidth}|p{\FullishSuiteTableEntryColumnWidth}}
\textbf{\FullishSuiteTableEntryTitleEntailmentPremiseGraph}
&
\textbf{\FullishSuiteTableEntryTitleEntailmentConclusionGraph} \\
\hline
\begin{verbatim}
ex:s ex:p _:o .
_:o ex:q ex:s .
\end{verbatim}
&
\begin{verbatim}
_:x ex:p _:y .
_:y ex:q _:x .
\end{verbatim}
\end{tabular}
\end{\FullishSuiteTableEntryFontSize}
\end{center}

\subsubsection{003\_Blank\_Nodes\_for\_Literals (Entailment)}
\label{toc:fullishtestsuite:testcases:003}

In OWL~2 Full, an RDF triple having a data literal in object position entails 
a corresponding triple with a blank node substituting the literal. This
feature stems from RDF Simple Entailment. It cannot be expected from
OWL DL reasoners, since OWL~2 DL treats such blank nodes as anonymous 
individuals, while the domains of individuals and data values are defined to 
be disjoint. Most RDF entailment-rule reasoners do not show this 
behavior, since they typically do not implement the existential semantics 
of blank nodes.

\begin{center}
\begin{\FullishSuiteTableEntryFontSize}
\begin{tabular}{p{\FullishSuiteTableEntryColumnWidth}|p{\FullishSuiteTableEntryColumnWidth}}
\textbf{\FullishSuiteTableEntryTitleEntailmentPremiseGraph}
&
\textbf{\FullishSuiteTableEntryTitleEntailmentConclusionGraph} \\
\hline
\begin{verbatim}
ex:s ex:p "foo" .
\end{verbatim}
&
\begin{verbatim}
ex:s ex:p _:x .
\end{verbatim}
\end{tabular}
\end{\FullishSuiteTableEntryFontSize}
\end{center}

\subsubsection{004\_Axiomatic\_Triples (Entailment)}
\label{toc:fullishtestsuite:testcases:004}

OWL~2 Full has many tautologies, i.e.\ statements that are entailed by the
empty premise graph. Some of these tautologies have the form of ``axiomatic 
triples'', as defined by RDF and RDFS, but OWL~2 Full goes beyond these 
specifications. An example is the triple ``\texttt{owl:Class} \texttt{rdfs:subClassOf} \texttt{owl:Thing}''. 
RDF entailment-rule reasoners, such as OWL~2 RL/RDF rule reasoners, often 
prove at least some of the tautologies that OWL~2 Full provides, while for 
OWL~2 DL, many of these tautologies are not valid, neither syntactically nor 
semantically.

\begin{center}
\begin{\FullishSuiteTableEntryFontSize}
\begin{tabular}{p{\FullishSuiteTableEntryColumnWidth}|p{\FullishSuiteTableEntryColumnWidth}}
\textbf{\FullishSuiteTableEntryTitleEntailmentPremiseGraph}
&
\textbf{\FullishSuiteTableEntryTitleEntailmentConclusionGraph} \\
\hline
\begin{verbatim}
\end{verbatim}
&
\begin{verbatim}
owl:Class rdf:type owl:Thing .
owl:Class rdf:type owl:Class .
owl:Class rdfs:subClassOf owl:Thing .
owl:Class owl:equivalentClass rdfs:Class .
rdfs:Datatype rdfs:subClassOf owl:Class .
\end{verbatim}
\end{tabular}
\end{\FullishSuiteTableEntryFontSize}
\end{center}

\subsubsection{005\_Everything\_is\_a\_Resource (Entailment)}
\label{toc:fullishtestsuite:testcases:005}

In OWL~2 Full, following the semantics of RDFS, all three nodes of an 
RDF triple denote RDF resources (\texttt{rdfs:Resource}) and OWL individuals 
(\texttt{owl:Thing}). In addition, the predicate node of an RDF triple denotes an 
RDF property (\texttt{rdf:Property}) and an OWL object property (\texttt{owl:ObjectProperty}). 
RDF entailment-rule reasoners will often support this view to at least some 
extent. While OWL~2 DL offers some support for this view syntactically in the 
form of ``punning'', the strict separation of individuals, classes and properties 
in the semantics of OWL~2 DL prevents compliant OWL DL reasoners from producing 
many of the conclusions known from OWL~2 Full. In addition, OWL DL has only
very limited support for RDF entity types such as \texttt{rdf:Property}.

\begin{center}
\begin{\FullishSuiteTableEntryFontSize}
\begin{tabular}{p{\FullishSuiteTableEntryColumnWidth}|p{\FullishSuiteTableEntryColumnWidth}}
\textbf{\FullishSuiteTableEntryTitleEntailmentPremiseGraph}
&
\textbf{\FullishSuiteTableEntryTitleEntailmentConclusionGraph} \\
\hline
\begin{verbatim}
ex:s ex:p ex:o .
\end{verbatim}
&
\begin{verbatim}
ex:s rdf:type rdfs:Resource .
ex:s rdf:type owl:Thing .
ex:p rdf:type rdfs:Resource .
ex:p rdf:type owl:Thing .
ex:p rdf:type rdf:Property .
ex:p rdf:type owl:ObjectProperty .
ex:o rdf:type rdfs:Resource .
ex:o rdf:type owl:Thing .
\end{verbatim}
\end{tabular}
\end{\FullishSuiteTableEntryFontSize}
\end{center}

\subsubsection{006\_Literal\_Values\_represented\_by\_URIs\_and\_Blank\_Nodes (Entailment)}
\label{toc:fullishtestsuite:testcases:006}

In OWL~2 Full, literals can be assigned URIs or blank nodes via \texttt{owl:sameAs} 
statements. One can then use these references to make further assertions 
about the literals and to draw semantic conclusions from them. This is an
often discussed replacement for literals in the subject position of RDF triples, 
which is not supported by the RDF syntax. It is often supported by 
RDF entailment-rule reasoners to some extent, but is not allowed in OWL~2 DL, 
where URIs and blank nodes are used to refer to individuals but not to data 
values.

\begin{center}
\begin{\FullishSuiteTableEntryFontSize}
\begin{tabular}{p{\FullishSuiteTableEntryColumnWidth}|p{\FullishSuiteTableEntryColumnWidth}}
\textbf{\FullishSuiteTableEntryTitleEntailmentPremiseGraph}
&
\textbf{\FullishSuiteTableEntryTitleEntailmentConclusionGraph} \\
\hline
\begin{verbatim}
ex:u owl:sameAs "abc" .
_:x owl:sameAs "abc" .
_:x owl:sameAs ex:w .
\end{verbatim}
&
\begin{verbatim}
ex:u owl:sameAs ex:w .
\end{verbatim}
\end{tabular}
\end{\FullishSuiteTableEntryFontSize}
\end{center}

\subsubsection{007\_Equal\_Classes (Entailment)}
\label{toc:fullishtestsuite:testcases:007}

In OWL~2 Full, asserting that two classes are equal makes them into equivalent
classes. This allows to substitute one class name for the other in all 
class-related axioms, such as class assertions, class subsumption axioms,
and property range axioms. This can be observed in the Linked Open Data cloud, 
which contains many sameAs links between entities that are sometimes used
as as classes in certain contexts. Many RDF entailment-rule reasoners 
provide for the expected semantic results. While syntactically allowed in 
OWL~2 DL via ``punning'', the semantic results are not available due to the 
strict separation of individuals and classes.

\begin{center}
\begin{\FullishSuiteTableEntryFontSize}
\begin{tabular}{p{\FullishSuiteTableEntryColumnWidth}|p{\FullishSuiteTableEntryColumnWidth}}
\textbf{\FullishSuiteTableEntryTitleEntailmentPremiseGraph}
&
\textbf{\FullishSuiteTableEntryTitleEntailmentConclusionGraph} \\
\hline
\begin{verbatim}
ex:c1 owl:sameAs ex:c2 .
ex:w rdf:type ex:c1 .
ex:c rdfs:subClassOf ex:c1 .
ex:p rdfs:range ex:c1 .
\end{verbatim}
&
\begin{verbatim}
ex:w rdf:type ex:c2 .
ex:c rdfs:subClassOf ex:c2 .
ex:p rdfs:range ex:c2 .
\end{verbatim}
\end{tabular}
\end{\FullishSuiteTableEntryFontSize}
\end{center}

\subsubsection{008\_Inverse\_Functional\_Data\_Properties (Entailment)}
\label{toc:fullishtestsuite:testcases:008}

In OWL~2 Full, data properties can be defined as inverse-functional properties. 
This option is, for example, frequently applied in the FOAF specification. 
While many RDF entailment-rule reasoners support the semantic consequences from 
these definitions, they are not supported by OWL~2 DL, which only allows 
object properties to be inverse-functional.

\begin{center}
\begin{\FullishSuiteTableEntryFontSize}
\begin{tabular}{p{\FullishSuiteTableEntryColumnWidth}|p{\FullishSuiteTableEntryColumnWidth}}
\textbf{\FullishSuiteTableEntryTitleEntailmentPremiseGraph}
&
\textbf{\FullishSuiteTableEntryTitleEntailmentConclusionGraph} \\
\hline
\begin{verbatim}
foaf:mbox_sha1sum 
  rdf:type owl:DatatypeProperty ;
  rdf:type owl:InverseFunctionalProperty .
ex:bob foaf:mbox_sha1sum "xyz" .
ex:robert foaf:mbox_sha1sum "xyz" .
\end{verbatim}
&
\begin{verbatim}
ex:bob owl:sameAs ex:robert .
\end{verbatim}
\end{tabular}
\end{\FullishSuiteTableEntryFontSize}
\end{center}

\subsubsection{009\_Existential\_Restriction\_Entailments (Entailment)}
\label{toc:fullishtestsuite:testcases:009}

In OWL~2 Full, a class assertion using an existential property restriction
entails a property assertion with a corresponding blank node. This inference 
is generally be provided by OWL DL reasoners, but in most cases is not 
provable by RDF entailment rule reasoners, which typically do not implement 
the existential semantics of blank nodes and existential property restrictions.

\begin{center}
\begin{\FullishSuiteTableEntryFontSize}
\begin{tabular}{p{\FullishSuiteTableEntryColumnWidth}|p{\FullishSuiteTableEntryColumnWidth}}
\textbf{\FullishSuiteTableEntryTitleEntailmentPremiseGraph}
&
\textbf{\FullishSuiteTableEntryTitleEntailmentConclusionGraph} \\
\hline
\begin{verbatim}
ex:p rdf:type owl:ObjectProperty .
ex:c rdf:type owl:Class .
ex:s rdf:type [
  rdf:type owl:Restriction ;
  owl:onProperty ex:p ;
  owl:someValuesFrom ex:c 
] .
\end{verbatim}
&
\begin{verbatim}
ex:s ex:p _:x .
_:x rdf:type ex:c .
\end{verbatim}
\end{tabular}
\end{\FullishSuiteTableEntryFontSize}
\end{center}

\subsubsection{010\_Negative\_Property\_Assertions (Entailment)}
\label{toc:fullishtestsuite:testcases:010}

OWL~2 has introduced explicit support for negative property assertions (NPAs). 
However, it was already possible to encode NPAs in OWL~1, in terms of 
OWL~1 axioms and class expressions. These definitions are rather complex and 
require strong semantic support for several of the OWL language features. 
OWL~2 Full can infer that the new explicit encoding of NPAs follows from 
the corresponding old encoding of OWL~1. The same holds for OWL~2 DL. In 
contrast, RDF entailment-rule reasoners typically do not allow for such 
inferences due to the high semantic requirements.

\begin{center}
\begin{\FullishSuiteTableEntryFontSize}
\begin{tabular}{p{\FullishSuiteTableEntryColumnWidth}|p{\FullishSuiteTableEntryColumnWidth}}
\textbf{\FullishSuiteTableEntryTitleEntailmentPremiseGraph}
&
\textbf{\FullishSuiteTableEntryTitleEntailmentConclusionGraph} \\
\hline
\begin{verbatim}
ex:p rdf:type owl:ObjectProperty .
ex:s rdf:type [
  owl:onProperty ex:p ;
  owl:allValuesFrom [
    owl:complementOf [
      owl:oneOf ( ex:o ) 
    ]
  ]
] .
\end{verbatim}
&
\begin{verbatim}
_:z rdf:type owl:NegativePropertyAssertion .
_:z owl:sourceIndividual ex:s .
_:z owl:assertionProperty ex:p .
_:z owl:targetIndividual ex:o .
\end{verbatim}
\end{tabular}
\end{\FullishSuiteTableEntryFontSize}
\end{center}

\subsubsection{011\_Entity\_Types\_as\_Classes (Inconsistency)}
\label{toc:fullishtestsuite:testcases:011}

In OWL~2 Full, entity types, such as \texttt{owl:Class}, are regular classes. This 
semantic property is basically inherited from RDFS. This makes it possible, 
for example, to state that the entity types of classes and properties are 
mutually disjoint, and to infer inconsistencies if an entity is used as both 
a class and a property. Some RDF entailment rule reasoners, such as those
implementing the OWL~2 RL/RDF rules, follow this semantics. OWL~2 DL,
on the other hand, does not support it, since it sees entity types are purely
syntactic information.

\begin{center}
\begin{\FullishSuiteTableEntryFontSize}
\begin{tabular}{p{\FullishSuiteTableEntryColumnWidth}}
\textbf{\FullishSuiteTableEntryTitleInconsistencyGraph} \\
\hline
\begin{verbatim}
owl:Class owl:disjointWith owl:ObjectProperty .
ex:x rdf:type owl:Class .
ex:x rdf:type owl:ObjectProperty .
\end{verbatim}
\end{tabular}
\end{\FullishSuiteTableEntryFontSize}
\end{center}

\subsubsection{012\_Template\_Class (Entailment)}
\label{toc:fullishtestsuite:testcases:012}

In OWL~2 Full, instead of explicitly assigning features to a property, 
such as an entity type, property characteristics, or a domain, it is
possible to build a class representing all these features and then make 
the property an instance of this ``template class''. Some RDF entailment rule 
reasoners, such as those implementing the OWL~2 RL/RDF rules, will support
this approach to a certain extent, while in OWL~2 DL, in most cases it is
be syntactically illegal and generally does not have the expected semantic
meaning.

\begin{center}
\begin{\FullishSuiteTableEntryFontSize}
\begin{tabular}{p{\FullishSuiteTableEntryColumnWidth}|p{\FullishSuiteTableEntryColumnWidth}}
\textbf{\FullishSuiteTableEntryTitleEntailmentPremiseGraph}
&
\textbf{\FullishSuiteTableEntryTitleEntailmentConclusionGraph} \\
\hline
\begin{verbatim}
foaf:Person rdf:type owl:Class .
ex:PersonAttribute owl:intersectionOf (
  owl:DatatypeProperty
  owl:FunctionalProperty [ 
    rdf:type owl:Restriction ;
    owl:onProperty rdfs:domain ;
    owl:hasValue foaf:Person 
  ]
) .
ex:name rdf:type ex:PersonAttribute .
ex:alice ex:name "alice" .
\end{verbatim}
&
\begin{verbatim}
ex:name rdf:type owl:FunctionalProperty .
ex:alice rdf:type foaf:Person .
\end{verbatim}
\end{tabular}
\end{\FullishSuiteTableEntryFontSize}
\end{center}

\subsubsection{013\_Cliques (Entailement)}
\label{toc:fullishtestsuite:testcases:013}

OWL~2 Full can define the metaclass of all cliques, for which each 
instance is a clique of people that know everyone else in that clique. 
The encoding is not supported by OWL~2 DL, since it uses built-in
vocabulary terms as regular entities. For RDF entailment rule reasoners,
the semantic requirements for producing all expected results are typically
too high.

\begin{center}
\begin{\FullishSuiteTableEntryFontSize}
\begin{tabular}{p{\FullishSuiteTableEntryColumnWidth}|p{\FullishSuiteTableEntryColumnWidth}}
\textbf{\FullishSuiteTableEntryTitleEntailmentPremiseGraph}
&
\textbf{\FullishSuiteTableEntryTitleEntailmentConclusionGraph} \\
\hline
\begin{verbatim}
ex:Clique rdf:type owl:Class .
ex:sameCliqueAs 
  rdfs:subPropertyOf owl:sameAs ;
  rdfs:range ex:Clique .
ex:Clique rdfs:subClassOf [
  rdf:type owl:Restriction ;
  owl:onProperty ex:sameCliqueAs ;
  owl:someValuesFrom ex:Clique 
] .
foaf:knows 
  rdf:type owl:ObjectProperty ;
  owl:propertyChainAxiom ( 
  rdf:type 
  ex:sameCliqueAs 
  [owl:inverseOf rdf:type] 
) .
ex:JoesGang rdf:type ex:Clique .
ex:alice rdf:type ex:JoesGang .
ex:bob rdf:type ex:JoesGang .
\end{verbatim}
&
\begin{verbatim}
ex:alice foaf:knows ex:bob .
\end{verbatim}
\end{tabular}
\end{\FullishSuiteTableEntryFontSize}
\end{center}

\subsubsection{014\_Harry\_belongs\_to\_some\_Species (Entailment)}
\label{toc:fullishtestsuite:testcases:014}

OWL~2 Full supports the combination of metamodelling and class union. For 
example, provided that the classes of eagles and falcons are both instances
of the metaclass of species, if one does not exactly know whether Harry is 
an eagle or a falcon, one can still conclude that Harry must belong to some 
species. OWL~2 DL does not support semantic conclusions from metamodeling,
although it allows for some metamodeling syntactically via ``punning''. While
many RDF entailment-rule reasoners have some restricted support for semantic 
metamodeling, drawing said conclusion from the union of classes typically
goes beyond the capabilities of these reasoners.

\begin{center}
\begin{\FullishSuiteTableEntryFontSize}
\begin{tabular}{p{\FullishSuiteTableEntryColumnWidth}|p{\FullishSuiteTableEntryColumnWidth}}
\textbf{\FullishSuiteTableEntryTitleEntailmentPremiseGraph}
&
\textbf{\FullishSuiteTableEntryTitleEntailmentConclusionGraph} \\
\hline
\begin{verbatim}
ex:Eagle rdf:type ex:Species .
ex:Falcon rdf:type ex:Species .
ex:harry rdf:type [ 
  owl:unionOf ( ex:Eagle ex:Falcon ) 
] .
\end{verbatim}
&
\begin{verbatim}
ex:harry rdf:type _:x .
_:x rdf:type ex:Species .
\end{verbatim}
\end{tabular}
\end{\FullishSuiteTableEntryFontSize}
\end{center}

\subsubsection{015\_Reflective\_Tautologies\_I (Entailment)}
\label{toc:fullishtestsuite:testcases:015}

In OWL~2 Full, the statement ``\texttt{owl:sameAs} \texttt{owl:sameAs} \texttt{owl:sameAs}'' is a tautology.
This is a classic example used to demonstrate the use of built-in vocabulary 
terms as regular entities, sometimes referred to as ``syntax reflection''. 
It is not allowed in OWL~2 DL. Some RDF entailment-rule reasoners, such as those
implementing the OWL~2 RL/RDF rules, do provide this result.

\begin{center}
\begin{\FullishSuiteTableEntryFontSize}
\begin{tabular}{p{\FullishSuiteTableEntryColumnWidth}|p{\FullishSuiteTableEntryColumnWidth}}
\textbf{\FullishSuiteTableEntryTitleEntailmentPremiseGraph}
&
\textbf{\FullishSuiteTableEntryTitleEntailmentConclusionGraph} \\
\hline
\begin{verbatim}
\end{verbatim}
&
\begin{verbatim}
owl:sameAs owl:sameAs owl:sameAs .
\end{verbatim}
\end{tabular}
\end{\FullishSuiteTableEntryFontSize}
\end{center}

\subsubsection{016\_Reflective\_Tautologies\_II (Entailment)}
\label{toc:fullishtestsuite:testcases:016}

In OWL~2 Full, the class equivalence property is a subproperty
of the class subsumption property. This is an example of the use of built-in
vocabulary terms as regular entities, occasionally referred to as 
``syntax reflection''. It is not allowed in OWL~2 DL. RDF entailment-rule
reasoners may contain this tautology as a special rule, but otherwise
cannot be expected to provide this result. For example, the result does not
follow from the OWL~2 RL/RDF rules.

\begin{center}
\begin{\FullishSuiteTableEntryFontSize}
\begin{tabular}{p{\FullishSuiteTableEntryColumnWidth}|p{\FullishSuiteTableEntryColumnWidth}}
\textbf{\FullishSuiteTableEntryTitleEntailmentPremiseGraph}
&
\textbf{\FullishSuiteTableEntryTitleEntailmentConclusionGraph} \\
\hline
\begin{verbatim}
\end{verbatim}
&
\begin{verbatim}
owl:equivalentClass 
  rdfs:subPropertyOf rdfs:subClassOf .
\end{verbatim}
\end{tabular}
\end{\FullishSuiteTableEntryFontSize}
\end{center}

\subsubsection{017\_Builtin\_Based\_Definitions (Entailment)}
\label{toc:fullishtestsuite:testcases:017}

In OWL~2 Full, custom properties can be defined based on existing built-in 
properties. For example, a property \texttt{ex:noInstanceOf} that is disjoint 
from \texttt{rdf:type} can be defined, and this new property can be used to 
state non-membership, which has semantic ramifications. 
OWL~2 DL does not allow this.
Entailment-rule reasoners can make such assertions, and may provide
some limited support for semantic conclusions.

\begin{center}
\begin{\FullishSuiteTableEntryFontSize}
\begin{tabular}{p{\FullishSuiteTableEntryColumnWidth}|p{\FullishSuiteTableEntryColumnWidth}}
\textbf{\FullishSuiteTableEntryTitleEntailmentPremiseGraph}
&
\textbf{\FullishSuiteTableEntryTitleEntailmentConclusionGraph} \\
\hline
\begin{verbatim}
ex:notInstanceOf 
  owl:propertyDisjointWith rdf:type .
ex:w rdf:type ex:c .
ex:u ex:notInstanceOf ex:c .
\end{verbatim}
&
\begin{verbatim}
ex:w owl:differentFrom ex:u .
\end{verbatim}
\end{tabular}
\end{\FullishSuiteTableEntryFontSize}
\end{center}

\subsubsection{018\_Modified\_Logical\_Vocabulary\_Semantics (Entailment)}
\label{toc:fullishtestsuite:testcases:018}

The semantics of OWL built-in vocabulary terms can be enriched in a way such
that their application leads to additional results that are not available 
from their original meaning. For example, the domain and range of 
\texttt{owl:sameAs} can be restricted to the class of persons, which renders 
all things that are equal into persons. 
OWL~2 DL does not allow this, while RDF entailment-rule
reasoners often provide some limited support.

\begin{center}
\begin{\FullishSuiteTableEntryFontSize}
\begin{tabular}{p{\FullishSuiteTableEntryColumnWidth}|p{\FullishSuiteTableEntryColumnWidth}}
\textbf{\FullishSuiteTableEntryTitleEntailmentPremiseGraph}
&
\textbf{\FullishSuiteTableEntryTitleEntailmentConclusionGraph} \\
\hline
\begin{verbatim}
owl:sameAs rdfs:domain ex:Person .
ex:w owl:sameAs ex:u .
\end{verbatim}
&
\begin{verbatim}
ex:u rdf:type ex:Person .
\end{verbatim}
\end{tabular}
\end{\FullishSuiteTableEntryFontSize}
\end{center}

\subsubsection{019\_Disjoint\_Annotation\_Properties (Inconsistency)}
\label{toc:fullishtestsuite:testcases:019}

In OWL~2 Full, annotation properties are normal object properties. Thus, two
annotation properties can be specified to be disjoint, and semantic conclusions
can be drawn from this disjointness. This feature is, for example, used
in the SKOS specification to define the meaning of lexical labels. OWL~2 DL
provides only limited syntactic support for putting axioms on annotation
properties, and does not provide any semantic conclusions. One can expect
limited semantic support from some RDF entailment-rule reasoners, such as
those implementing the OWL~2 RL/RDF rules.

\begin{center}
\begin{\FullishSuiteTableEntryFontSize}
\begin{tabular}{p{\FullishSuiteTableEntryColumnWidth}}
\textbf{\FullishSuiteTableEntryTitleInconsistencyGraph} \\
\hline
\begin{verbatim}
skos:prefLabel rdf:type owl:AnnotationProperty .
skos:prefLabel rdfs:subPropertyOf rdfs:label .
skos:altLabel rdf:type owl:AnnotationProperty .
skos:altLabel rdfs:subPropertyOf rdfs:label .
skos:prefLabel owl:propertyDisjointWith skos:altLabel .
ex:foo skos:prefLabel "foo" .
ex:foo skos:altLabel "foo" .
\end{verbatim}
\end{tabular}
\end{\FullishSuiteTableEntryFontSize}
\end{center}

\subsubsection{020\_Logical\_Complications (Entailment)}
\label{toc:fullishtestsuite:testcases:020}

OWL~2 Full allows complex logical reasoning to be performed. For example, 
non-obvious subsumption relationships between two classes can be inferred 
based on the application of disjointness and different Boolean connectives. 
This kind of reasoning is generally possible in unrestricted form in OWL~2 DL,
but typically not with RDF entailment-rule reasoners.

\begin{center}
\begin{\FullishSuiteTableEntryFontSize}
\begin{tabular}{p{\FullishSuiteTableEntryColumnWidth}|p{\FullishSuiteTableEntryColumnWidth}}
\textbf{\FullishSuiteTableEntryTitleEntailmentPremiseGraph}
&
\textbf{\FullishSuiteTableEntryTitleEntailmentConclusionGraph} \\
\hline
\begin{verbatim}
ex:c owl:unionOf ( ex:c1 ex:c2 ex:c3 ) .
ex:d owl:disjointWith ex:c1 .
ex:d rdfs:subClassOf [ 
  owl:intersectionOf (
    ex:c 
    [ owl:complementOf ex:c2 ] 
  ) 
] .
\end{verbatim}
&
\begin{verbatim}
ex:d rdfs:subClassOf ex:c3 .
\end{verbatim}
\end{tabular}
\end{\FullishSuiteTableEntryFontSize}
\end{center}

\subsubsection{021\_Composite\_Enumerations (Entailment)}
\label{toc:fullishtestsuite:testcases:021}

OWL~2 Full allows for the composition of enumerations via boolean connectives.
For example, the union of the classes $\{w1,w2\}$ and $\{w2,w3\}$ can be inferred
to be equivalent to the class $\{w1,w2,w3\}$. OWL~2 DL reasoners can be expected
to provide this result, while RDF entailment-rule reasoners are typically 
unable to produce the result.

\begin{center}
\begin{\FullishSuiteTableEntryFontSize}
\begin{tabular}{p{\FullishSuiteTableEntryColumnWidth}|p{\FullishSuiteTableEntryColumnWidth}}
\textbf{\FullishSuiteTableEntryTitleEntailmentPremiseGraph}
&
\textbf{\FullishSuiteTableEntryTitleEntailmentConclusionGraph} \\
\hline
\begin{verbatim}
ex:c1 owl:oneOf ( ex:w1 ex:w2 ) .
ex:c2 owl:oneOf ( ex:w2 ex:w3 ) .
ex:c3 owl:oneOf ( ex:w1 ex:w2 ex:w3 ) .
ex:c4 owl:unionOf ( ex:c1 ex:c2 ) .
\end{verbatim}
&
\begin{verbatim}
ex:c3 owl:equivalentClass ex:c4 .
\end{verbatim}
\end{tabular}
\end{\FullishSuiteTableEntryFontSize}
\end{center}

\subsubsection{022\_List\_Member\_Access (Entailment)}
\label{toc:fullishtestsuite:testcases:022}

In OWL~2 Full, one can refer to all items within an RDF list. For example, 
Chapter~9 of the SKOS Reference defines ordered concept collections via 
the property \texttt{skos:memberList} applied to some RDF list consisting of 
items of type \texttt{skos:Concept}. SKOS further defines non-ordered concept 
collections by applying the property \texttt{skos:member} repeatedly to single 
entities of type \texttt{skos:Concept}. SKOS statement~S36 says that 
a non-ordered concept collection can be inferred from an ordered collection. 
An example is given in Section~9.6.1 of the SKOS Reference. 
OWL~2 Full allows this statement to be expressed semantically. 
Both the encoding of S36 and the example inference is given here. 
RDF entailment-rule reasoners implementing the OWL~2 RL/RDF rules also 
produce the result. OWL~2 DL cannot make assertions about RDF lists.

\begin{center}
\begin{\FullishSuiteTableEntryFontSize}
\begin{tabular}{p{\FullishSuiteTableEntryColumnWidth}|p{\FullishSuiteTableEntryColumnWidth}}
\textbf{\FullishSuiteTableEntryTitleEntailmentPremiseGraph}
&
\textbf{\FullishSuiteTableEntryTitleEntailmentConclusionGraph} \\
\hline
\begin{verbatim}
skos:memberList rdfs:subPropertyOf _:pL .
skos:member owl:propertyChainAxiom ( 
  _:pL 
  rdf:first 
) .
_:pL owl:propertyChainAxiom ( 
  _:pL 
  rdf:rest 
) .
ex:MyOrderedCollection 
  rdf:type skos:OrderedCollection ;
  skos:memberList ( ex:X ex:Y ex:Z ) .
\end{verbatim}
&
\begin{verbatim}
ex:MyOrderedCollection skos:member ex:X .
ex:MyOrderedCollection skos:member ex:Y .
ex:MyOrderedCollection skos:member ex:Z .
\end{verbatim}
\end{tabular}
\end{\FullishSuiteTableEntryFontSize}
\end{center}

\subsubsection{023\_Unique\_List\_Components (Entailment)}
\label{toc:fullishtestsuite:testcases:023}

In principle, it is possible to create argument lists of OWL constructs 
that are non-linear. Section~3.3.3 of the RDF Semantics specification allows 
semantic extensions to place extra syntactic wellformedness 
restrictions on the use of the RDF Collections vocabulary in order to 
rule out graphs containing non-linear lists. While OWL~2 Full does not 
provide this directly, it can state that the List vocabulary 
property \texttt{rdf:first} is a functional property. This has semantic consequences 
even if the argument list of an OWL construct is given in a non-linear form.
RDF entailment-rule reasoners often have some limited support for these
kinds of results. OWL~2 DL cannot make assertions about RDF lists.

\begin{center}
\begin{\FullishSuiteTableEntryFontSize}
\begin{tabular}{p{\FullishSuiteTableEntryColumnWidth}|p{\FullishSuiteTableEntryColumnWidth}}
\textbf{\FullishSuiteTableEntryTitleEntailmentPremiseGraph}
&
\textbf{\FullishSuiteTableEntryTitleEntailmentConclusionGraph} \\
\hline
\begin{verbatim}
rdf:first rdf:type owl:FunctionalProperty .
ex:w rdf:type [
  rdf:type owl:Class ;
  owl:oneOf _:l
] .
_:l rdf:first ex:u .
_:l rdf:first ex:v .
_:l rdf:rest rdf:nil .
\end{verbatim}
&
\begin{verbatim}
ex:w owl:sameAs ex:u .
ex:w owl:sameAs ex:v .
\end{verbatim}
\end{tabular}
\end{\FullishSuiteTableEntryFontSize}
\end{center}

\subsubsection{024\_Cardinality\_Restrictions\_on\_Complex\_Properties (Entailment)}
\label{toc:fullishtestsuite:testcases:024}

OWL~2 DL does cannot place cardinality restrictions on transitive 
properties. OWL~2 Full allows this. This can, for example, be used to state 
that every person has at least one ancestor. The existence of an ancestor can 
then be inferred for any given person. RDF entailment-rule reasoners may
provide some limited support but typically are unable to produce the
result of this particular example.

\begin{center}
\begin{\FullishSuiteTableEntryFontSize}
\begin{tabular}{p{\FullishSuiteTableEntryColumnWidth}|p{\FullishSuiteTableEntryColumnWidth}}
\textbf{\FullishSuiteTableEntryTitleEntailmentPremiseGraph}
&
\textbf{\FullishSuiteTableEntryTitleEntailmentConclusionGraph} \\
\hline
\begin{verbatim}
ex:hasAncestor 
  rdf:type owl:TransitiveProperty .
ex:Person rdfs:subClassOf [
  rdf:type owl:Restriction ;
  owl:onProperty ex:hasAncestor ;
  owl:minCardinality 
    "1"^^xsd:nonNegativeInteger 
] .
ex:alice rdf:type ex:Person .
ex:bob rdf:type ex:Person .
ex:alice ex:hasAncestor ex:bob .
\end{verbatim}
&
\begin{verbatim}
ex:bob ex:hasAncestor _:x .
ex:alice ex:hasAncestor _:x .
\end{verbatim}
\end{tabular}
\end{\FullishSuiteTableEntryFontSize}
\end{center}

\subsubsection{025\_Cyclic\_Dependencies\_between\_Complex\_Properties (Entailment)}
\label{toc:fullishtestsuite:testcases:025}

OWL~2 DL does not allow cyclic dependencies between complex properties that
are defined via subproperty chain axioms. OWL~2 Full allows this. For example, 
the uncle relation and the cousin relation can be expressed mutually in terms
of the other relation using two subproperty chain axioms. This
provides for more precise characterizations of properties than it is 
possible in OWL~2 DL. RDF entailment rule reasoners that implement the 
OWL~2 RL/RDF rules provide limited support for reasoning in such scenarios.

\begin{center}
\begin{\FullishSuiteTableEntryFontSize}
\begin{tabular}{p{\FullishSuiteTableEntryColumnWidth}|p{\FullishSuiteTableEntryColumnWidth}}
\textbf{\FullishSuiteTableEntryTitleEntailmentPremiseGraph}
&
\textbf{\FullishSuiteTableEntryTitleEntailmentConclusionGraph} \\
\hline
\begin{verbatim}
ex:hasUncle owl:propertyChainAxiom (
  ex:hasCousin 
  ex:hasFather
) .
ex:hasCousin owl:propertyChainAxiom ( 
  ex:hasUncle 
  [ owl:inverseOf ex:hasFather ] 
) .
ex:alice ex:hasFather ex:dave .
ex:alice ex:hasCousin ex:bob .
ex:bob ex:hasFather ex:charly .
ex:bob ex:hasUncle ex:dave .
\end{verbatim}
&
\begin{verbatim}
ex:alice ex:hasUncle ex:charly .
ex:bob ex:hasCousin ex:alice .
\end{verbatim}
\end{tabular}
\end{\FullishSuiteTableEntryFontSize}
\end{center}

\subsubsection{026\_Inferred\_Property\_Characteristics\_I (Entailment)}
\label{toc:fullishtestsuite:testcases:026}

In OWL~2 Full, as in OWL~2 DL, a property that has a domain and a range being
singleton classes is entailed to be an inverse-functional property.
RDF entailment-rule reasoners cannot be expected to provide this result,
since it requires sophisticated reasoning.

\begin{center}
\begin{\FullishSuiteTableEntryFontSize}
\begin{tabular}{p{\FullishSuiteTableEntryColumnWidth}|p{\FullishSuiteTableEntryColumnWidth}}
\textbf{\FullishSuiteTableEntryTitleEntailmentPremiseGraph}
&
\textbf{\FullishSuiteTableEntryTitleEntailmentConclusionGraph} \\
\hline
\begin{verbatim}
ex:p rdfs:domain [ owl:oneOf ( ex:w ) ] .
ex:p rdfs:range [ owl:oneOf ( ex:u ) ] .
\end{verbatim}
&
\begin{verbatim}
ex:p rdf:type owl:InverseFunctionalProperty .
\end{verbatim}
\end{tabular}
\end{\FullishSuiteTableEntryFontSize}
\end{center}

\subsubsection{027\_Inferred\_Property\_Characteristics\_II (Entailment)}
\label{toc:fullishtestsuite:testcases:027}

In OWL~2 Full, if the chain of a property and its inverse property builds a 
subproperty chain of \texttt{owl:sameAs}, then that property is 
inverse-functional.
The application of the built-in vocabulary term \texttt{owl:sameAs} is 
not allowed in OWL~2 DL. 
Newer RDF entailment-rule reasoners, such as those implementing
the OWL~2 RL/RDF rules, may provide some limited semantic support.

\begin{center}
\begin{\FullishSuiteTableEntryFontSize}
\begin{tabular}{p{\FullishSuiteTableEntryColumnWidth}|p{\FullishSuiteTableEntryColumnWidth}}
\textbf{\FullishSuiteTableEntryTitleEntailmentPremiseGraph}
&
\textbf{\FullishSuiteTableEntryTitleEntailmentConclusionGraph} \\
\hline
\begin{verbatim}
owl:sameAs owl:propertyChainAxiom ( 
  ex:p 
  [owl:inverseOf ex:p] 
) .
\end{verbatim}
&
\begin{verbatim}
ex:p rdf:type owl:InverseFunctionalProperty .
\end{verbatim}
\end{tabular}
\end{\FullishSuiteTableEntryFontSize}
\end{center}

\subsubsection{028\_Inferred\_Property\_Characteristics\_III (Entailment)}
\label{toc:fullishtestsuite:testcases:028}

In OWL~2 Full, instead of using the built-in property characteristics of
inverse-functional properties, properties can be made into 
instances of the custom class of the inverses of all functional properties.
OWL~2 DL does not allow the use of built-in vocabulary terms as regular
entities. For RDF entailment-rule reasoners, the semantic result given in this 
example is typically too demanding.

\begin{center}
\begin{\FullishSuiteTableEntryFontSize}
\begin{tabular}{p{\FullishSuiteTableEntryColumnWidth}|p{\FullishSuiteTableEntryColumnWidth}}
\textbf{\FullishSuiteTableEntryTitleEntailmentPremiseGraph}
&
\textbf{\FullishSuiteTableEntryTitleEntailmentConclusionGraph} \\
\hline
\begin{verbatim}
ex:InversesOfFunctionalProperties 
  owl:equivalentClass [
    rdf:type owl:Restriction ;
    owl:onProperty owl:inverseOf ;
    owl:someValuesFrom owl:FunctionalProperty 
  ] .
\end{verbatim}
&
\begin{verbatim}
ex:InversesOfFunctionalProperties 
  rdfs:subClassOf owl:InverseFunctionalProperty .
\end{verbatim}
\end{tabular}
\end{\FullishSuiteTableEntryFontSize}
\end{center}

\subsubsection{029\_Ex\_Falso\_Quodlibet (Entailment)}
\label{toc:fullishtestsuite:testcases:029}

In OWL~2 Full, an inconsistent premise ontology entails arbitrary conclusion
ontologies (``principle of explosion'', ``ex falso sequitur quodlibet''). OWL~2 DL
has the same semantic property, but many existing OWL~2 DL reasoners 
signal an error when given an inconsistent premise ontology, and do not
produce the expected result (however, it would be trivial to extend 
an OWL~2 DL reasoner to give the result as a reaction to an inconsistency
error). 
RDF entailment-rule reasoners cannot be expected to produce tyhis result, 
since it requires full semantic support for classical negation.

\begin{center}
\begin{\FullishSuiteTableEntryFontSize}
\begin{tabular}{p{\FullishSuiteTableEntryColumnWidth}|p{\FullishSuiteTableEntryColumnWidth}}
\textbf{\FullishSuiteTableEntryTitleEntailmentPremiseGraph}
&
\textbf{\FullishSuiteTableEntryTitleEntailmentConclusionGraph} \\
\hline
\begin{verbatim}
ex:A rdf:type owl:Class .
ex:B rdf:type owl:Class .
ex:w rdf:type [ 
  owl:intersectionOf ( 
    ex:A 
    [owl:complementOf ex:A]
  ) 
] .
\end{verbatim}
&
\begin{verbatim}
ex:w rdf:type ex:B .
\end{verbatim}
\end{tabular}
\end{\FullishSuiteTableEntryFontSize}
\end{center}

\subsubsection{030\_Bad\_Class (Inconsistency)}
\label{toc:fullishtestsuite:testcases:030}

If an OWL~2 Full ontology contains a class that has the Russell Set 
as its class extension, then the ontology is inconsistent. This situation
would occur for even the empty ontology if the so-called OWL~2 Full 
comprehension conditions, as non-normatively defined in Chapter~8 of the 
OWL~2 RDF-Based Semantics, were a normative part of OWL~2 Full, as 
explained in Chapter~9 of the specification document. OWL~2 DL does not 
know about this issue, and RDF entailment-rule reasoners cannot be expected to 
know about it due to their relatively weak semantics.

\begin{center}
\begin{\FullishSuiteTableEntryFontSize}
\begin{tabular}{p{\FullishSuiteTableEntryColumnWidth}}
\textbf{\FullishSuiteTableEntryTitleInconsistencyGraph} \\
\hline
\begin{verbatim}
ex:c rdf:type owl:Class . 
ex:c owl:complementOf [
  rdf:type owl:Restriction ;
  owl:onProperty rdf:type ;
  owl:hasSelf "true"^^xsd:boolean 
] .
\end{verbatim}
\end{tabular}
\end{\FullishSuiteTableEntryFontSize}
\end{center}

\subsubsection{031\_Large\_Universe (Inconsistency)}
\label{toc:fullishtestsuite:testcases:031}

The universe of an OWL~2 Full interpretation cannot consist of only a single 
individual. This means that \texttt{owl:Thing} cannot be equivalent to 
a singleton enumeration class, without leading to an inconsistent ontology. 
This is different from OWL~2 DL, for which the only restriction on the 
universe is that it has to be non-empty. RDF entailment-rule reasoners 
cannot be expected to provide the inconsistency result, since this requires
strong logic-based reasoning.

\begin{center}
\begin{\FullishSuiteTableEntryFontSize}
\begin{tabular}{p{\FullishSuiteTableEntryColumnWidth}}
\textbf{\FullishSuiteTableEntryTitleInconsistencyGraph} \\
\hline
\begin{verbatim}
owl:Thing owl:equivalentClass [ 
  owl:oneOf ( ex:w ) 
] .
\end{verbatim}
\end{tabular}
\end{\FullishSuiteTableEntryFontSize}
\end{center}

\subsubsection{032\_Datatype\_Relationships (Entailment)}
\label{toc:fullishtestsuite:testcases:032}

According to the XSD Datatypes specification, the value spaces of the
datatypes \texttt{xsd:decimal} and \texttt{xsd:string} are disjoint, while the value space
of \texttt{xsd:integer} is a subset of the value space of \texttt{xsd:decimal}. In OWL~2 Full,
these relationships between the data values of datatypes can be observed as 
corresponding relationships between the classes representing these datatypes.
OWL~2 DL also follows the XSD semantics, but it does not support to explicitly
query for subsumption or disjointness relationships between datatypes. Some
RDF entailment-rule reasoners may decide to provide the different relationships
between XSD datatypes as explicit facts or rules, but cannot, in general, be
expected to do so.

\begin{center}
\begin{\FullishSuiteTableEntryFontSize}
\begin{tabular}{p{\FullishSuiteTableEntryColumnWidth}|p{\FullishSuiteTableEntryColumnWidth}}
\textbf{\FullishSuiteTableEntryTitleEntailmentPremiseGraph}
&
\textbf{\FullishSuiteTableEntryTitleEntailmentConclusionGraph} \\
\hline
\begin{verbatim}
\end{verbatim}
&
\begin{verbatim}
xsd:decimal owl:disjointWith xsd:string .
xsd:integer rdfs:subClassOf xsd:decimal .
\end{verbatim}
\end{tabular}
\end{\FullishSuiteTableEntryFontSize}
\end{center}

\subsection{Correctness Proofs}
\label{toc:fullishtestsuite:proofs}

This section provides \emph{correctness proofs}
for all test cases listed in Section~\ref{toc:fullishtestsuite:testcases}.
The proofs have been constructed
with respect to the OWL~2 RDF Based Semantics~\cite{w3c09-owl2-rdf-semantics}
and the RDF Semantics~\cite{w3c04-rdf-semantics},
which conjointly specify the model-theoretic semantics of OWL~2 Full.

\subsubsection{001\_Subgraph\_Entailment (Proof)}
\label{toc:fullishtestsuite:proofs:001}

Let $I$ be an OWL~2 RDF-Based interpretation that satisfies the premise graph,
so the following becomes true:
\[
\begin{array}{l}
    \langle I(\texttt{ex:c}), I(\texttt{ex:r})\rangle  \in \mathrm{IEXT}(I(\texttt{rdfs:subClassOf})) \\
    \langle I(\texttt{ex:r}), I(\texttt{owl:Restriction})\rangle  \in \mathrm{IEXT}(I(\texttt{rdf:type})) \\
    \langle I(\texttt{ex:r}), I(\texttt{ex:p})\rangle  \in \mathrm{IEXT}(I(\texttt{owl:onProperty})) \\
    \langle I(\texttt{ex:r}), I(\texttt{ex:d})\rangle  \in \mathrm{IEXT}(I(\texttt{owl:someValuesFrom}))
\end{array}
\]
Then, in particular, the conjunction of the subset of atoms
\[
\begin{array}{l}
    \langle I(\texttt{ex:r}), I(\texttt{owl:Restriction})\rangle  \in \mathrm{IEXT}(I(\texttt{rdf:type})) \\
    \langle I(\texttt{ex:r}), I(\texttt{ex:p})\rangle  \in \mathrm{IEXT}(I(\texttt{owl:onProperty}))
\end{array}
\]
is also satisfied.

\subsubsection{002\_Existential\_Blank\_Nodes (Proof)}
\label{toc:fullishtestsuite:proofs:002}

Let $I$ be an OWL~2 RDF-Based interpretation interpretation and $B$ be a blank node 
mapping for the blank nodes in the premise graph, such that $I+B$ satisfies 
the premise graph. This gives
\[
    \exists o:\: \langle I(\texttt{ex:s}), o\rangle  \in \mathrm{IEXT}(I(\texttt{ex:p})) 
    \;\wedge\; 
    \langle o, I(\texttt{ex:s})\rangle  \in \mathrm{IEXT}(I(\texttt{ex:q}))
\]
Weakening this statement by introducing an existentially quantified variable 
for $I(\texttt{ex:s})$ logically implies
\[
    \exists x,y:\: \langle x, y\rangle  \in \mathrm{IEXT}(I(\texttt{ex:p})) 
    \;\wedge\; 
    \langle y, x\rangle  \in \mathrm{IEXT}(I(\texttt{ex:q}))
\]
Thus, there is a blank node mapping~$B'$, such that $I+B'$ satisfies 
the conclusion graph.

\subsubsection{003\_Blank\_Nodes\_for\_Literals (Proof)}
\label{toc:fullishtestsuite:proofs:003}

Let $I$ be an OWL~2 RDF-Based interpretation that satisfies the premise graph. 
Then from
\[
    \langle I(\texttt{ex:s}), I(\texttt{"foo"})\rangle  \in \mathrm{IEXT}(I(\texttt{ex:p}))
\]
and taking into account that literals denote individuals in the universe,
we receive the formally weaker assertion
\[
    \exists x:\: \langle I(\texttt{ex:s}), x\rangle  \in \mathrm{IEXT}(I(\texttt{ex:p}))
\]
Thus, there is a blank node mapping~$B$, such that $I+B$ satisfies the 
conclusion graph.

\subsubsection{004\_Axiomatic\_Triples (Proof)}
\label{toc:fullishtestsuite:proofs:004}

Given a satisfying OWL~2 RDF-Based interpretation~$I$ for the empty graph.

\noindent\emph{1) Claim:} 
$\langle I(\texttt{owl:Class}), I(\texttt{owl:Thing})\rangle  
\in 
\mathrm{IEXT}(I(\texttt{rdf:type}))$.

\noindent\emph{Proof:} 
The denotation of $\texttt{owl:Class}$ is in the universe, 
i.e., $\texttt{I(owl:Class)} \in \mathrm{IR}$.
The claim follows from 
$\mathrm{ICEXT}(I(\texttt{owl:Thing})) = \mathrm{IR}$ 
(OWL2/Tab5.2) 
and from the RDFS semantic condition defining ``$\mathrm{ICEXT}$''.

\noindent\emph{2) Claim:} 
$\langle I(\texttt{owl:Class}), I(\texttt{owl:Class})\rangle  
\in 
\mathrm{IEXT}(I(\texttt{rdf:type}))$.

\noindent\emph{Proof:} 
$I(\texttt{owl:Class}) \in \mathrm{IC}$ 
and
$\mathrm{ICEXT}(I(\texttt{owl:Class})) = \mathrm{IC}$ 
(OWL2/Tab5.2).
The claim follows from the RDFS semantic condition defining ``$\mathrm{ICEXT}$''.

\noindent\emph{3) Claim:} 
$\langle I(\texttt{owl:Class}), I(\texttt{owl:Thing})\rangle  
\in \mathrm{IEXT}(I(\texttt{rdfs:subClassOf}))$.

\noindent\emph{Proof:} 
According to 2), 
$I(\texttt{owl:Class}) \in \mathrm{IC}$. 
Further, $I(\texttt{owl:Thing}) \in \mathrm{IC}$ 
according to OWL2/Tab5.2. 
Given arbitrary $x \in \mathrm{ICEXT}(I(\texttt{owl:Class}))$, 
then $x \in \mathrm{IR}$, 
and thus $x \in \mathrm{ICEXT}(I(\texttt{owl:Thing}))$ 
according to OWL2/Tab5.2.
The claim follows from using the ``$\leftarrow$'' direction 
of the OWL~2 semantic condition for class subsumption (OWL2/Tab5.8).

\noindent\emph{4) Claim:} 
$\langle I(\texttt{owl:Class}), I(\texttt{rdfs:Class})\rangle  
\in 
\mathrm{IEXT}(I(\texttt{owl:equivalentClass}))$.

\noindent\emph{Proof:} 
According to 2), we get
$I(\texttt{owl:Class}) \in \mathrm{IC}$. 
According to OWL2/Tab5.2, we get
$I(\texttt{rdfs:Class}) \in \mathrm{IC}$. 
From OWL2/Tab5.2 we get
$\mathrm{ICEXT}(I(\texttt{owl:Class})) 
= \mathrm{IC} 
= \mathrm{ICEXT}(I(\texttt{rdfs:Class}))$. 
The claim follows from using 
the ``$\leftarrow$'' direction of the OWL~2 semantic condition 
for class equivalence
(OWL2/Tab5.9).

\noindent\emph{5) Claim:}
$\langle I(\texttt{rdfs:Datatype}), I(\texttt{owl:Class})\rangle  
\in \mathrm{IEXT}(I(\texttt{rdfs:subClassOf}))$.

\noindent\emph{Proof:} 
According to 2), we get $I(\texttt{owl:Class}) \in \mathrm{IC}$. 
According to OWL2/Tab5.2, we get
$I(\texttt{rdfs:Datatype}) \in \mathrm{IC}$. 
Given arbitrary $x \in \mathrm{ICEXT}(I(\texttt{rdfs:Datatype}))$.
By OWL2/Tab5.2 we get $x \in \mathrm{IDC}$. 
Then, OWL2/Tab5.1 gives $x \in \mathrm{IC}$.
Finally, OWL2/Tab5.2 gives $x \in \mathrm{ICEXT}(I(\texttt{owl:Class}))$. 
The claim now follows
from the ``$\leftarrow$'' direction of the OWL~2 semantic condition 
for class subsumption 
(OWL2/Tab5.8).

\subsubsection{005\_Everything\_is\_a\_Resource (Proof)}
\label{toc:fullishtestsuite:proofs:005}

Let $I$ be an OWL~2 RDF-Based interpretation that satisfies the premise graph.

\noindent\emph{1a) Claim:} 
$I(\texttt{ex:s}) \in \mathrm{IR}$, 
$I(\texttt{ex:p}) \in \mathrm{IR}$, 
$I(\texttt{ex:o}) \in \mathrm{IR}$.

\noindent\emph{Proof:} 
Since $I$ is a simple-interpretation, 
$I(\texttt{ex:s})$ and $I(\texttt{ex:o})$ are in $\mathrm{IR}$, 
and $I(\texttt{ex:p})$ is in $\mathrm{IP}$. 
According to the RDF semantic condition that defines ``$\mathrm{IP}$'',
$\langle I(\texttt{ex:p}), I(\texttt{rdf:Property})\rangle  
\in \mathrm{IEXT}(I(\texttt{rdf:type})$, 
and thus $I(\texttt{ex:p}) \in \mathrm{IR}$.

\noindent\emph{1b) Claim:} 
$\langle I(\texttt{ex:s}), I(\texttt{rdfs:Resource})\rangle  
\in \mathrm{IEXT}(I(\texttt{rdf:type}))$, 
and ditto for $\texttt{ex:p}$ and $\texttt{ex:o}$.

\noindent\emph{Proof:} 
1a) showed $I(\texttt{ex:s}) \in IR$, 
and from the RDFS semantic condition
that defines the class extension of $\texttt{rdfs:Resource}$ 
to be the set $\mathrm{IR}$ follows 
$I(\texttt{ex:s}) \in \mathrm{ICEXT}(I(\texttt{rdfs:Resource}))$. 
The claim follows from the RDFS semantic 
condition that defines ``$\mathrm{ICEXT}$''. 
Analog proofs apply to $\texttt{ex:p}$ and $\texttt{ex:o}$.

\noindent\emph{1c) Claim:} 
$\langle I(\texttt{ex:s}), I(\texttt{owl:Thing})\rangle  
\in \mathrm{IEXT}(I(\texttt{rdf:type}))$, 
and ditto for $\texttt{ex:p}$ and $\texttt{ex:o}$.

\noindent\emph{Proof:} 
As for 1b), but applying the OWL~2 semantic condition that defines
the extension of $\texttt{owl:Thing}$ (OWL2/Tab5.2) 
instead of $\texttt{rdfs:Resource}$.

\noindent\emph{2a) Claim:} 
$I(\texttt{ex:p}) \in \mathrm{IP}$.

\noindent\emph{Proof:} 
This follows directly from $I$ being a simple-interpretation 
that satisfies the premise graph.

\noindent\emph{2b) Claim:} 
$\langle I(\texttt{ex:p}), I(\texttt{rdf:Property})\rangle  
\in \mathrm{IEXT}(I(\texttt{rdf:type}))$.

\noindent\emph{Proof:} 
According to 2a), 
$I(\texttt{ex:p}) \in \mathrm{IP}$, 
and the RDF semantic condition that
defines ``$\mathrm{IP}$'' provides the claim.

\noindent\emph{2c) Claim:} 
$\langle I(\texttt{ex:p}), I(\texttt{owl:ObjectProperty})\rangle  
\in \mathrm{IEXT}(I(\texttt{rdf:type}))$.

\noindent\emph{Proof:} 
According to 2a), 
$I(\texttt{ex:p}) \in \mathrm{IP}$, 
and according to OWL2/Tab5.2
the class extension of $I(\texttt{owl:ObjectProperty})$ is $\mathrm{IP}$. 
The claim follows from the RDFS semantic extension that defines 
``$\mathrm{ICEXT}$''.

\subsubsection{006\_Literal\_Values\_represented\_by\_URIs\_and\_Blank\_Nodes (Proof)}
\label{toc:fullishtestsuite:proofs:006}

Let $I$ be an OWL~2 RDF-Based interpretation and $B$ be a blank node mapping 
for the blank nodes in the premise graph such that $I+B$ satisfies the
premise graph.
Given an~$x$, such that
\[
\begin{array}{rl}
    (1) & \langle I(\texttt{ex:u}), I(\texttt{"abc"})\rangle  \in \mathrm{IEXT}(I(\texttt{owl:sameAs})) \text{, and} \\
    (2) & \langle x, I(\texttt{"abc"})\rangle  \in \mathrm{IEXT}(I(\texttt{owl:sameAs})) \text{, and} \\
    (3) & \langle x, I(\texttt{ex:w})\rangle  \in \mathrm{IEXT}(I(\texttt{owl:sameAs})) \text{.}
\end{array}
\]
By the ``$\rightarrow$'' direction of the semantic condition 
for $\texttt{owl:sameAs}$ (OWL2/Tab5.9), 
we receive that
\[
\begin{array}{rl}
    (1') & I(\texttt{ex:u}) = I(\texttt{"abc"}) \text{, and} \\
    (2') & x = I(\texttt{"abc"}) \text{, and} \\
    (3') & x = I(\texttt{ex:w}) \text{.}
\end{array}
\]
From (2') and (3') we conclude that
\[
\begin{array}{rl}
    (4) & I(\texttt{"abc"}) = I(\texttt{ex:w}) \text{.}
\end{array}
\]
From (1') and (4) we conclude
\[
\begin{array}{rl}
    (5) & I(\texttt{ex:u}) = I(\texttt{ex:w}) \text{.}
\end{array}
\]
From the ``$\leftarrow$'' direction of the semantic condition 
for $\texttt{owl:sameAs}$ (OWL2/Tab5.9),
we conclude
\[
\begin{array}{rl}
    (6) \langle I(\texttt{ex:u}), I(\texttt{ex:w})\rangle  
    \in \mathrm{IEXT}(I(\texttt{owl:sameAs})) \text{.}
\end{array}
\]

\subsubsection{007\_Equal\_Classes (Proof)}
\label{toc:fullishtestsuite:proofs:007}

Let $I$ be an OWL~2 RDF-Based interpretation that satisfies the premise graph.
From the fact
\[
    \langle I(\texttt{ex:c1}), I(\texttt{ex:c2})\rangle  \in \mathrm{IEXT}(I(\texttt{owl:sameAs}))
\]
the ``$\rightarrow$'' direction of the semantic condition 
for $\texttt{owl:sameAs}$ (OWL2/Tab5.9) 
provides
\[
    I(\texttt{ex:c1}) = I(\texttt{ex:c2}) \text{.}
\]
So we can substitute any occurrence 
of $I(\texttt{ex:c1})$ by $I(\texttt{ex:c2})$. 
Hence, from the premises
\[
\begin{array}{l}
    \langle I(\texttt{ex:w}), I(\texttt{ex:c1})\rangle  \in \mathrm{IEXT}(I(\texttt{rdf:type})) \\
    \langle I(\texttt{ex:c}), I(\texttt{ex:c1})\rangle  \in \mathrm{IEXT}(I(\texttt{rdfs:subClassOf})) \\
    \langle I(\texttt{ex:p}), I(\texttt{ex:c1})\rangle  \in \mathrm{IEXT}(I(\texttt{rdfs:range}))
\end{array}
\]
we receive the corresponding conclusions
\[
\begin{array}{l}
    \langle I(\texttt{ex:w}), I(\texttt{ex:c2})\rangle  \in \mathrm{IEXT}(I(\texttt{rdf:type})) \\
    \langle I(\texttt{ex:c}), I(\texttt{ex:c2})\rangle  \in \mathrm{IEXT}(I(\texttt{rdfs:subClassOf})) \\
    \langle I(\texttt{ex:p}), I(\texttt{ex:c2})\rangle  \in \mathrm{IEXT}(I(\texttt{rdfs:range}))
\end{array}
\]

\subsubsection{008\_Inverse\_Functional\_Data\_Properties (Proof)}
\label{toc:fullishtestsuite:proofs:008}

Let $I$ be an OWL~2 RDF-Based interpretation that satisfies the premise graph.
We start from
\[
\begin{array}{rl}
    (1a) & \langle I(\texttt{ex:bob}), I(\texttt{"xyz"})\rangle  \in \mathrm{IEXT}(I(\texttt{foaf:mbox\_sha1sum})) \text{, and} \\
    (1b) & \langle I(\texttt{ex:robert}), I(\texttt{"xyz"})\rangle  \in \mathrm{IEXT}(I(\texttt{foaf:mbox\_sha1sum})) \text{,}
\end{array}
\]
as well as from
\[
\begin{array}{rl}
    (2) & \langle I(\texttt{foaf:mbox\_sha1sum}), I(\texttt{owl:InverseFunctionalProperty})\rangle  \in \mathrm{IEXT}(I(\texttt{rdf:type})) .
\end{array}
\]
From (2) and the ``$\rightarrow$'' direction of the RDFS semantic condition for $\mathrm{ICEXT}$ 
we receive
\[
\begin{array}{rl}
    (2') & I(\texttt{foaf:mbox\_sha1sum}) \in \mathrm{ICEXT}(I(\texttt{owl:InverseFunctionalProperty})) .
\end{array}
\]
This allows to apply the ``$\rightarrow$'' direction of semantic condition for
inverse-functional properties (OWL2/Tab5.13), which provides
\[
\begin{array}{rl}
    (3) & \forall x_1\,, x_2\,, y:\: \\ 
        & \quad \langle x_1, y\rangle  \in \mathrm{IEXT}(I(\texttt{foaf:mbox\_sha1sum})) \;\wedge \\ 
        & \quad \langle x_2, y\rangle  \in \mathrm{IEXT}(I(\texttt{foaf:mbox\_sha1sum})) \\ 
        & \quad \Rightarrow x_1 = x_2
\end{array}
\]
Applying (3) to (1a) and (1b) with
\[
\begin{array}{rcl}
    x_1 & := & I(\texttt{ex:bob}) \text{,} \\
    x_2 & := & I(\texttt{ex:robert}) \text{, and} \\
    y & := & I(\texttt{"xyz"})
\end{array}
\]
results in
\[
\begin{array}{rl}
    (4) & I(\texttt{ex:bob}) = I(\texttt{ex:robert}) .
\end{array}
\]
Using the ``$\leftarrow$'' direction of the semantic condition 
for $\texttt{owl:sameAs}$ (OWL2/Tab5.9) 
results in
\[
\begin{array}{rl}
    (5) & \langle I(\texttt{ex:bob}), I(\texttt{ex:robert})\rangle  \in \mathrm{IEXT}(I(\texttt{owl:sameAs})) \text{.}
\end{array}
\]

\subsubsection{009\_Existential\_Restriction\_Entailments (Proof)}
\label{toc:fullishtestsuite:proofs:009}

Let $I$ be an OWL~2 RDF-Based interpretation and B be a blank node mapping 
for the blank nodes in the premise graph such that $I+B$ satisfies the
premise graph.
Given a~$z$ such that the following holds:
\[
\begin{array}{rl}
    (1a) & \langle I(\texttt{ex:p}), I(\texttt{owl:ObjectProperty})\rangle  
           \in \mathrm{IEXT}(I(\texttt{rdf:type})) \text{ ,} \\
    (1b) & \langle I(\texttt{ex:c}), I(\texttt{owl:Class})\rangle  
           \in \mathrm{IEXT}(I(\texttt{rdf:type})) \text{ ,} \\
    (1c) & \langle I(\texttt{ex:s}), z\rangle  
           \in \mathrm{IEXT}(I(\texttt{rdf:type})) \text{ ,} \\
    (1d) & \langle z, I(\texttt{owl:Restriction})\rangle  
           \in \mathrm{IEXT}(I(\texttt{rdf:type})) \text{ ,} \\
    (1e) & \langle z, I(\texttt{ex:p})\rangle  
           \in \mathrm{IEXT}(I(\texttt{owl:onProperty})) \text{ ,} \\
    (1f) & \langle z, I(\texttt{ex:c})\rangle  
           \in \mathrm{IEXT}(I(\texttt{owl:someValuesFrom})) \text{ .}
\end{array}
\]
From the RDFS semantic condition for $\mathrm{ICEXT}$ 
(``$\rightarrow$'' direction) and (1c) 
follows
    (2) $I(\texttt{ex:s}) \in \mathrm{ICEXT}(z)$ .
From the semantic condition of $\texttt{owl:someValuesFrom}$ (OWL2/Tab5.6), 
(1e) and (1f) follows
\[
\begin{array}{rll}
    (3) & \forall y:\: & y \in \mathrm{ICEXT}(z)
          \Leftrightarrow \\
        & & \exists x:\: 
            \left[\, \langle y, x\rangle  \in \mathrm{IEXT}(I(\texttt{ex:p})) 
            \;\wedge\; 
            x \in \mathrm{ICEXT}(I(\texttt{ex:c})) 
            \,\right] \text{ .}
\end{array}
\]
From (2) and (3) follows
\[
\begin{array}{rl}
    (4) & \exists x:\: 
          \langle I(\texttt{ex:s}), x\rangle  \in \mathrm{IEXT}(I(\texttt{ex:p})) 
          \;\wedge\;
          x \in \mathrm{ICEXT}(I(\texttt{ex:c})) 
          \text{ .}
\end{array}
\]
By the RDFS semantic condition for $\mathrm{ICEXT}$ 
(``$\leftarrow$'' direction) and (4) we receive
\[
\begin{array}{rl}
    (5) & \exists x:\: 
          \langle I(\texttt{ex:s}), x\rangle  \in \mathrm{IEXT}(I(\texttt{ex:p})) 
          \;\wedge\;
          \langle x, I(\texttt{ex:c})\rangle  \in \mathrm{IEXT}(I(\texttt{rdf:type})) 
          \text{ .}
\end{array}
\]

\subsubsection{010\_Negative\_Property\_Assertions (Proof)}
\label{toc:fullishtestsuite:proofs:010}

Let $I$ be an OWL~2 RDF-Based interpretation and $B$ be a blank node mapping 
for the blank nodes in the premise graph such that $I+B$ satisfies the
premise graph.
Let $x_1$, $x_2$, $x_3$ and $x_4$ be individuals, such that the following holds
\[
\begin{array}{rl}
    (1a) & \langle I(\texttt{ex:p}), I(\texttt{owl:ObjectProperty})\rangle  
           \in \mathrm{IEXT}(I(\texttt{rdf:type})) \text{ ,} \\ 
    (1b) & \langle I(\texttt{ex:s}), x_1\rangle  
           \in \mathrm{IEXT}(I(\texttt{rdf:type})) \text{ ,} \\
    (1c) & \langle x_1, I(\texttt{ex:p})\rangle  
           \in \mathrm{IEXT}(I(\texttt{owl:onProperty})) \text{ ,} \\
    (1d) & \langle x_1, x_2\rangle  
           \in \mathrm{IEXT}(I(\texttt{owl:allValuesFrom})) \text{ ,} \\
    (1e) & \langle x_2, x_3\rangle  
           \in \mathrm{IEXT}(I(\texttt{owl:complementOf})) \text{ ,} \\
    (1f) & \langle x_3, x_4\rangle  
           \in \mathrm{IEXT}(I(\texttt{owl:oneOf})) \text{ ,} \\
    (1g) & \langle x_4, I(\texttt{ex:o})\rangle  
           \in \mathrm{IEXT}(I(\texttt{rdf:first})) \text{ ,} \\
    (1h) & \langle x_4, I(\texttt{rdf:nil})\rangle  
           \in \mathrm{IEXT}(I(\texttt{rdf:rest})) \text{ .}
\end{array}
\]
From the RDFS semantic condition for $\textrm{ICEXT}$ 
(``$\rightarrow$'' direction) and (1b) follows
\[
\begin{array}{rl}
    (2b) & I(\texttt{ex:s}) \in \mathrm{ICEXT}(x_1) \text{ .}
\end{array}
\]
From the semantic condition for $\texttt{owl:allValuesFrom}$ (OWL2/Tab5.6), (1c) and (1d) 
follows
\[
\begin{array}{rl}
    (2c) & \forall y:\: y \in \mathrm{ICEXT}(x_1) \Leftrightarrow \\ 
         & \forall z:\: \left[ 
           \langle y, z\rangle  \in \mathrm{IEXT}(I(\texttt{ex:p})) 
           \Rightarrow 
           z \in \mathrm{ICEXT}(x_2) 
           \right] \text{ .}
\end{array}
\]
From the ``$\rightarrow$'' direction of the semantic condition 
for class complement (OWL2/Tab5.4) and (1e) follows
\[
\begin{array}{rl}
    (2e) & \forall y:\: y \in \mathrm{ICEXT}(x_2)
           \Leftrightarrow 
           y \notin \mathrm{ICEXT}(x_3) \text{ .}
\end{array}
\]
From the ``$\rightarrow$'' direction of the semantic condition 
for singleton enumerations
(OWL2/Tab5.5) and (1f), (1g) and (1h) follows
\[
\begin{array}{rl}
    (2f) & \forall y:\: 
           y \in \mathrm{ICEXT}(x_3) 
           \Leftrightarrow 
           y = I(\texttt{ex:o}) \text{ .}
\end{array}
\]
Assume that
\[
\begin{array}{rl}
    (3) & \langle I(\texttt{ex:s}), I(\texttt{ex:o})\rangle  
          \in IEXT(I(\texttt{ex:p})) \text{ .}
\end{array}
\]
By (2b), (2c) and (3) we receive
\[
\begin{array}{rl}
    (4) & I(\texttt{ex:o}) \in \mathrm{ICEXT}(x_2) \text{ .}
\end{array}
\]
By (4) and (2e) follows
\[
\begin{array}{rl}
    (5) & I(\texttt{ex:o}) \notin \mathrm{ICEXT}(x_3) \text{ .}
\end{array}
\]
By (5) and (2f) follows
\[
\begin{array}{rl}
    (6) & I(\texttt{ex:o}) \neq I(\texttt{ex:o}) \text{ .}
\end{array}
\]
This is a contradiction, hence assumption (3) was wrong. So we have:
\[
\begin{array}{rl}
    (3') & \langle I(\texttt{ex:s}), I(\texttt{ex:o})\rangle  
           \notin \mathrm{IEXT}(I(\texttt{ex:p})) \text{ .}
\end{array}
\]
Since I is a simple-interpretation, we receive
\[
\begin{array}{rl}
    (7a) & I(\texttt{ex:s}) \in \mathrm{IR} \text{ ,} \\
    (7b) & I(\texttt{ex:o}) \in \mathrm{IR} \text{ .}
\end{array}
\]
By the property extension of $\texttt{owl:onProperty}$ (OWL2/Tab5.3) follows
\[
\begin{array}{rl}
    (7c) & I(\texttt{ex:p}) \in \mathrm{IP} \text{ .}
\end{array}
\]
From the second semantic condition for NPAs in OWL2/Tab5.15, (3'), 
(7a), (7b) and (7c) follows that there exists some~$z$, such that
\[
\begin{array}{rl}
    (8a) & \langle z, I(\texttt{ex:s})\rangle  
           \in \mathrm{IEXT}(I(\texttt{owl:sourceIndividual})) \text{ ,} \\
    (8b) & \langle z, I(\texttt{ex:p})\rangle  
           \in \mathrm{IEXT}(I(\texttt{owl:assertionProperty})) \text{ ,} \\
    (8c) & \langle z, I(\texttt{ex:o})\rangle  
           \in \mathrm{IEXT}(I(\texttt{owl:targetIndividual})) \text{ .}
\end{array}
\]
Finally, from the property extension of $\texttt{owl:sourceIndividual}$ and (8a) follows
\[
\begin{array}{rl}
    (9) & z \in \mathrm{ICEXT}(I(\texttt{owl:NegativePropertyAssertion})) \text{ ,}
\end{array}
\]
which by the ``$\leftarrow$'' direction of the RDFS semantic condition for $\mathrm{ICEXT}$ results in
\[
\begin{array}{rl}
    (9') & \langle z, I(\texttt{owl:NegativePropertyAssertion})\rangle  
           \in \mathrm{IEXT}(I(\texttt{rdf:type})) \text{ .}
\end{array}
\]
The result consists of (9), (8a), (8b) and~(8c).

\subsubsection{011\_Entity\_Types\_as\_Classes (Proof)}
\label{toc:fullishtestsuite:proofs:011}

Let $I$ be an OWL~2 RDF-Based interpretation that satisfies the premise graph.
We start from the facts:
\[
\begin{array}{rl}
    (1a) & \langle I(\texttt{owl:Class}), I(\texttt{owl:ObjectProperty})) 
           \in \mathrm{IEXT}(I(\texttt{owl:disjointWith})) \text{ ,} \\
    (1b) & \langle I(\texttt{ex:x}), I(\texttt{owl:Class})\rangle  
           \in \mathrm{IEXT}(I(\texttt{rdf:type})) \text{ ,} \\
    (1c) & \langle I(\texttt{ex:x}), I(\texttt{owl:ObjectProperty})\rangle  
           \in \mathrm{IEXT}(I(\texttt{rdf:type})) \text{ .}
\end{array}
\]
First, by the ``$\rightarrow$'' direction of RDFS semantic condition 
of $\mathrm{ICEXT}$, we rewrite (1b) and (1c) to
\[
\begin{array}{rl}
    (1b') & I(\texttt{ex:x}) \in \mathrm{ICEXT}(I(\texttt{owl:Class})) \text{ ,} \\
    (1c') & I(\texttt{ex:x}) \in \mathrm{ICEXT}(I(\texttt{owl:ObjectProperty})) \text{ .}
\end{array}
\]
By the ``$\rightarrow$'' direction of the semantic condition 
for class disjointness 
(OWL2/Tab5.9) and (1a) follows
\[
\begin{array}{rl}
    (2) & \forall z: \neg \left[ 
            z \in \mathrm{ICEXT}(I(\texttt{owl:Class})) \;\wedge\; 
            z \in \mathrm{ICEXT}(I(\texttt{owl:ObjectProperty})) \right] \text{ .}
\end{array}
\]
Specialization (2) to $z := I(\texttt{ex:x})$ implies
\[
\begin{array}{rl}
    (3) & \neg \left[ 
            I(\texttt{ex:x}) \in \mathrm{ICEXT}(I(\texttt{owl:Class})) \;\wedge\;
            I(\texttt{ex:x}) \in \mathrm{ICEXT}(\texttt{owl:ObjectProperty})) \right] \text{ .}
\end{array}
\]
Now (3) is in contradiction with (1b') and (1c').

\subsubsection{012\_Template\_Class (Proof)}
\label{toc:fullishtestsuite:proofs:012}

Let $I$ be an OWL~2 RDF-Based interpretation and $B$ be a blank node mapping 
for the blank nodes in the premise graph such that $I+B$ satisfies the
premise graph.
Given the existence of individuals $l_1$, $l_2$, $l_3$ and $r$, such that
\[
\begin{array}{rl}
    (1a1) & \langle I(\texttt{foaf:Person}), I(\texttt{owl:Class})\rangle  
            \in \mathrm{IEXT}(I(\texttt{rdf:type})) \text{ ,} \\
    (1b1) & \langle I(\texttt{ex:PersonAttribute}), l_1\rangle  
            \in \mathrm{IEXT}(I(\texttt{owl:intersectionOf})) \text{ ,} \\
    (1c1) & \langle l_1, I(\texttt{owl:DatatypeProperty})\rangle  
            \in \mathrm{IEXT}(I(\texttt{rdf:first})) \text{ ,} \\
    (1c2) & \langle l_1, l_2\rangle  
            \in \mathrm{IEXT}(I(\texttt{rdf:rest})) \text{ ,} \\
    (1c3) & \langle l_2, I(\texttt{owl:FunctionalProperty})\rangle  
            \in \mathrm{IEXT}(I(\texttt{rdf:first})) \text{ ,} \\
    (1c4) & \langle l_2, l_3\rangle  
            \in \mathrm{IEXT}(I(\texttt{rdf:rest})) \text{ ,} \\
    (1c5) & \langle l_3, r\rangle  
            \in \mathrm{IEXT}(I(\texttt{rdf:first})) \text{ ,} \\
    (1c6) & \langle l_3, I(\texttt{rdf:nil})\rangle  
            \in \mathrm{IEXT}(I(\texttt{rdf:rest})) \text{ ,} \\
    (1d1) & \langle r, I(\texttt{owl:Restriction})\rangle  
            \in \mathrm{IEXT}(I(\texttt{rdf:type})) \text{ ,} \\
    (1d2) & \langle r, I(\texttt{rdfs:domain})\rangle  
            \in \mathrm{IEXT}(I(\texttt{owl:onProperty})) \text{ ,} \\
    (1d3) & \langle r, I(\texttt{foaf:Person})\rangle  
            \in \mathrm{IEXT}(I(\texttt{owl:hasValue})) \text{ ,} \\
    (1e1) & \langle I(\texttt{ex:name}), I(\texttt{ex:PersonAttribute})\rangle  
            \in \mathrm{IEXT}(I(\texttt{rdf:type})) \text{ ,} \\
    (1f1) & \langle I(\texttt{ex:alice}), I(\texttt{"alice"})\rangle  
            \in \mathrm{IEXT}(I(\texttt{ex:name})) \text{ .}
\end{array}
\]
From (1b1), (1c1) -- (1c6), and the semantic condition for class intersection
(OWL2/Tab5.4, ``$\rightarrow$'', ternary) follows
\[
\begin{array}{rll}
    (2) & \forall x:\: & x \in \mathrm{ICEXT}(I(\texttt{ex:PersonAttribute})) 
            \Leftrightarrow \\
        & &   x \in \mathrm{ICEXT}(I(\texttt{owl:DatatypeProperty})) \;\wedge\; \\ 
        & &   x \in \mathrm{ICEXT}(I(\texttt{owl:FunctionalProperty})) \;\wedge\; \\
        & &   x \in \mathrm{ICEXT}(r) \text{ .}
\end{array}
\]
From (1e1) and the RDFS semantic condition for $\mathrm{ICEXT}$ (``$\rightarrow$'') follows
\[
\begin{array}{rl}
    (3) & I(\texttt{ex:name}) 
          \in \mathrm{ICEXT}(I(\texttt{ex:PersonAttribute})) \text{ .}
\end{array}
\]
From (3) and (2) follows
\[
\begin{array}{rl}
    (4) & I(\texttt{ex:name}) 
          \in \mathrm{ICEXT}(I(\texttt{owl:FunctionalProperty})) \text{ .}
\end{array}
\]
From (4) and the RDFS semantic condition for $\mathrm{ICEXT}$ (``$\leftarrow$'') follows
\[
\begin{array}{rl}
    (5) & \langle I(\texttt{ex:name}), I(\texttt{owl:FunctionalProperty})\rangle  
          \in \mathrm{IEXT}(I(\texttt{rdf:type})) \text{ .}
\end{array}
\]
From (1d2), (1d3) and the semantic condition for has-value restrictions
(OWL2/Tab5.6) follows
\[
\begin{array}{rll}
    (6) & \forall x: & x \in ICEXT(r) \Leftrightarrow \\
        & &  \langle x, I(\texttt{foaf:Person})\rangle  
            \in \mathrm{IEXT}(I(\texttt{rdfs:domain})) \text{ .}
\end{array}
\]
From (3) and (2) follows
\[
\begin{array}{rl}
    (7) & I(\texttt{ex:name}) \in \mathrm{ICEXT}(r) \text{ .}
\end{array}
\]
From (7) and (6) follows
\[
\begin{array}{rl}
    (8) & \langle I(\texttt{ex:name}), I(\texttt{foaf:Person})\rangle  
          \in \mathrm{IEXT}(I(\texttt{rdfs:domain})) \text{ .}
\end{array}
\]
From (1f1), (8) and the RDFS semantic condition for property domains follows
\[
\begin{array}{rl}
    (9) & I(\texttt{ex:alice}) 
          \in \mathrm{ICEXT}(I(\texttt{foaf:Person})) \text{ .}
\end{array}
\]
From (9) and the RDFS semantic condition for $\mathrm{ICEXT}$ (``$\leftarrow$'') follows
\[
\begin{array}{rl}
    (10) & \langle I(\texttt{ex:alice}), I(\texttt{foaf:Person})\rangle  
           \in \mathrm{IEXT}(I(\texttt{rdf:type})) \text{ .}
\end{array}
\]
The conjecture follows from (5) and (10).

\subsubsection{013\_Cliques (Proof)}
\label{toc:fullishtestsuite:proofs:013}

Let $I$ be an OWL~2 RDF-Based interpretation and $B$ be a blank node mapping 
for the blank nodes in the premise graph such that $I+B$ satisfies the
premise graph.
Let there be $r$, $i$, $l_1$, $l_2$ and~$l_3$ such that
\[
\begin{array}{rl}
    (1a) & \langle I(\texttt{ex:Clique}), I(\texttt{owl:Class})\rangle  \in \mathrm{IEXT}(I(\texttt{rdf:type})) \text{ ,} \\
    (1b) & \langle I(\texttt{ex:sameCliqueAs}), I(\texttt{owl:sameAs})\rangle  \in \mathrm{IEXT}(I(\texttt{rdfs:subPropertyOf})) \text{ ,} \\
    (1c) & \langle I(\texttt{ex:sameCliqueAs}), I(\texttt{ex:Clique})\rangle  \in \mathrm{IEXT}(I(\texttt{rdfs:range})) \text{ ,} \\
    (1d) & \langle I(\texttt{ex:Clique}), r\rangle  \in \mathrm{IEXT}(I(\texttt{rdfs:subClassOf})) \text{ ,} \\
    (1e) & \langle r, I(\texttt{owl:Restriction})\rangle  \in \mathrm{IEXT}(I(\texttt{rdf:type})) \text{ ,} \\
    (1f) & \langle r, I(\texttt{ex:sameCliqueAs})\rangle  \in \mathrm{IEXT}(I(\texttt{owl:onProperty})) \text{ ,} \\
    (1g) & \langle r, I(\texttt{ex:Clique})\rangle  \in \mathrm{IEXT}(I(\texttt{owl:someValuesFrom})) \text{ ,} \\
    (1h) & \langle I(\texttt{foaf:knows}), I(\texttt{owl:ObjectProperty})\rangle  \in \mathrm{IEXT}(I(\texttt{rdf:type})) \text{ ,} \\
\end{array}
\]
From (1s) and the RDFS semantic condition of ``$\mathrm{ICEXT}$'' 
(``$\rightarrow$'') follows
\[
\begin{array}{rl}
    (1s') & I(\texttt{ex:JoesGang}) 
    \in \mathrm{ICEXT}(I(\texttt{ex:Clique})) \text{ .}
\end{array}
\]
From (1d) and the OWL~2 semantic condition of class subsumption 
(OWL2/Tab5.8, ``$\rightarrow$'') follows
\[
\begin{array}{rl}
    (2) & \forall x:\: 
          x \in \mathrm{ICEXT}(I(\texttt{ex:Clique})) 
          \Rightarrow 
          x \in \mathrm{ICEXT}(r) \text{ .}
\end{array}
\]
From (1f) -- (1g) and the semantic condition for existential property 
restrictions (OWL2/Tab5.6) follows
\[
\begin{array}{rll}
    (3) & \forall x:\: & x \in \mathrm{ICEXT}(r) \Leftrightarrow \\ 
        & & \exists y:\: \langle x, y\rangle  \in \mathrm{IEXT}(I(\texttt{ex:sameCliqueAs})) 
                          \;\wedge\; y \in \mathrm{ICEXT}(I(\texttt{ex:Clique})) \text{ .}
\end{array}
\]
From (1s'), (2) and (3) follows
\[
\begin{array}{rl}
    (4) & \exists y: \langle I(\texttt{ex:JoesGang}), y\rangle  \in \mathrm{IEXT}(I(\texttt{ex:sameCliqueAs})) \text{ .}
\end{array}
\]
According to (4), we can find some~$y$ such that
\[
\begin{array}{rl}
    (5) & \langle I(\texttt{ex:JoesGang}), y\rangle  \in \mathrm{IEXT}(I(\texttt{ex:sameCliqueAs})) \text{ .}
\end{array}
\]
By applying the OWL~2 semantic condition for property subsumption 
OWL2/Tab5.8, ``$\leftrightarrow$'') and (1b) to (5), we receive
\[
\begin{array}{rl}
    (6) & \langle I(\texttt{ex:JoesGang}), y\rangle  \in \mathrm{IEXT}(I(\texttt{ex:sameAs})) \text{ .}
\end{array}
\]
Now, the semantic condition for $\texttt{owl:sameAs}$ 
(OWL2/Tab5.9, ``$\rightarrow$'') applied to (6) which yields
\[
\begin{array}{rl}
    (6) & y = I(\texttt{ex:JoesGang}) \text{ .}
\end{array}
\]
By (4) and (6) we receive
\[
\begin{array}{rl}
    (7) & \langle I(\texttt{ex:JoesGang}), I(\texttt{ex:JoesGang})\rangle  
          \in \mathrm{IEXT}(I(\texttt{ex:sameCliqueAs})) \text{ .}
\end{array}
\]
From (1r) and the semantic condition for inverse properties 
(OWL2/Tab5.12, ``$\rightarrow$'') follows
\[
\begin{array}{rl}
    (8) & \forall x\,\,y:\: \langle x, y\rangle  \in \mathrm{IEXT}(i) 
          \Leftrightarrow 
          \langle y, x\rangle  \in \mathrm{IEXT}(I(\texttt{rdf:type})) \text{ .}
\end{array}
\]
From (1u) and (8) follows:
\[
\begin{array}{rl}
    (9) & \langle I(\texttt{ex:JoesGang}), I(\texttt{ex:bob})\rangle  
          \in \mathrm{IEXT}(i) \text{ .}
\end{array}
\]
From (1j) -- (1q) and the semantic condition for sub property chains 
(OWL2/Tab5.11, ``$\rightarrow$'', ternary) we receive
\[
\begin{array}{rl}
    (10) & \forall y_0,\,y_1,\,y_2,\,y_3:\: \\ 
         & \quad \langle y_0, y_1\rangle  \in \mathrm{IEXT}(I(\texttt{rdf:type})) \;\wedge\; \\
         & \quad \langle y_1, y_2\rangle  \in \mathrm{IEXT}(I(\texttt{ex:sameCliqueAs})) \;\wedge\; \\
         & \quad \langle y_2, y_3\rangle  \in \mathrm{IEXT}(i) \\ 
         & \quad \Rightarrow
             \langle y_0, y_3\rangle  \in \mathrm{IEXT}(I(\texttt{foaf:knows})) \text{ .}
\end{array}
\]
Finally, from (10), (1t), (7) and (9) follows
\[
\begin{array}{rl}
    (11) & \langle I(\texttt{ex:alice}), I(\texttt{ex:bob})\rangle  
           \in \mathrm{IEXT}(I(\texttt{foaf:knows})) \text{ .}
\end{array}
\]

\subsubsection{014\_Harry\_belongs\_to\_some\_Species (Proof)}
\label{toc:fullishtestsuite:proofs:014}

Let $I$ be an OWL~2 RDF-Based interpretation and $B$ be a blank node mapping 
for the blank nodes in the premise graph such that $I+B$ satisfies the
premise graph.
Let $u$, $l_1$ and~$l_2$ be individuals such that the following holds:
\[
\begin{array}{rl}
    (1a) & \langle I(\texttt{ex:Eagle}), I(\texttt{ex:Species})\rangle  
           \in \mathrm{IEXT}(I(\texttt{rdf:type})) \text{ ,} \\
    (1b) & \langle I(\texttt{ex:Falcon}), I(\texttt{ex:Species})\rangle  
           \in \mathrm{IEXT}(I(\texttt{rdf:type})) \text{ ,} \\
    (1c) & \langle I(\texttt{ex:harry}), u\rangle  
           \in \mathrm{IEXT}(I(\texttt{rdf:type})) \text{ ,} \\
    (1d1) & \langle u, l_1\rangle  
            \in \mathrm{IEXT}(I(\texttt{owl:unionOf})) \text{ ,} \\
    (1d2) & \langle l_1, I(\texttt{ex:Eagle})\rangle  
            \in \mathrm{IEXT}(I(\texttt{rdf:first})) \text{ ,} \\
    (1d3) & \langle l_1, l_2\rangle  
            \in \mathrm{IEXT}(I(\texttt{rdf:rest})) \text{ ,} \\
    (1d4) & \langle l_2, I(\texttt{ex:Falcon})\rangle  
            \in \mathrm{IEXT}(I(\texttt{rdf:first})) \text{ ,} \\
    (1d5) & \langle l_2, I(\texttt{rdf:nil})\rangle  
            \in \mathrm{IEXT}(I(\texttt{rdf:rest})) \text{ .}
\end{array}
\]
We prove the claim by classical dilemma.

\noindent\emph{Case 1:} 
$\langle I(\texttt{ex:harry}), I(\texttt{ex:Eagle})\rangle  
 \in \mathrm{IEXT}(I(\texttt{rdf:type}))$ .

Then, together with (1a) follows:
\[
\begin{array}{rll}
    (2) & \exists x:\: & 
            \langle I(\texttt{ex:harry}), x\rangle  
            \in \mathrm{IEXT}(I(\texttt{rdf:type})) \;\wedge\; \\ 
        & & \langle x, I(\texttt{ex:Species})\rangle  
            \in \mathrm{IEXT}(I(\texttt{rdf:type})) \text{ .}
\end{array}
\]

\noindent\emph{Case 2:} 
$\langle I(\texttt{ex:harry}), I(\texttt{ex:Eagle})\rangle  
 \notin \mathrm{IEXT}(I(\texttt{rdf:type}))$ .

By the RDFS semantic condition of $\mathrm{ICEXT}$ 
(``$\rightarrow$'') we get
\[
\begin{array}{rl}
    (3) & I(\texttt{ex:harry}) 
          \notin \mathrm{ICEXT}(I(\texttt{ex:Eagle})) \text{ .}
\end{array}
\]
Likewise, from (1c) and the RDFS semantic condition 
of $\mathrm{ICEXT}$ (``$\rightarrow$'') we get
\[
\begin{array}{rl}
    (4) & I(\texttt{ex:harry}) \in \mathrm{ICEXT}(u) \text{ .}
\end{array}
\]
From (1d1) -- (1d5) and the semantic condition of class union 
(OWL2/Tab5.4, ``$\rightarrow$'', binary) follows
\[
\begin{array}{rll}
    (5) & \forall x:\: & x \in \mathrm{ICEXT}(u) 
          \Leftrightarrow \\ 
        & & x \in \mathrm{ICEXT}(I(\texttt{ex:Eagle})) 
            \vee 
            x \in \mathrm{ICEXT}(I(\texttt{ex:Falcon})) \text{ .}
\end{array}
\]
Specializing (4) to $x := I(\texttt{ex:harry})$ implies
\[
\begin{array}{rl}
    (6) & I(\texttt{ex:harry}) \in \mathrm{ICEXT}(u) 
          \Leftrightarrow \\ 
        & I(\texttt{ex:harry}) \in \mathrm{ICEXT}(\texttt{ex:Eagle}) 
          \vee 
          I(\texttt{ex:harry}) \in \mathrm{ICEXT}(\texttt{ex:Falcon}) \text{ .}
\end{array}
\]
By (4), (6) and (3) we get
\[
\begin{array}{rl}
    (7) & I(\texttt{ex:harry}) \in \mathrm{ICEXT}(I(\texttt{ex:Falcon})) \text{ .}
\end{array}
\]
Using the RDFS semantic condition of $\mathrm{ICEXT}$ (``$\leftarrow$'') 
results in
\[
\begin{array}{rl}
    (8) & \langle I(\texttt{ex:harry}), I(\texttt{ex:Falcon})\rangle  
          \in \mathrm{IEXT}(I(\texttt{rdf:type})) \text{ .}
\end{array}
\]
From (8) and (1b) follows:
\[
\begin{array}{rll}
    (9) & \exists x:\: &  
            \langle I(\texttt{ex:harry}), x\rangle  
            \in \mathrm{IEXT}(I(\texttt{rdf:type})) 
            \;\wedge\; \\
        & & \langle x, I(\texttt{ex:Species})\rangle  
            \in \mathrm{IEXT}(I(\texttt{rdf:type})) \text{ .}
\end{array}
\]

Since (2) and~(9) are the same result from the contrary assumed cases, 
we get the claimed result.

\subsubsection{015\_Reflective\_Tautologies\_I (Proof)}
\label{toc:fullishtestsuite:proofs:015}

Let $I$ be a satisfying OWL~2 RDF-Based interpretation for the empty graph.
It is true that
\[
    I(\texttt{owl:sameAs}) = I(\texttt{owl:sameAs}) \text{ .}
\]
By the ``$\leftarrow$'' direction of semantic condition for 
$\texttt{owl:sameAs}$ (OWL2/Tab5.9)
we receive
\[
    \langle I(\texttt{owl:sameAs}), I(\texttt{owl:sameAs})\rangle  
    \in \mathrm{IEXT}(I(\texttt{owl:sameAs})) \text{ .}
\]

\subsubsection{016\_Reflective\_Tautologies\_II (Proof)}
\label{toc:fullishtestsuite:proofs:016}

Let $I$ be a satisfying OWL~2 RDF-Based interpretation for the empty graph.
Given arbitrary $c_1$, $c_2$, and assume the following to hold:
\[
\begin{array}{rl}
    (1) & \langle c_1, c_2\rangle  \in \mathrm{IEXT}(I(\texttt{owl:equivalentClass})) \text{ .}
\end{array}
\]
From the ``$\rightarrow$'' direction 
of the semantic condition for class equivalence (OWL2/Tab5.9) 
and from the property extension of $\texttt{owl:equivalentClass}$ 
(OWL2/Tab5.3) follows
\[
\begin{array}{rl}
    (2a) & c_1 \in \mathrm{IC} \text{ ,}\\
    (2b) & c_2 \in \mathrm{IC} \text{ ,}\\
    (2c) & \forall x:\: 
           x \in \mathrm{ICEXT}(c_1) 
           \Leftrightarrow 
           x \in \mathrm{ICEXT}(c_2)
           \text{ .}
\end{array}
\]
From (2c) follows the weaker result
\[
\begin{array}{rl}
    (3) & \forall x:\: 
          x \in \mathrm{ICEXT}(c_1) 
          \Rightarrow 
          x \in \mathrm{ICEXT}(c_2) 
          \text{ .}
\end{array}
\]
From (2a), (2b) and (3) and the ``$\leftarrow$'' direction 
of the OWL 2 semantic condition of class subsumption (OWL2/Tab5.8) 
follows
\[
\begin{array}{rl}
    (4) & \langle c_1, c_2\rangle  \in \mathrm{IEXT}(I(\texttt{rdfs:subClassOf})) \text{ .}
\end{array}
\]
Since (4) follows from (1) and since $c_1$ and~$c_2$ were chosen arbitrarily, 
we get
\[
\begin{array}{rll}
    (5) & \forall x,\,y:\: & 
            \langle x, y\rangle  \in \mathrm{IEXT}(I(\texttt{owl:equivalentClass})) \\ 
        & & \Rightarrow \langle x, y\rangle  \in \mathrm{IEXT}(I(\texttt{rdfs:subClassOf}))
            \text{ .}
\end{array}
\]
For property $\texttt{owl:equivalentClass}$ we receive from OWL2/Tab5.3
\[
\begin{array}{rl}
    (6a) & I(\texttt{owl:equivalentClass}) \in \mathrm{IP} \text{ .}
\end{array}
\]
and for property $\texttt{rdfs:subClassOf}$ 
we receive from the RDFS axiomatic triples
\[
\begin{array}{rl}
    (6b) & \langle I(\texttt{rdfs:subClassOf}), I(\texttt{rdf:Property})\rangle  
    \in \mathrm{IEXT}(I(\texttt{rdf:type})) \text{ .}
\end{array}
\]
By the ``$\leftarrow$'' direction of the RDF semantic condition 
of $\mathrm{IP}$ and $\mathrm{IEXT}$ 
we receive from (6b):
\[
\begin{array}{rl}
    (6b') & I(\texttt{rdfs:subClassOf}) \in \mathrm{IP} \text{ .}
\end{array}
\]
From (5), (6a) and (6b') and from the ``$\leftarrow$'' direction 
of the OWL 2 semantic condition for property subsumption (OWL2/Tab5.8) 
we finally receive
\[
\begin{array}{rl}
    (7) & \langle I(\texttt{owl:equivalentClass}), I(\texttt{rdfs:subClassOf})\rangle  
    \in \mathrm{IEXT}(I(\texttt{rdfs:subPropertyOf})) \text{ .}
\end{array}
\]

\subsubsection{017\_Builtin\_Based\_Definitions (Proof)}
\label{toc:fullishtestsuite:proofs:017}

Let $I$ be an OWL~2 RDF-Based interpretation that satisfies the premise graph.
Let the following assertions hold:
\[
\begin{array}{cl}
    (1a) & \langle I(\texttt{ex:notInstanceOf}), I(\texttt{rdf:type})\rangle  
         \in \mathrm{IEXT}(I(\texttt{owl:propertyDisjointWith})) \text{ ,} \\
    (1b) & \langle I(\texttt{ex:w}), I(\texttt{ex:c})\rangle  
         \in \mathrm{IEXT}(I(\texttt{rdf:type})) \text{ ,} \\
    (1c) & \langle I(\texttt{ex:u}), I(\texttt{ex:c})\rangle  
         \in \mathrm{IEXT}(I(\texttt{ex:notInstanceOf})) \text{ .}
\end{array}
\]
From (1a) and the semantic condition for 
disjoint properties (OWL2/5.9, ``$\rightarrow$'') 
follows:
\[
\begin{array}{cll}
    (1a') & \forall x,\,y: \neg [\, &
            \langle x, y\rangle  \in \mathrm{IEXT}(I(\texttt{ex:notInstanceOf})) \;\wedge\; \\ 
          & & \langle x, y\rangle  \in \mathrm{IEXT}(I(\texttt{rdf:type})) \,] \text{ .}
\end{array}
\]
Specializing (1a') to $x := I(\texttt{ex:w})$ 
and $y := I(\texttt{ex:c})$ implies
\[
\begin{array}{cll}
    (2) & \neg [\, & 
            \langle I(\texttt{ex:w}), I(\texttt{ex:c})\rangle  
            \in \mathrm{IEXT}(I(\texttt{ex:notInstanceOf})) \;\wedge\; \\ 
        & & \langle I(\texttt{ex:w}), I(\texttt{ex:c})\rangle  
            \in \mathrm{IEXT}(I(\texttt{rdf:type})) \,] \text{ .}
\end{array}
\]
From (1b) and (2) follows
\[
\begin{array}{cl}
    (3) & \langle I(\texttt{ex:w}), I(\texttt{ex:c})\rangle  
          \notin \mathrm{IEXT}(I(\texttt{ex:notInstanceOf})) \text{ .}
\end{array}
\]
From (1c) and~(3) follows
\[
\begin{array}{cl}
    (4) & I(\texttt{ex:w}) \neq I(\texttt{ex:u}) \text{ .}
\end{array}
\]
From (4) and the semantic condition for 
$\texttt{owl:differentFrom}$ (OWL2/Tab5.9, ``$\leftarrow$'') 
follows:
\[
\begin{array}{cl}
    (5)  \langle I(\texttt{ex:w}), I(\texttt{ex:u})\rangle  
         \in \mathrm{IEXT}(I(\texttt{owl:differentFrom})) \text{ .}
\end{array}
\]

\subsubsection{018\_Modified\_Logical\_Vocabulary\_Semantics (Proof)}
\label{toc:fullishtestsuite:proofs:018}

Let $I$ be an OWL~2 RDF-Based interpretation that satisfies the premise graph.
We start from:
\[
\begin{array}{rl}
    (1a) & \langle I(\texttt{owl:sameAs}), I(\texttt{ex:Person})\rangle  
           \in \mathrm{IEXT}(I(\texttt{rdfs:domain})) \text{ ,} \\
    (1b) & \langle I(\texttt{ex:w}), I(\texttt{ex:u})\rangle  
           \in \mathrm{IEXT}(I(\texttt{owl:sameAs})) \text{ .}
\end{array}
\]
From this we get via the RDFS semantic condition for property domains:
\[
\begin{array}{rl}
    (2) & I(\texttt{ex:w}) \in \mathrm{ICEXT}(I(\texttt{ex:Person})) \text{ .}
\end{array}
\]
Further, from (1b) and the semantic condition for 
$\texttt{owl:sameAs}$ (OWL2/Tab5.9, ``$\rightarrow$'') 
we get
\[
\begin{array}{rl}
    (3) & I(\texttt{ex:w}) = I(\texttt{ex:u}) \text{ .}
\end{array}
\]
This allows for substitution in (2), providing
\[
\begin{array}{rl}
    (4) & I(\texttt{ex:u}) \in \mathrm{ICEXT}(I(\texttt{ex:Person})) \text{ .}
\end{array}
\]
Finally, by applying the RDFS semantic extension 
for ICEXT (``$\leftarrow$'') to (4) we get
\[
\begin{array}{rl}
    (5) & \langle I(\texttt{ex:u}), I(\texttt{ex:Person})\rangle  
          \in \mathrm{IEXT}(I(\texttt{rdf:type})) \text{ .}
\end{array}
\]

\subsubsection{019\_Disjoint\_Annotation\_Properties (Proof)}
\label{toc:fullishtestsuite:proofs:019}

Let $I$ be an OWL~2 RDF-Based interpretation that satisfies the premise graph.
Starting from:
\[
\begin{array}{rl}
    (1a) & \langle I(\texttt{skos:prefLabel}), I(\texttt{skos:altLabel})\rangle  
           \in \mathrm{IEXT}(I(\texttt{owl:propertyDisjointWith})) \text{ ,} \\
    (1b) & \langle I(\texttt{ex:foo}), I(\texttt{"foo"})\rangle  
           \in \mathrm{IEXT}(I(\texttt{skos:prefLabel})) \text{ ,} \\
    (1c) & \langle I(\texttt{ex:foo}), I(\texttt{"foo"})\rangle  
           \in \mathrm{IEXT}(I(\texttt{skos:altLabel})) \text{ .}
\end{array}
\]
From (1a) and the semantic condition of property disjointness 
(OWL2/Tab5.9, ``$\rightarrow$'') 
we receive
\[
\begin{array}{rll}
    (2) & \forall x,\,y:\: \neg [\, &
            \langle x, y\rangle  \in \mathrm{IEXT}(I(\texttt{skos:prefLabel})) \;\wedge\; \\ 
        & & \langle x, y\rangle  \in \mathrm{IEXT}(I(\texttt{skos:altLabel})) 
            \,] \text{ .}
\end{array}
\]
Specifically, we receive:
\[
\begin{array}{rll}
    (3) & \neg [ \; & 
            \langle I(\texttt{ex:foo}), I(\texttt{"foo"})\rangle  
            \in \mathrm{IEXT}(I(\texttt{skos:prefLabel})) \;\wedge\; \\ 
        & & \langle I(\texttt{ex:foo}), I(\texttt{"foo"})\rangle  
            \in \mathrm{IEXT}(I(\texttt{skos:altLabel})) 
            \,] \text{ .}
\end{array}
\]
We now have a contradiction between (1b), (1c) and~(3).

\subsubsection{020\_Logical\_Complications (Proof)}
\label{toc:fullishtestsuite:proofs:020}

Let $I$ be an OWL 2 RDF-Based interpretation and $B$ be a blank node mapping 
for the blank nodes in the premise graph such that $I+B$ satisfies the
premise graph.
Let there be $xs$, $xc$, $lu_1$, $lu_2$, $lu_3$, $li_1$, $li_2$, $li_3$, such that
\[
\begin{array}{rl}
    (1a) & \langle I(\texttt{ex:c}), lu_1\rangle  \in \mathrm{IEXT}(I(\texttt{owl:unionOf})) \text{ ,} \\
    (1b) & \langle lu_1, I(\texttt{ex:c1})\rangle  \in \mathrm{IEXT}(I(\texttt{rdf:first})) \text{ ,} \\
    (1c) & \langle lu_1, lu_2\rangle  \in \mathrm{IEXT}(I(\texttt{rdf:rest})) \text{ ,} \\
    (1d) & \langle lu_2, I(\texttt{ex:c2})\rangle  \in \mathrm{IEXT}(I(\texttt{rdf:first})) \text{ ,} \\
    (1e) & \langle lu_2, lu_3\rangle  \in \mathrm{IEXT}(I(\texttt{rdf:rest})) \text{ ,} \\
    (1f) & \langle lu_3, I(\texttt{ex:c3})\rangle  \in \mathrm{IEXT}(I(\texttt{rdf:first})) \text{ ,} \\
    (1g) & \langle lu_3, I(\texttt{rdf:nil})\rangle  \in \mathrm{IEXT}(I(\texttt{rdf:rest})) \text{ ,} \\
    (1h) & \langle I(\texttt{ex:d}), I(\texttt{ex:c1})\rangle  \in \mathrm{IEXT}(I(\texttt{owl:disjointWith})) \text{ ,} \\
    (1j) & \langle I(\texttt{ex:d}), xs\rangle  \in \mathrm{IEXT}(I(\texttt{rdfs:subClassOf})) \text{ ,} \\
    (1k) & \langle xs, li_1\rangle  \in \mathrm{IEXT}(I(\texttt{owl:intersectionOf})) \text{ ,} \\
    (1m) & \langle li_1, I(\texttt{ex:c})\rangle  \in \mathrm{IEXT}(I(\texttt{rdf:first})) \text{ ,} \\
    (1n) & \langle li_1, li_2\rangle  \in \mathrm{IEXT}(I(\texttt{rdf:rest})) \text{ ,} \\
    (1o) & \langle li_2, xc\rangle  \in \mathrm{IEXT}(I(\texttt{rdf:first})) \text{ ,} \\
    (1p) & \langle li_2, I(\texttt{rdf:nil})\rangle  \in \mathrm{IEXT}(I(\texttt{rdf:rest})) \text{ ,} \\
    (1q) & \langle xc, I(\texttt{ex:c2})\rangle  \in \mathrm{IEXT}(I(\texttt{owl:complementOf})) \text{ .}
\end{array}
\]
From (1a), (1b) -- (1g) and the semantic condition for class union
(OWL2/Tab.5.4, ``$\rightarrow$'', ternary) follows
\[
\begin{array}{rl}
    (2) & I(\texttt{ex:c3}) \in \mathrm{IC} \text{ .}
\end{array}
\]
and
\[
\begin{array}{rll}
    (3) & \forall x:\: & x \in \mathrm{ICEXT}(I(\texttt{ex:c})) 
          \Leftrightarrow \\
        & & x \in \mathrm{ICEXT}(I(\texttt{ex:c1})) \;\vee\;
            x \in \mathrm{ICEXT}(I(\texttt{ex:c2})) \;\vee\;
            x \in \mathrm{ICEXT}(I(\texttt{ex:c3})) \text{ .}
\end{array}
\]
From (1h) and the semantic condition for class disjointness follows
\[
\begin{array}{rl}
    (4) & I(\texttt{ex:d}) \in \mathrm{IC}
\end{array}
\]
and from (1h) and the semantic condition for class disjointness 
(OWL2/Tab.5.9, ``$\rightarrow$'') follows
\[
\begin{array}{rl}
    (5) & \forall x:\: \neg [\, 
          x \in \mathrm{ICEXT}(I(\texttt{ex:d})) \;\wedge\; 
          x \in \mathrm{ICEXT}(I(\texttt{ex:c1})) \,] \text{ .}
\end{array}
\]
From (1j) and the OWL 2 semantic condition of class subsumption 
(OWL2/Tab5.9, ``$\rightarrow$'') follows
\[
\begin{array}{rl}
    (6) & \forall x:\: 
          x \in \mathrm{ICEXT}(I(\texttt{ex:d})) 
          \Rightarrow 
          x \in \mathrm{ICEXT}(xs) \text{ .}
\end{array}
\]
From (1k), (1m) -- (1p) and the semantic condition for class intersection
(OWL2/Tab5.4, ``$\rightarrow$'', binary) follows
\[
\begin{array}{rl}
    (7) & \forall x:\: 
          x \in \mathrm{ICEXT}(xs) 
          \Leftrightarrow 
          x \in \mathrm{ICEXT}(I(\texttt{ex:c})) \;\wedge\; 
          x \in \mathrm{ICEXT}(xc) \text{ .}
\end{array}
\]
From (1q) and the semantic condition for class complement 
(OWL2/Tab5.4, ``$\rightarrow$'') follows
\[
\begin{array}{rl}
    (8) & \forall x:\: 
          x \in ICEXT(xc) 
          \Leftrightarrow 
          x \notin \mathrm{ICEXT}(I(\texttt{ex:c2})) \text{ .}
\end{array}
\]
From (6), (7) and~(8) follows
\[
\begin{array}{rl}
    (9) & \forall x:\: 
          x \in \mathrm{ICEXT}(I(\texttt{ex:d}))
          \Rightarrow 
          x \in \mathrm{ICEXT}(I(\texttt{ex:c}))
          \;\wedge\; 
          x \notin \mathrm{ICEXT}(I(\texttt{ex:c2})) \text{ .}
\end{array}
\]
From (9) and (3) follows
\[
\begin{array}{rl}
    (10) & \forall x:\: 
           x \in \mathrm{ICEXT}(I(\texttt{ex:d})) 
           \Rightarrow 
           x \in \mathrm{ICEXT}(I(\texttt{ex:c1})) 
           \;\vee\; 
           x \in \mathrm{ICEXT}(I(\texttt{ex:c3})) \text{ .}
\end{array}
\]
From (10) and (5) follows
\[
\begin{array}{rl}
    (11) & \forall x:\: 
           x \in \mathrm{ICEXT}(I(\texttt{ex:d})) 
           \Rightarrow 
           x \in \mathrm{ICEXT}(I(\texttt{ex:c3})) \text{ .}
\end{array}
\]
Finally, from (4), (2), (11) and the OWL~2 semantic extension of
class subsumption (OWL2/Tab5.8, ``$\leftarrow$'') follows
\[
\begin{array}{rl}
    (12) & \langle I(\texttt{ex:d}), I(\texttt{ex:c3})\rangle  
           \in \mathrm{IEXT}(I(\texttt{rdfs:subClassOf})) \text{ .}
\end{array}
\]

\subsubsection{021\_Composite\_Enumerations (Proof)}
\label{toc:fullishtestsuite:proofs:021}

Let $I$ be an OWL~2 RDF-Based interpretation and $B$ be a blank node mapping 
for the blank nodes in the premise graph such that $I+B$ satisfies the
premise graph.
Let there be $l_{11}$, $l_{12}$, $l_{21}$, $l_{22}$, $l_{31}$, $l_{32}$, $l_{41}$, $l_{42}$, such that the following statements hold:
\[
\begin{array}{rl}
    (1a1) & \langle I(\texttt{ex:c1}), l_{11}\rangle  \in \mathrm{IEXT}(I(\texttt{owl:oneOf})) \text{ ,} \\
    (1a2) & \langle l_{11}, I(\texttt{ex:w1})\rangle  \in \mathrm{IEXT}(I(\texttt{rdf:first})) \text{ ,} \\
    (1a3) & \langle l_{11}, l_{12})\rangle  \in \mathrm{IEXT}(I(\texttt{rdf:rest})) \text{ ,} \\
    (1a4) & \langle l_{12}, I(\texttt{ex:w2})\rangle  \in \mathrm{IEXT}(I(\texttt{rdf:first})) \text{ ,} \\
    (1a5) & \langle l_{12}, I(\texttt{rdf:nil})\rangle  \in \mathrm{IEXT}(I(\texttt{rdf:rest})) \text{ ,} \\
    (1b1) & \langle I(\texttt{ex:c2}), l_{21})\rangle  \in \mathrm{IEXT}(I(\texttt{owl:oneOf})) \text{ ,} \\
    (1b2) & \langle l_{21}, I(\texttt{ex:w2})\rangle  \in \mathrm{IEXT}(I(\texttt{rdf:first})) \text{ ,} \\
    (1b3) & \langle l_{21}, l_{22})\rangle  \in \mathrm{IEXT}(I(\texttt{rdf:rest})) \text{ ,} \\
\end{array}
\]
By applying the semantic condition for enumeration classes 
(OWL2/Tab5.5, ``$\rightarrow$'', binary and ternary) 
to (1a1)--(1a5), (1b1)--(1b5) 
and (1c1)--(1c7), respectively,
we receive:
\[
\begin{array}{rl}
    (2a) & \forall x:\: x \in \mathrm{ICEXT}(I(\texttt{ex:c1})) \Leftrightarrow x = I(\texttt{ex:w1}) \vee x = I(\texttt{ex:w2}) \text{ ,} \\
    (2b) & \forall x:\: x \in \mathrm{ICEXT}(I(\texttt{ex:c2})) \Leftrightarrow x = I(\texttt{ex:w2}) \vee x = I(\texttt{ex:w3}) \text{ ,} \\
    (2c) & \forall x:\: x \in \mathrm{ICEXT}(I(\texttt{ex:c3})) \Leftrightarrow x = I(\texttt{ex:w1}) \vee x = I(\texttt{ex:w2}) \vee x = I(\texttt{ex:w3}) \text{ .}
\end{array}
\]
By applying the semantic condition for class union 
(OWL2/Tab5.4, ``$\rightarrow$'', binary) 
to (1d1)--(1d5), we receive:
\[
\begin{array}{rl}
    (2d) & \forall x:\: x \in \mathrm{ICEXT}(\texttt{ex:c4}) \Leftrightarrow x \in \mathrm{ICEXT}(I(\texttt{ex:c1})) \vee x \in \mathrm{ICEXT}(I(\texttt{ex:c2})) \text{ .}
\end{array}
\]
From the property extension of $\texttt{owl:oneOf}$ (OWL2/Tab5.3) and (1c1) follows
\[
\begin{array}{rl}
    (3a) & I(\texttt{ex:c3}) \in \mathrm{IC} \text{ .}
\end{array}
\]
From the property extension of $\texttt{owl:unionOf}$ (OWL2/Tab5.3) and (1d1) follows
\[
\begin{array}{rl}
    (3b) & I(\texttt{ex:c4}) \in \mathrm{IC} \text{ .}
\end{array}
\]
From (2a), (2b) and (2c) follows
\[
\begin{array}{rl}
    (4) & \forall x:\: x \in \mathrm{ICEXT}(I(\texttt{ex:c3})) \Leftrightarrow 
            x \in \mathrm{ICEXT}(I(\texttt{ex:c1})) \vee x \in \mathrm{ICEXT}(I(\texttt{ex:c2})) \text{ .}
\end{array}
\]
From (2d) and (4) follows
\[
\begin{array}{rl}
    (5) & \forall x:\: x \in \mathrm{ICEXT}(I(\texttt{ex:c3})) \Leftrightarrow x \in \mathrm{ICEXT}(I(\texttt{ex:c4})) \text{ .}
\end{array}
\]
From the semantic condition for class equivalence 
(OWL2/Tab5.9, $\leftarrow$), 
(3a), (3b) and (5) follows
\[
\begin{array}{rl}
    (6) & \langle I(\texttt{ex:c3}), I(\texttt{ex:c4})\rangle  \in \mathrm{IEXT}(I(\texttt{owl:equivalentClass})) .
\end{array}
\]

\subsubsection{022\_List\_Member\_Access (Proof)}
\label{toc:fullishtestsuite:proofs:022}

Let $I$ be an OWL~2 RDF-Based interpretation and $B$ be a blank node mapping 
for the blank nodes in the premise graph such that $I+B$ satisfies the
premise graph.
Given an individual~$pL$
as well as list individuals $l_{11}$, $l_{12}$, $l_{21}$, $l_{22}$, $l_{31}$, $l_{32}$ and~$l_{33}$,
such that the following assertions hold:
\[
\begin{array}{rl}
    (1a1) & \langle I(\texttt{skos:memberList}), pL\rangle  \in \mathrm{IEXT}(I(\texttt{rdfs:subPropertyOf})) \text{ ,} \\
    (1b1) & \langle I(\texttt{skos:member}), l_{11}\rangle  \in \mathrm{IEXT}(I(\texttt{owl:propertyChainAxiom})) \text{ ,} \\
    (1b2) & \langle l_{11}, pL\rangle  \in \mathrm{IEXT}(I(\texttt{rdf:first})) \text{ ,} \\
    (1b3) & \langle l_{11}, l_{12}\rangle  \in \mathrm{IEXT}(I(\texttt{rdf:rest})) \text{ ,} \\
    (1b4) & \langle l_{12}, I(\texttt{rdf:first})\rangle  \in \mathrm{IEXT}(I(\texttt{rdf:first})) \text{ ,} \\
    (1b5) & \langle l_{12}, I(\texttt{rdf:nil})\rangle  \in \mathrm{IEXT}(I(\texttt{rdf:rest})) \text{ ,} \\
    (1c1) & \langle pL, l_{21}\rangle  \in \mathrm{IEXT}(I(\texttt{owl:propertyChainAxiom})) \text{ ,} \\
    (1c2) & \langle l_{21}, pL\rangle  \in \mathrm{IEXT}(I(\texttt{rdf:first})) \text{ ,} \\
\end{array}
\]
By the RDFS semantic condition for property subsumption, (1a1) and (1e1) 
we receive
\[
\begin{array}{rl}
    (1e1') & \langle I(\texttt{ex:MyOrderedCollection}), l_{31}\rangle  \in \mathrm{IEXT}(pL) \text{ .}
\end{array}
\]
From the semantic condition of sub property chains 
(OWL2/Tab5.11, ``$\rightarrow$'', binary version) we get from (1b1) to (1b5):
\[
\begin{array}{rll}
    (2b) & \forall y_0,\,y_1,\,y_2:\: &
              \langle y_0, y_1\rangle  \in \mathrm{IEXT}(pL) \;\wedge\; \\
         & &  \langle y_1, y_2\rangle  \in \mathrm{IEXT}(I(\texttt{rdf:first})) \\ 
         & & \Rightarrow \langle y_0, y_2\rangle  \in \mathrm{IEXT}(I(\texttt{skos:member})) \text{ .} 
\end{array}
\]
and from (1c1) to (1c5) we get
\[
\begin{array}{rll}
    (2c) & \forall y_0,\,y_1,\,y_2:\: & 
             \langle y_0, y_1\rangle  \in \mathrm{IEXT}(pL) \;\wedge\; \\
         & & \langle y_1, y_2\rangle  \in \mathrm{IEXT}(I(\texttt{rdf:rest})) \\ 
         & & \Rightarrow \langle y_0, y_2\rangle  \in \mathrm{IEXT}(pL) \text{ .}
\end{array}
\]
We receive the first result triple
\[
\begin{array}{rl}
    (3a) & \langle I(\texttt{ex:MyOrderedCollection}), I(\texttt{ex:X})\rangle  
           \in \mathrm{IEXT}(I(\texttt{skos:member})) \text{ .}
\end{array}
\]
from (2b), (1e1') and (1e2).
Further, by (2c), (1e1') and (1e3) we receive
\[
\begin{array}{rl}
    (1e1'') & \langle I(\texttt{ex:MyOrderedCollection}), l_{32}\rangle  
              \in \mathrm{IEXT}(pL) \text{ .}
\end{array}
\]
In the same way, from (2b), (1e1''), (1e4) and (1e5) we receive
\[
\begin{array}{rl}
    (3b) & \langle I(\texttt{ex:MyOrderedCollection}), I(\texttt{ex:Y})\rangle  
           \in \mathrm{IEXT}(I(\texttt{skos:member})) \text{ .}
\end{array}
\]
and
\[
\begin{array}{rl}
    (1e1''') & \langle I(\texttt{ex:MyOrderedCollection}), l_{33}\rangle  
               \in \mathrm{IEXT}(pL) \text{ .}
\end{array}
\]
And likewise, we receive
\[
\begin{array}{rl}
    (3c) & \langle I(\texttt{ex:MyOrderedCollection}), I(\texttt{ex:Z})\rangle  
           \in \mathrm{IEXT}(I(\texttt{skos:member})) \text{ .}
\end{array}
\]

\subsubsection{023\_Unique\_List\_Components (Proof)}
\label{toc:fullishtestsuite:proofs:023}

Let $I$ be an OWL~2 RDF-Based interpretation and $B$ be a blank node mapping 
for the blank nodes in the premise graph such that $I+B$ satisfies the
premise graph.
Let there be individuals $o$ and~$l$ such that
\[
\begin{array}{rl}
    (1a) & \langle I(\texttt{rdf:first}), I(\texttt{owl:FunctionalProperty})\rangle  \in \mathrm{IEXT}(I(\texttt{rdf:type})) \text{ ,} \\
    (1b) & \langle I(\texttt{ex:w}), o\rangle  \in \mathrm{IEXT}(I(\texttt{rdf:type})) \text{ ,} \\
    (1c) & \langle o, I(\texttt{owl:Class})\rangle  \in \mathrm{IEXT}(I(\texttt{rdf:type})) \text{ ,} \\
    (1d) & \langle o, l\rangle  \in \mathrm{IEXT}(I(\texttt{owl:oneOf})) \text{ ,} \\
    (1e) & \langle l, I(\texttt{ex:u})\rangle  \in \mathrm{IEXT}(I(\texttt{rdf:first})) \text{ ,} \\
    (1f) & \langle l, I(\texttt{ex:v})\rangle  \in \mathrm{IEXT}(I(\texttt{rdf:first})) \text{ ,} \\
    (1g) & \langle l, I(\texttt{rdf:nil})\rangle  \in \mathrm{IEXT}(I(\texttt{rdf:rest})) \text{ .}
\end{array}
\]
By the RDFS semantic condition for $\mathrm{ICEXT}$ and (1b) follows
\[
\begin{array}{rl}
    (1b') & I(\texttt{ex:w}) \in \mathrm{ICEXT}(o) \text{ .}
\end{array}
\]
By the semantic condition for enumeration classes 
(OWL2/Tab5.5, ``$\rightarrow$'', singleton) 
and (1d), (1e) and (1g) follows
\[
\begin{array}{rl}
    (2) & \forall x:\: x \in \mathrm{ICEXT}(o) \Leftrightarrow x = I(\texttt{ex:u}) \text{ .}
\end{array}
\]
From (1b') and specializing (2) to $x := I(\texttt{ex:w})$ follows
\[
\begin{array}{rl}
    (3) & I(\texttt{ex:w}) = I(\texttt{ex:u}) \text{ .}
\end{array}
\]
By the semantic condition for functional properties 
(OWL2/Tab5.13, ``$\rightarrow$'') 
and (1a) follows
\[
\begin{array}{rll}
    (4) & \forall x,\, y1,\, y2:\: &
            \langle x, y1\rangle  \in \mathrm{IEXT}(I(\texttt{rdf:first})) \;\wedge\; \\
        & & \langle x, y2\rangle  \in \mathrm{IEXT}(I(\texttt{rdf:first})) \\ 
        & &  \Rightarrow y1 = y2 \text{ .}
\end{array}
\]
From (1e), (1f) and from specializing (4) to $x := l$ follows
\[
\begin{array}{rl}
    (5) & I(\texttt{ex:u}) = I(\texttt{ex:v}) \text{ .}
\end{array}
\]
From the semantic condition of $\texttt{owl:sameAs}$ 
(OWL2/Tab5.9, ``$\leftarrow$'') and (3) and (5) 
follows
\[
\begin{array}{rl}
    (6a) & \langle I(\texttt{ex:w}), I(\texttt{ex:u})\rangle  \in \mathrm{IEXT}(I(\texttt{owl:sameAs})) \text{ ,} \\
    (6b) & \langle I(\texttt{ex:w}), I(\texttt{ex:v})\rangle  \in \mathrm{IEXT}(I(\texttt{owl:sameAs})) \text{ .}
\end{array}
\]

\subsubsection{024\_Cardinality\_Restrictions\_on\_Complex\_Properties (Proof)}
\label{toc:fullishtestsuite:proofs:024}

Let $I$ be an OWL~2 RDF-Based interpretation and $B$ be a blank node mapping 
for the blank nodes in the premise graph such that $I+B$ satisfies the
premise graph.
Let there be z such that
\[
\begin{array}{rl}
    (1a) & \langle I(\texttt{ex:hasAncestor}), I(\texttt{owl:TransitiveProperty})\rangle  \in \mathrm{IEXT}(I(\texttt{rdf:type})) \text{ ,} \\
    (1b) & \langle I(\texttt{ex:Person}), z\rangle  \in \mathrm{IEXT}(I(\texttt{rdfs:subClassOf})) \text{ ,} \\
    (1c) & \langle z, I(\texttt{owl:Restriction})\rangle  \in \mathrm{IEXT}(I(\texttt{rdf:type})) \text{ ,} \\
    (1d) & \langle z, I(\texttt{ex:hasAncestor})\rangle  \in \mathrm{IEXT}(I(\texttt{owl:onProperty})) \text{ ,} \\
    (1e) & \langle z, I(\texttt{"1"\^{}\^{}xsd:nonNegativeInteger})\rangle  \in \mathrm{IEXT}(I(\texttt{owl:minCardinality})) \text{ ,} \\
    (1f) & \langle I(\texttt{ex:alice}), I(\texttt{ex:Person})\rangle  \in \mathrm{IEXT}(I(\texttt{rdf:type})) \text{ ,} \\
    (1g) & \langle I(\texttt{ex:bob}), I(\texttt{ex:Person})\rangle  \in \mathrm{IEXT}(I(\texttt{rdf:type})) \text{ ,} \\
    (1h) & \langle I(\texttt{ex:alice}), I(\texttt{ex:bob})\rangle  \in \mathrm{IEXT}(I(\texttt{ex:hasAncestor})) \text{ .}
\end{array}
\]
From (1a) and the semantic condition of 
transitive properties (OWL2/Tab5.13, ``$\rightarrow$'') 
follows
\[
\begin{array}{rll}
    (2) & \forall y_1,\, y_2,\, y_3:\: & 
            \langle y_1, y_2\rangle  \in \mathrm{IEXT}(I(\texttt{ex:hasAncestor})) \;\wedge\; \\ 
        & & \langle y_2, y_3\rangle  \in \mathrm{IEXT}(I(\texttt{ex:hasAncestor})) \\ 
        & & \Rightarrow \langle y_1, y_3\rangle  \in \mathrm{IEXT}(I(\texttt{ex:hasAncestor})) \text{ .}
\end{array}
\]
From (1b) and the OWL~2 semantic condition for class subsumption 
(OWL2/Tab5.8, ``$\rightarrow$'') follows
\[
\begin{array}{rl}
    (3) & \forall y:\: y \in \mathrm{ICEXT}(I(\texttt{ex:Person})) \Rightarrow y \in \mathrm{ICEXT}(z) \text{ .}
\end{array}
\]
From (1c)--(1e) and the semantic condition for 
\mbox{1-min} cardinality restrictions 
(OWL2/Tab5.6) follows
\[
\begin{array}{rl}
    (4) & \forall y:\: y \in \mathrm{ICEXT}(z) 
            \Leftrightarrow 
            \exists x:\: \langle y, x\rangle  \in \mathrm{ICEXT}(I(\texttt{ex:hasAncestor})) \text{ .}
\end{array}
\]
From (3) and (4) follows
\[
\begin{array}{rl}
    (5) & \forall y:\: \mathrm{ICEXT}(\texttt{ex:Person}, y) \Rightarrow 
            \exists x:\: \langle y, x\rangle  \in \mathrm{IEXT}(\texttt{ex:hasAncestor}) \text{ .}
\end{array}
\]
Applying the RDFS semantic condition for $\mathrm{ICEXT}$ (``$\rightarrow$'') to (1g) yields
\[
\begin{array}{rl}
    (6) & I(\texttt{ex:bob}) \in \mathrm{ICEXT}(I(\texttt{ex:Person})) \text{ .}
\end{array}
\]
From (6) and (5) follows the existence of some~$x$ such that
\[
\begin{array}{rl}
    (7) & \langle I(\texttt{ex:bob}), x\rangle  \in \mathrm{IEXT}(I(\texttt{ex:hasAncestor})) \text{ .}
\end{array}
\]
From (1h), (7) and (2) follows, for the same~$x$,
\[
\begin{array}{rl}
    (8) & \langle I(\texttt{ex:alice}), x\rangle  \in \mathrm{IEXT}(I(\texttt{ex:hasAncestor})) \text{ .}
\end{array}
\]
Hence we have shown in (7) and (8) that
\[
\begin{array}{rll}
    (9) & \exists x:\: &
            \langle I(\texttt{ex:bob}), x\rangle  \in \mathrm{IEXT}(I(\texttt{ex:hasAncestor})) \;\wedge\; \\
        & & \langle I(\texttt{ex:alice}), x\rangle  \in \mathrm{IEXT}(I(\texttt{ex:hasAncestor})) \text{ .}
\end{array}
\]
Therefore, there is some blank node mapping~$B'$ for the blank nodes in the
conclusion graph such that $I+B'$ satisfies the conclusion graph.

\subsubsection{025\_Cyclic\_Dependencies\_between\_Complex\_Properties (Proof)}
\label{toc:fullishtestsuite:proofs:025}

Let $I$ be an OWL~2 RDF-Based interpretation and $B$ be a blank node mapping 
for the blank nodes in the premise graph such that $I+B$ satisfies the
premise graph.
Let there be individuals $l_{11}$, $l_{12}$, $l_{21}$, $l_{22}$, $l_{3}$, such that:
\[

\] 
From the semantic condition for sub property chains (OWL2/Tab5.11, ``$\rightarrow$'', binary) 
and (1a1) to (1a5) follows
\[
\begin{array}{rll}
    (2a) & \forall y_{0},\,y_{1},\,y_{2}:\: &
             \langle y_{0}, y_{1}\rangle  \in \mathrm{IEXT}(I(\texttt{ex:hasCousin})) \;\wedge\; \\ 
         & & \langle y_{1}, y_{2}\rangle  \in \mathrm{IEXT}(I(\texttt{ex:hasFather})) \\ 
         & & \Rightarrow \langle y_{0}, y_{2}\rangle  \in \mathrm{IEXT}(I(\texttt{ex:hasUncle})) \text{ .}
\end{array}
\]
From the semantic condition for sub property chains (OWL2/Tab5.11, ``$\rightarrow$'', binary) 
and (1b1) to (1b5) follows
\[
\begin{array}{rll}
    (2b) & \forall y_{0},\,y_{1},\,y_{2}:\: &
             \langle y_{0}, y_{1}\rangle  \in \mathrm{IEXT}(I(\texttt{ex:hasUncle})) \;\wedge\; \\
         & & \langle y_{1}, y_{2}\rangle  \in \mathrm{IEXT}(l_{3}) \\
         & & \Rightarrow \langle y_{0}, y_{2}\rangle  \in \mathrm{IEXT}(I(\texttt{ex:hasCousin})) \text{ .}
\end{array}
\] 
From the semantic condition for inverse properties (OWL2/Tab5.12, ``$\rightarrow$'') 
and (1c) follows
\[
\begin{array}{rl}
    (2c) & \forall x,\, y:\: \langle x,y\rangle  \in \mathrm{IEXT}(l_{3}) 
         \Leftrightarrow 
         \langle y, x\rangle  \in \mathrm{IEXT}(I(\texttt{ex:hasFather})) \text{ .}
\end{array}
\] 
From (2c) and (1d) follows
\[
\begin{array}{rl}
    (1d') & \langle I(\texttt{ex:dave}), I(\texttt{ex:alice})\rangle  \in \mathrm{IEXT}(l_{3}) \text{ .}
\end{array}%
\] 
The resulting triples are (3a) and (3b).

\subsubsection{026\_Inferred\_Property\_Characteristics\_I (Proof)}
\label{toc:fullishtestsuite:proofs:026}

Let $I$ be an OWL~2 RDF-Based interpretation and $B$ be a blank node mapping 
for the blank nodes in the premise graph such that $I+B$ satisfies the
premise graph.
Let $x_{1}$, $x_{2}$, $l_{1}$ and~$l_{2}$ be individuals such that the following holds:
\[
%
\]
From the semantic condition for 
enumeration classes (OWL2/Tab5.5, ``$\rightarrow$'', singleton) 
and (1b) to (1d) follows:
\[
\begin{array}{rl}
    (2) & \forall z:\: 
        z \in \mathrm{ICEXT}(x_{1}) 
        \Leftrightarrow 
        z = I(\texttt{ex:w}) \text{ .}
\end{array}
\]
Now let us assume that there are $s_{1}$, $s_{2}$ and~$o$ such that
\[
%
\]
From the RDFS semantic condition for property domains, (1a) and (3a) follows
\[
%
\]
From the RDFS semantic condition for property domains, (1a) and (3b) follows
\[
%
\]
Since $s_{1}$, $s_{2}$ and~$o$ were chosen arbitrarily, we can generalize
\[
%
\]
From the RDFS axiomatic triple for the domain of $\texttt{rdfs:domain}$, we receive
\[
%
\]
From the RDFS semantic condition for property domains, (8) and (1a) follows
\[
%
\]
With the RDFS semantic condition for $\mathrm{ICEXT}$ and the RDF semantic condition 
for $\mathrm{IP}$ and $\mathrm{IEXT}$ follows
\[
%
\]
By the semantic condition for inverse-functional properties 
(OWL2/Tab5.13, ``$\leftarrow$''), 
(9') and (7) follows
\[
%
\]
And by the RDFS semantic condition for $\mathrm{ICEXT}$ and (10) follows
\[
%
\]

\subsubsection{027\_Inferred\_Property\_Characteristics\_II (Proof)}
\label{toc:fullishtestsuite:proofs:027}

Let $I$ be an OWL~2 RDF-Based interpretation and $B$ be a blank node mapping 
for the blank nodes in the premise graph such that $I+B$ satisfies the
premise graph.
Let there be $l_{1}$, $l_{2}$ and~$v$, such that
\[
%
\]
From (1a), (1b) -- (1e) and the semantic condition for sub property chains
(OWL2/Tab5.11, ``$\rightarrow$'', binary) follows
\[
%
\]
From (1f) and the semantic condition for 
inverse properties (OWL2/Tab5.12, ``$\rightarrow$'') 
follows:
\[
%
\]
From the semantic condition for $\texttt{owl:sameAs}$ 
(OWL2/Tab5.9, ``$\rightarrow$'') follows:
\[
%
\]
From (2), (6) and the semantic condition for inverse-functional properties
(OWL2/Tab.13, ``$\leftarrow$'') follows
\[
%
\]
Finally, from (7) and the RDFS semantic condition 
for $\mathrm{\mathrm{ICEXT}}$ (``$\leftarrow$'') follows
\[
%
\]

\subsubsection{028\_Inferred\_Property\_Characteristics\_III (Proof)}
\label{toc:fullishtestsuite:proofs:028}

Let $I$ be an OWL~2 RDF-Based interpretation and $B$ be a blank node mapping 
for the blank nodes in the premise graph such that $I+B$ satisfies the
premise graph.
Let there be a~$z$ such that
\[
%
\]
From (1a) and the semantic condition for 
class equivalence (OWL2/Tab5.9, ``$\rightarrow$'') 
follows
\[
%
\]
From (1b) -- (1d) and the semantic condition for existential property 
restrictions (OWL2/Tab5.6) follows
\[
%
\]
Let $p$ be an arbitrary individual such that
\[
%
\]
From (7) and the semantic condition for 
inverse properties (OWL2/Tab5.12, ``$\rightarrow$'') 
follows
\[
%
\]
From (7) and the semantic condition for functional properties 
(OWL2/Tab5.13, ``$\rightarrow$'') follows:
\[
%
\]
From (7) and the property extension of 
$\texttt{owl:inverseOf}$ (OWL2/Tab.5.3) follows
\[
%
\]
From (11), (10) and the semantic condition for inverse-functional properties
(OWL2/Tab5.13, ``$\leftarrow$'') follows
\[
%
\]
Since (12) follows from (5) and since $p$ has been arbitrarily chosen, we receive
\[
%
\]
From (1a) and the property extension of $\texttt{owl:equivalentClass}$ (OWL2/Tab5.3) 
follows
\[
%
\]
Finally, from (14), (15), (13) and the OWL~2 semantic condition for
class subsumption (OWL2/Tab5.8, ``$\leftarrow$'') follows
\[
%
\]

\subsubsection{029\_Ex\_Falso\_Quodlibet (Proof)}
\label{toc:fullishtestsuite:proofs:029}

Let $I$ be an OWL~2 RDF-Based interpretation and $B$ be a blank node mapping 
for the blank nodes in the premise graph such that $I+B$ satisfies the
premise graph.
Let $x$, $y$, $l_1$, $l_2$ be individuals such that
\[
%
\]
From (1d1), (1e1) -- (1e4), and the semantic condition for class intersection
(OWL2/Tab5.4, ``$\rightarrow$'', binary) follows
\[
%
\]
From (1f1) and the semantic condition for class complement 
(OWL2/Tab5.4, ``$\rightarrow$'') 
follows
\[
%
\]
From (1c1) and the RDFS semantic condition of $\mathrm{ICEXT}$ (``$\rightarrow$'') follows
\[
%
\]
From (5) and (4) follows a contradiction, i.e.\ the set of premises is 
contradictory.
From a contradiction follows arbitrary 
(\emph{``ex falso sequitur quodlibet''}), 
hence we receive:
\[
%
\]

\subsubsection{030\_Bad\_Class (Proof)}
\label{toc:fullishtestsuite:proofs:030}

Let $I$ be an OWL~2 RDF-Based interpretation and $B$ be a blank node mapping 
for the blank nodes in the premise graph such that $I+B$ satisfies the
premise graph.
Let there be an~$x$, such that
\[
%
\]
From (1a) and the semantic condition for class complement 
(OWL2/Tab5.4, ``$\rightarrow$'') 
follows
\[
%
\]
From (1b), (1c), (1d) and the semantic condition for self-restrictions
(OWL2/Tab5.6) follows
\[
%
\]
Now the following equivalence holds:
\[
%
\]
Hence, we receive a contradiction solely from the original settings 
(1a), (1b), (1c) and (1d). That is, the original setting is an 
inconsistent ontology.

\subsubsection{031\_Large\_Universe (Proof)}
\label{toc:fullishtestsuite:proofs:031}

Let $I$ be an OWL~2 RDF-Based interpretation and $B$ be a blank node mapping 
for the blank nodes in the premise graph such that $I+B$ satisfies the
premise graph.
Let there be $x$ and~$l$, such that the following holds:
\[
%
\]
From (1a) and the semantic condition for 
class equivalence (OWL2/Tab5.9, ``$\rightarrow$'') 
follows
\[
%
\]
From (1b), (1c), (1d) and semantic condition for enumeration classes
(OWL2/Tab5.5, ``$\rightarrow$'') follows
\[
%
\]
Since $I$ is a simple interpretation and from the 
class extension of $\texttt{owl:Thing}$
(OWL2/Tab5.2, ``$\leftarrow$'') we receive
\[
%
\]
Applying (2) and (3) to (4a) and (4b), respectively, leads to
\[
%
\]
However, from the class extension of 
$\texttt{owl:Nothing}$ (OWL2/Tab5.2) follows
\[
%
\]
By (7) and~(8) we get a contradiction.
Hence the original setting (1a), (1b), (1c) and~(1d)
is an inconsistent ontology.

\subsubsection{032\_Datatype\_Relationships (Proof)}
\label{toc:fullishtestsuite:proofs:032}

Let $I$ be an OWL~2 RDF-Based interpretation that satisfies the empty graph.

As a consequence of OWL2/Def4.2, 
$I$~must be specified with respect to some 
OWL~2 RDF-Based datatype map~$D$. 
According to OWL2/Def4.1, 
$D$ must include the datatypes denoted by the URIs 
$\texttt{xsd:string}$, 
$\texttt{xsd:integer}$, 
and 
$\texttt{xsd:decimal}$.
The denotations are given by name-datatype pairs~``$(u,d)$'' 
provided by the datatype map, 
and the value spaces are given as~``$\mathrm{VS}(d)$''. 
According to the \emph{``general semantic conditions for datatypes''} 
in the specification of D-entailment, 
the datatypes are identified by 
``$I(\texttt{xsd:string})$'', 
``$I(\texttt{xsd:integer})$'', 
and 
``$I(\texttt{xsd:decimal})$'', 
respectively. 
Secondly, the datatypes
$I(u)$, 
for $u$ one of 
``$\texttt{xsd:string}$'', 
``$\texttt{xsd:integer}$'', 
and ``$\texttt{xsd:decimal}$'', 
are instances of the set $\mathrm{ICEXT}(I(\texttt{rdfs:Datatype}))$. 
OWL2/Tab5.2 implies  
$\mathrm{ICEXT}(I(\texttt{rdfs:Datatype})) 
=
\mathrm{IDC}$. 
From OWL2/Tab5.1 follows 
that $\mathrm{IDC}$ is a sub set of~$\mathrm{IC}$. 
From OWL2/Tab5.2 follows that 
$\mathrm{ICEXT}(I(\texttt{owl:Class})) = \mathrm{IC}$. 
Hence, we get:
\[
\begin{array}{rl}
  (1a) & I(\texttt{xsd:string}) \in \mathrm{IC} \text{ ;} \\
  (1b) & I(\texttt{xsd:integer}) \in \mathrm{IC} \text{ ;} \\
  (1c) & I(\texttt{xsd:decimal}) \in \mathrm{IC} \text{ .}
\end{array}
\]
Further, according to the 
\emph{``general semantic conditions for datatypes''} 
in the specification of D-entailment 
the datatypes have the following value spaces: 
$\mathrm{ICEXT}(I(\texttt{xsd:string}))$, 
$\mathrm{ICEXT}(I(\texttt{xsd:integer}))$, 
and 
$\mathrm{ICEXT}(I(\texttt{xsd:decimal}))$. 
According to OWL2/Def4.1 (referring to the OWL 2 Structural Specification), 
the value spaces of the three datatypes are defined 
according to the XSD Datatype specification. 
This has the following consequences. 
Firstly, the value spaces of 
$\texttt{xsd:decimal}$ and $\texttt{xsd:string}$ are disjoint sets:
\[
\begin{array}{rl}
    (2a) & \forall x:\: \neg [\, 
           x \in \mathrm{ICEXT}(I(\texttt{xsd:decimal})) \;\wedge\; 
           x \in \mathrm{ICEXT}(I(\texttt{xsd:string})) \,] \text{ .}
\end{array}
\]
Secondly, the value space of $\texttt{xsd:integer}$ 
is a subset of the value space of $\texttt{xsd:decimal}$:
\[
\begin{array}{rl}
    (2b) & \forall x:\: 
           x \in \mathrm{ICEXT}(I(\texttt{xsd:integer})) 
           \Rightarrow 
           x \in \mathrm{ICEXT}(I(\texttt{xsd:decimal})) \text{ .}
\end{array}
\]
Using (1c), (1a), (2a), 
and the ``$\leftarrow$'' direction of the semantic condition 
for class disjointness (OWL2/Tab5.9), we get:
\[
\begin{array}{rl}
    (3a) & \langle I(\texttt{xsd:decimal}), I(\texttt{xsd:string})\rangle  \in \mathrm{IEXT}(I(\texttt{owl:disjointWith})) \text{ .}
\end{array}
\]
Using (1b), (1c), (2b), 
and the ``$\leftarrow$'' direction 
of the OWL~2 semantic condition of class subsumption 
(OWL2/Tab5.8), we get:
\[
\begin{array}{rl}
    (3b) & \langle I(\texttt{xsd:integer}), I(\texttt{xsd:decimal})\rangle  
         \in \mathrm{IEXT}(I(\texttt{rdfs:subClassOf})) \text{ .}
\end{array}
\]
The combination of (3a) and (3b) was the conjecture.

\newpage

\section{Translation into TPTP}
\label{toc:tptptranslation}
In Section~\ref{toc:approach} it was explained 
how RDF graphs and the semantics of OWL~2 Full
(see Section~\ref{toc:preliminaries:owl2full})
are translated into FOL,
and Section~\ref{toc:setting:axiomset} mentioned
that the \emph{TPTP language}~\cite{Sut09} 
is used as a concrete FOL serialization syntax.
In this appendix,
the translation into TPTP are demonstrated
by means of a concrete example.
The translation is demonstrated using the test case
\begin{quote}
020\_Logical\_Complications
\end{quote}
from the test suite of \emph{characteristic OWL 2 Full conclusions},
which has been defined in Appendix~\ref{toc:fullishtestsuite}.
The example translation will be complete in the sense 
that the resulting TPTP encoding can be used with FOL ATPs
that understand the TPTP language\footnote
{
The reasoners available online 
as part of the SystemOnTPTP service
can be used for this purpose:
\url{http://www.tptp.org/cgi-bin/SystemOnTPTP/}.
}
in order to obtain the reasoning result of the test case.
The TPTP translations for the example test case
and for all other characteristic conclusions test cases 
are included in the electronic version of the test suite;
see Appendix~\ref{toc:fullishtestsuite} for pointers.
In addition,
the \emph{supplementary material} for this paper
(see the download link at the beginning of Section~\ref{toc:setting})
contains a translation of a large fragment
of the OWL~2 Full semantics into TPTP
(see Section~\ref{toc:setting:axiomset} 
for a characterization of the fragment) 
and provides an executable software tool
for the conversion of arbitrary RDF graphs into TPTP.

\subsection{RDF Graphs and Test Case Data}
\label{toc:tptptranslation:graph}

In this section it is shown how RDF graphs and test case data
are converted into the TPTP language.
Example translations are given
for the premise and conclusion graphs of the entailment test case 
\emph{020\_Logical\_Complications}
from the ``characteristic OWL~2 Full conclusions'' test suite.

According to Section~\ref{toc:fullishtestsuite:testcases:020},
the \emph{premise graph} of the example test case 
is given in Turtle syntax\footnote
{
Turtle RDF syntax: \url{http://www.w3.org/TeamSubmission/turtle/}
} 
as:

\begin{quote}
\begin{verbatim}
@prefix ex:   <http://www.example.org/> .
@prefix rdf:  <http://www.w3.org/1999/02/22-rdf-syntax-ns#> .
@prefix rdfs: <http://www.w3.org/2000/01/rdf-schema#> .
@prefix owl:  <http://www.w3.org/2002/07/owl#> .

ex:c owl:unionOf ( ex:c1 ex:c2 ex:c3 ) .
ex:d owl:disjointWith ex:c1 .
ex:d rdfs:subClassOf [ 
    owl:intersectionOf (
        ex:c 
        [ owl:complementOf ex:c2 ] 
    ) 
] .
\end{verbatim}
\end{quote}

This encoding uses some of the ``syntactic sugar'' 
that Turtle offers for concisely representing certain language constructs, 
such as RDF collections. 
For the purpose of translating the RDF graph into TPTP,
it is advisable to restate the above representation
into an equivalent form that consists of only RDF triples:

\begin{quote}
\begin{verbatim}
@prefix ex:   <http://www.example.org/> .
@prefix rdf:  <http://www.w3.org/1999/02/22-rdf-syntax-ns#> .
@prefix rdfs: <http://www.w3.org/2000/01/rdf-schema#> .
@prefix owl:  <http://www.w3.org/2002/07/owl#> .

ex:c owl:unionOf _:lu1 .
_:lu1 rdf:first ex:c1 .
_:lu1 rdf:rest _:lu2 .
_:lu2 rdf:first ex:c2 .
_:lu2 rdf:rest _:lu3 .
_:lu3 rdf:first ex:c3 .
_:lu3 rdf:rest rdf:nil .
ex:d owl:disjointWith ex:c1 .
ex:d rdfs:subClassOf _:xs .
_:xs owl:intersectionOf _:li1 .
_:li1 rdf:first ex:c .
_:li1 rdf:rest _:li2 .
_:li2 rdf:first _:xc .
_:li2 rdf:rest rdf:nil .
_:xc owl:complementOf ex:c2 .
\end{verbatim}
\end{quote}

Premise graphs of entailment test cases 
are translated into TPTP \emph{axiom} formulae.
Following the explanation in Section~\ref{toc:approach}
on how to translate RDF graphs into FOL,
the translation into TPTP is as follows:

\begin{quote}
\begin{verbatim}
fof(testcase_premise, axiom, (
    ? [B_xs, B_xc, B_lu1, B_lu2, B_lu3, B_li1, B_li2] : (
          iext(uri_owl_unionOf, uri_ex_c, B_lu1)
        & iext(uri_rdf_first, B_lu1, uri_ex_c1)
        & iext(uri_rdf_rest, B_lu1, B_lu2)
        & iext(uri_rdf_first, B_lu2, uri_ex_c2)
        & iext(uri_rdf_rest, B_lu2, B_lu3)
        & iext(uri_rdf_first, B_lu3, uri_ex_c3)
        & iext(uri_rdf_rest, B_lu3, uri_rdf_nil)
        & iext(uri_owl_disjointWith, uri_ex_d, uri_ex_c1)
        & iext(uri_rdfs_subClassOf, uri_ex_d, B_xs)
        & iext(uri_owl_intersectionOf, B_xs, B_li1)
        & iext(uri_rdf_first, B_li1, uri_ex_c)
        & iext(uri_rdf_rest, B_li1, B_li2)
        & iext(uri_rdf_first, B_li2, B_xc)
        & iext(uri_rdf_rest, B_li2, uri_rdf_nil)
        & iext(uri_owl_complementOf, B_xc, uri_ex_c2) ))) .
\end{verbatim}
\end{quote}

The Turtle representation of the \emph{conclusion graph} of the test case 
is given as:

\begin{quote}
\begin{verbatim}
@prefix ex:   <http://www.example.org/> .
@prefix rdf:  <http://www.w3.org/1999/02/22-rdf-syntax-ns#> .
@prefix rdfs: <http://www.w3.org/2000/01/rdf-schema#> .
@prefix owl:  <http://www.w3.org/2002/07/owl#> .

ex:d rdfs:subClassOf ex:c3 .
\end{verbatim}
\end{quote}

Conclusion graphs of entailment test cases 
are translated into TPTP \emph{conjecture} formulae,
which is done as follows: 

\begin{quote}
\begin{verbatim}
fof(testcase_conclusion, conjecture, (
    iext(uri_rdfs_subClassOf, uri_ex_d, uri_ex_c3) )) .
\end{verbatim}
\end{quote}

\subsection{Semantic Conditions of the OWL 2 RDF-Based Semantics}
\label{toc:tptptranslation:semcond}

In this section it is shown 
how the semantic conditions of the OWL~2 RDF-Based Semantics
are translated into the TPTP language.
An example translation is given
for a small subset of semantic conditions that are sufficient
to entail the conclusion graph of the entailment test case 
\emph{020\_Logical\_Complications}
from its premise graph.
The selection of the small sufficient subset of semantic conditions
was made based on the correctness proof for the test case,
as given in Section~\ref{toc:fullishtestsuite:proofs:020}.
The following semantic conditions are used to prove correctness:
\begin{itemize}
  \item 
extension of property \texttt{owl:disjointWith} 
(Section~5.3 of OWL~2 RDF-Based Semantics);
  \item
class complement 
(Section~5.4 of OWL~2 RDF-Based Semantics);
  \item
binary class intersection 
(Section 5.4 of OWL~2 RDF-Based Semantics);
  \item
ternary class union 
(Section 5.4 of OWL~2 RDF-Based Semantics);
  \item
class subsumption, OWL version 
(Section~5.8 of OWL~2 RDF-Based Semantics);
  \item
class disjointness 
(Section~5.9 of OWL~2 RDF-Based Semantics).
\end{itemize}

Semantic conditions are translated into TPTP \emph{axiom} formulae,
since, technically, they act as further premises
in addition to the axiom that represents the premise graph
of a test case.
Following the explanation in Section~\ref{toc:approach}
on how to translate semantic conditions into FOL,
the translation into TPTP is as follows:

\begin{quote}
\begin{verbatim}
% extension of property owl:disjointWith 
% (Section 5.3 of OWL 2 RDF-Based Semantics)
fof(owl_prop_disjointwith_ext, axiom, (
    ! [X, Y] : (
            iext(uri_owl_disjointWith, X, Y)
        => (
              ic(X)
            & ic(Y) )))) .

% class complement 
% (Section 5.4 of OWL 2 RDF-Based Semantics)
fof(owl_bool_complementof_class, axiom, (
    ! [Z, C] : (
            iext(uri_owl_complementOf, Z, C)
        =>
            ( ic(Z)
            & ic(C)
            & ( ! [X] : (
                    icext(Z, X)
                <=>
                    ~ icext(C, X) )))))) .

% binary class intersection 
% (Section 5.4 of OWL 2 RDF-Based Semantics)
fof(owl_bool_intersectionof_class_002, axiom, (
    ! [Z, S1, C1, S2, C2] : ( 
            ( iext(uri_rdf_first, S1, C1)
            & iext(uri_rdf_rest, S1, S2)
            & iext(uri_rdf_first, S2, C2)
            & iext(uri_rdf_rest, S2, uri_rdf_nil) )
        => (
                iext(uri_owl_intersectionOf, Z, S1)
            <=> (
                  ic(Z)
                & ic(C1)
                & ic(C2)
                & ( ! [X] : (
                        icext(Z, X)
                    <=> (
                          icext(C1, X)
                        & icext(C2, X) )))))))) .

% ternary class union 
% (Section 5.4 of OWL 2 RDF-Based Semantics)
fof(owl_bool_unionof_class_003, axiom, (
    ! [Z, S1, C1, S2, C2] : (
           ( iext(uri_rdf_first, S1, C1)
            & iext(uri_rdf_rest, S1, S2)
            & iext(uri_rdf_first, S2, C2)
            & iext(uri_rdf_rest, S2, S3)
            & iext(uri_rdf_first, S3, C3)
            & iext(uri_rdf_rest, S3, uri_rdf_nil) )
        => (
                iext(uri_owl_unionOf, Z, S1)
            <=> ( 
                  ic(Z)
                & ic(C1)
                & ic(C2)
                & ic(C3)
                & ( ! [X] : (
                        icext(Z, X)
                    <=> ( 
                          icext(C1, X)
                        | icext(C2, X)
                        | icext(C3, X) )))))))) .

% class subsumption, OWL version 
% (Section 5.8 of OWL 2 RDF-Based Semantics)
fof(owl_rdfsext_subclassof, axiom, (
    ! [C1, C2] : (
            iext(uri_rdfs_subClassOf, C1, C2)
        <=> ( 
              ic(C1)
            & ic(C2)
            & ( ! [X] : (
                    icext(C1, X)
                =>
                    icext(C2, X) )))))) .

% class disjointness 
% (Section 5.9 of OWL 2 RDF-Based Semantics)
fof(owl_eqdis_disjointwith, axiom, (
    ! [C1, C2] : (
            iext(uri_owl_disjointWith, C1, C2)
        <=> (
              ic(C1)
            & ic(C2)
            & ( ! [X] : ( 
                ~ ( icext(C1, X) 
                  & icext(C2, X) ))))))) .
\end{verbatim}
\end{quote}

}{}

\end{document}